\documentclass[journal]{IEEEtran}
\usepackage{cite}
\ifCLASSINFOpdf
\else
\fi
\usepackage{graphics}                        % for pdf, bitmapped graphics files
\usepackage{graphicx}
\usepackage{varwidth}
\graphicspath{ {./images/} }
\usepackage{booktabs}
\usepackage[flushleft]{threeparttable}
\usepackage{subcaption}
\usepackage{epsfig}                          % for postscript graphics files
\usepackage{mathptmx}                        % assumes new font selection scheme installed
\usepackage{times}                           % assumes new font selection scheme installed
\usepackage{algorithm,algpseudocode}
\usepackage{amsmath}                         % assumes amsmath package installed
\usepackage{amssymb}                         % assumes amsmath package installed
\usepackage{multicol,multirow,blindtext}
\usepackage{booktabs,amsfonts,dcolumn}
\newcolumntype{d}[1]{D..{#1}}

\usepackage{balance}
\usepackage{array}
\usepackage{hyperref}
\usepackage{esvect}
\usepackage{cuted}
\usepackage[dvipsnames]{xcolor}
\usepackage{tabularray}
\usepackage{fixmath}
\usepackage{comment}

\usepackage{xcolor,colortbl}

\definecolor{Gray}{gray}{0.85}
\definecolor{LightCyan}{rgb}{0.88,1,1}
\newcolumntype{a}{>{\columncolor{Gray}}c}

\makeatletter
\newcommand\notsotinyone{\@setfontsize\notsotinyone{6.3}{7}}
\newcommand\notsotinytwo{\@setfontsize\notsotinytwo{5.55}{7}}
\newcommand\notsotinythree{\@setfontsize\notsotinythree{5.73}{7}}
\newcommand\notsotinyfour{\@setfontsize\notsotinyfour{8.87}{10}}
\makeatother

\usepackage{array,etoolbox}
\preto\tabular{\setcounter{magicrownumbers}{0}}
\newcounter{magicrownumbers}

\begin{document}
%
% paper title
% Titles are generally capitalized except for words such as a, an, and, as,
% at, but, by, for, in, nor, of, on, or, the, to and up, which are usually
% not capitalized unless they are the first or last word of the title.
% Linebreaks \\ can be used within to get better formatting as desired.
% Do not put math or special symbols in the title.
\title{Achieving Unit-Consistent Pseudo-Inverse-based Path-Planning for Redundant Incommensurate \\ Robotic Manipulators}
%
%
% author names and IEEE memberships
% note positions of commas and nonbreaking spaces ( ~ ) LaTeX will not break
% a structure at a ~ so this keeps an author's name from being broken across
% two lines.
% use \thanks{} to gain access to the first footnote area
% a separate \thanks must be used for each paragraph as LaTeX2e's \thanks
% was not built to handle multiple paragraphs
%

\author{Jacket~Demby's,~\IEEEmembership{Student Member,~IEEE,}
        Jeffrey~Uhlmann,~\IEEEmembership{Member,~IEEE,}
        and~Guilherme~N.~DeSouza,~\IEEEmembership{Member,~IEEE}% <-this % stops a space
\thanks{All authors are with the Department of Electrical Engineering and Computer Science (EECS), University of Missouri-Columbia, Columbia, Missouri, 65201.}%
\thanks{Jacket Demby's and Guilherme N. DeSouza are with the Vision-Guided and Intelligent Robotics (ViGIR) Laboratory. (email: udembys@mail.missouri.edu; uhlmannj@missouri.edu; desouzag@missouri.edu)}%
%\thanks{Manuscript received April 19, 2005; revised August 26, 2015.}
}

\maketitle
%\noindent\textcolor{red}{[TRANSACTION DRAFT FOR REVIEW]}

% As a general rule, do not put math, special symbols or citations
% in the abstract or keywords.
\begin{abstract}
 In this paper, we review and compare several velocity-level and acceleration-level Pseudo-Inverse-based Path Planning (PPP) and  Pseudo-Inverse-based Repetitive Motion Planning (PRMP) schemes based on the kinematic model of robotic manipulators. We show that without unit consistency in the pseudo-inverse computation, path planning of incommensurate robotic manipulators will fail. Also, we investigated the robustness and noise tolerance of six PPP and PRMP schemes in the literature against various noise types (i.e. zero, constant, time-varying and random noises). We compared the simulated results using two redundant robotic manipulators: a 3DoF (2RP), and a 7DoF (2RP4R). These experimental results demonstrate that the improper Generalized Inverse (GI) with arbitrary selection of unit and/or in the presence of noise can lead to unexpected behavior of the robot, while producing wrong instantaneous outputs in the task space, which results in distortions and/or failures in the execution of the planned path. Finally, we propose and demonstrate the efficacy of the Mixed Inverse (MX) as the proper GI to achieve unit-consistency in path planning.

\end{abstract}


\section{Introduction}\label{sec:intro}
\IEEEPARstart{P}{ath} planning relies on formal schemes to control robotic manipulators to follow a desired Cartesian trajectory while avoiding joint limits, singularities, and obstacles \cite{faroni2018predictive, faroni2020inverse, park2020trajectory, ademovic2016path, lacevic2020improved, guo2020repetitive, chan1995weighted}. These schemes are of great interest to the robotics community because they span many engineering fields and real-world robotic applications (e.g. assembly, welding, handling tasks, surgery, etc.)\cite{pardi2020path, guo2020repetitive, corke2011robotics}. In essence, the schemes offer different ways to relate a desired trajectory defined by a sequence of Cartesian locations, $D(t) \in \mathbb{R}^{m}$, to the corresponding values in joint space, $Q(t) \in \mathbb{R}^{n}$ -- where $m$ is the dimension of the Cartesian or task space and $n$ the number of Degrees of Freedom (DoF) of the manipulator \cite{chan1995weighted, flacco2012motion, flacco2015control, guo2017new, wang2019feedback}. In this same regard, Repetitive Motion Planning (RMP) focuses on achieving repeatability in path planning tasks \cite{lanari1992control ,guo2020repetitive, li2018new, xie2021acceleration}.

\begin{figure}[!t]
\centering

\begin{subfigure}{.22\textwidth}
  \centering
  % include first image
  \includegraphics[width=4.1cm,trim={1cm 1cm 1cm 2cm},clip]{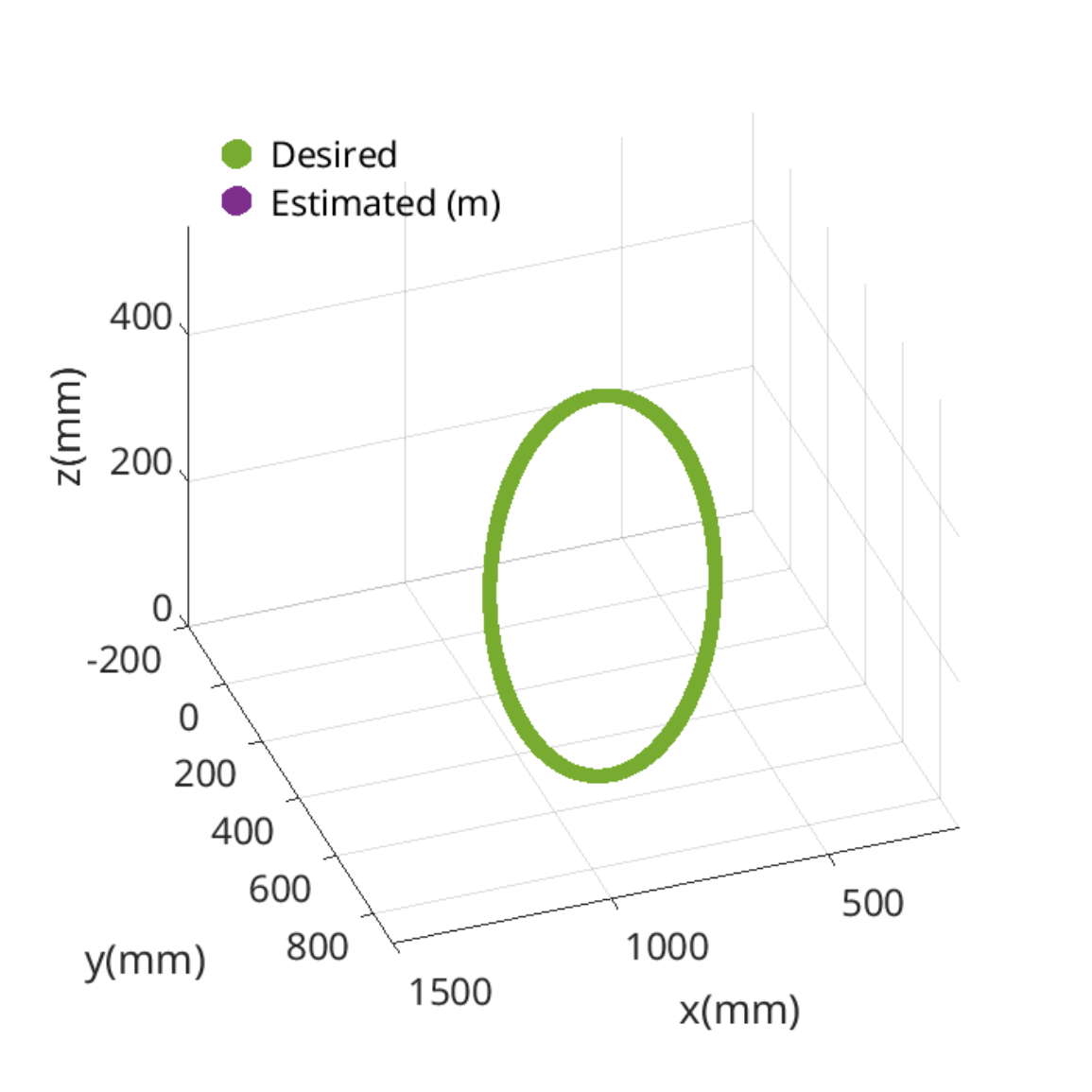}  
  \caption{\footnotesize  MP / m / zero noise}
  \label{fig:7dof-m-mp-zero}
\end{subfigure}
\hspace{6mm}
\begin{subfigure}{.22\textwidth}
  \centering
  % include third image
  \includegraphics[width=4.1cm,trim={1cm 1cm 1cm 2cm},clip]{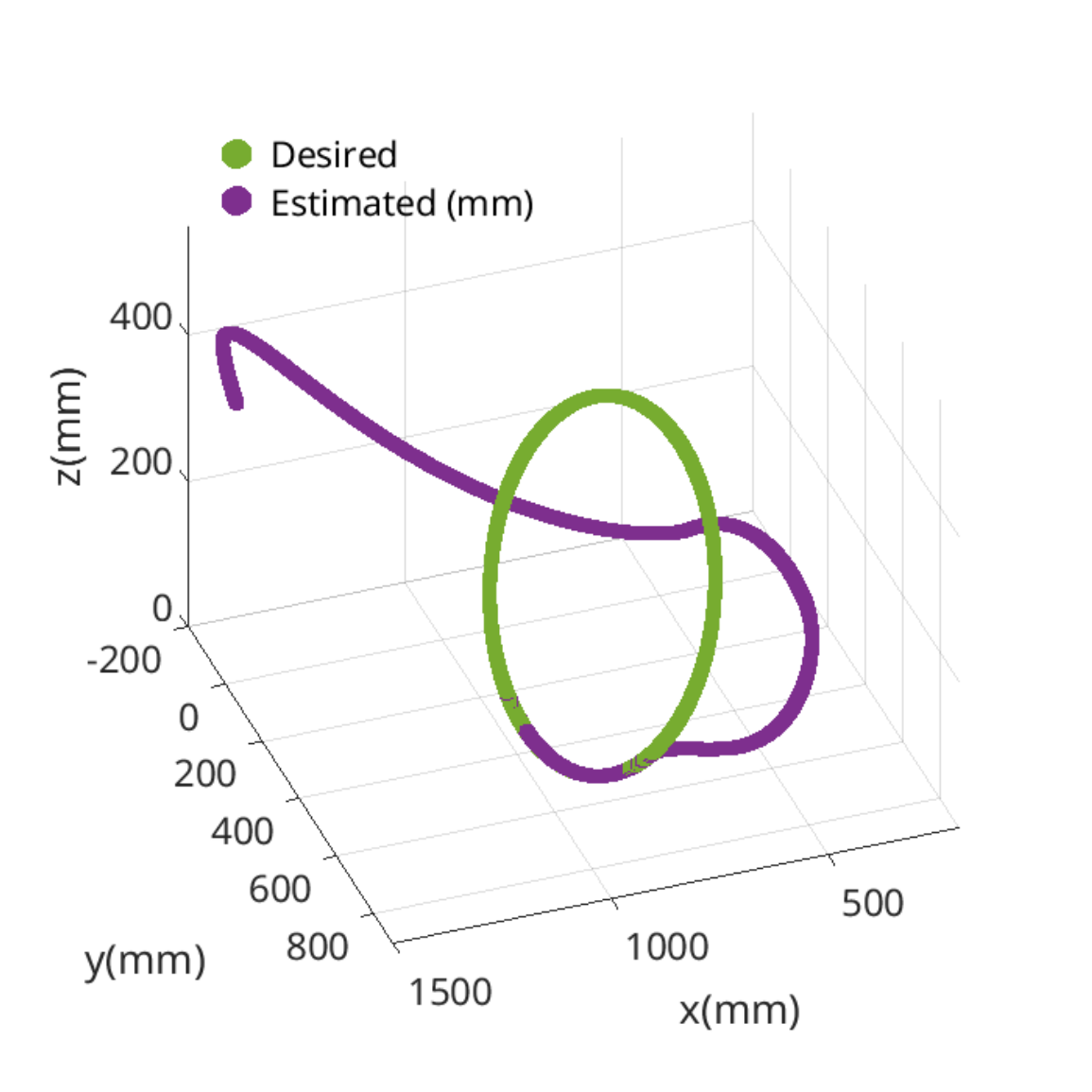}  
  \caption{\footnotesize MP / mm / zero noise}
  \label{fig:7dof-mm-mp-zero}
\end{subfigure}

\caption{\footnotesize Desired path (blue) versus estimated path (red) for the 7DoF (2RP4R) using the MVN scheme based on the Moore-Penrose (MP) Generalized Inverse while varying the units of the prismatic joint from $m$ to $mm$.}
\label{fig:path-planning-7DoF-MP}
\vspace{-4mm}
\end{figure}

Over the years, various schemes for Pseudo-inverse-based Path Planning (PPP) and Pseudo-inverse-based Repetitive Motion Planning (PRMP) have been proposed \cite{chan1995weighted, flacco2012motion, flacco2015control, guo2017new, wang2019feedback, lanari1992control ,guo2020repetitive, li2018new}. PPP and PRMP are usually formulated either up to the joint velocity level or the joint acceleration level. In that sense, Jacobians and Generalized Inverse (GI) Jacobians are fundamental notions and heavily incorporated in most (if not all) path planners. They must be accurately estimated using respectively: 1) analytical \cite{sciavicco2001modelling}, numerical \cite{farzan2013dh}, geometric \cite{mayer1981differential, sciavicco2001modelling}, and elementary transform sequence \cite{haviland2020systematic}) approaches (for the Jacobian); and 2) Moore-Penrose \cite{penrose1955generalized}, Error Damping \cite{chan1988general}, Filtered Jacobian \cite{chiaverini1991achieving}, Damped Jacobian \cite{chiaverini1994review}, Selective Damping \cite{buss2005selectively}, Improved Error Damping \cite{sugihara2011solvability}, Singular Value Filtering \cite{colome2012redundant}, Unit-Consistent \cite{uhlmann2018generalized}, and Mixed GI's \cite{uhlmann2018generalized} (for the inverse Jacobians). Mathematically, most methods in the literature derived from a Velocity-level PPP scheme (V-PPP) or an Acceleration-level PPP scheme (A-PPP). In V-PPP, the relationship between $D$ and $Q$ is formulated as the summation of two terms: a minimum-norm solution and a homogeneous solution \cite{sciavicco2001modelling, guo2017new, wang2019feedback, angeles2003fundamentals}:
\begin{equation}\label{eq:velocify-formulation}
    \dot{Q} = J^{\widetilde{-1}}\dot{D}_{d}+(I-J^{\widetilde{-1}}J)z
\end{equation}
where $\dot{Q} \in \mathbb{R}^{n}$ is the first-order partial derivative of the joint vector $Q$ with respect to time, $\dot{D}_{d} \in \mathbb{R}^{m}$ is the first-order time derivative of the desired Cartesian vector $D_{d}$, $I \in \mathbb{R}^{n\mathrm{x}n}$ is the identity matrix, $z \in \mathbb{R}^{n}$ is an arbitrary vector selected via some optimal criteria at the velocity level, and $J^{\widetilde{-1}}  \in \mathbb{R}^{n\mathrm{x}m}$ is a GI of the Jacobian matrix $J \in R^{m\mathrm{x}n}$. Similarly, A-PPP is mathematically formulated as it follows \cite{sciavicco2001modelling, wang2019feedback, angeles2003fundamentals}:
\begin{equation}\label{eq:acceleration-formulation}
    \ddot{Q} = J^{\widetilde{-1}}(\ddot{D_{d}} - \dot{J}\dot{Q})+(I-J^{\widetilde{-1}}J)z
\end{equation}
where $\ddot{Q} \in \mathbb{R}^{n}$ is the second-order time derivative of the joint vector $Q$, $\ddot{D}_{d} \in \mathbb{R}^{m}$ is the second-order time derivative of the desired Cartesian vector $D_{d}$, $z \in \mathbb{R}^{n}$ is an arbitrary vector selected via some optimal criteria, this time at the acceleration level, and $\dot{J} \in \mathbb{R}^{m\mathrm{x}n}$ is the first order time derivative of the Jacobian matrix $J \in R^{m\mathrm{x}n}$. Setting $z=0$ in equations (\ref{eq:velocify-formulation}) and (\ref{eq:acceleration-formulation}) reduces these equations respectively to the Minimum Velocity Norm (MVN) and the Minimum Acceleration Norm (MAN) \cite{sciavicco2001modelling, siciliano2008springer}.

The Moore-Penrose \cite{penrose1955generalized} ($J^{\dagger}$) is certainly the most widely employed GI ($J^{\widetilde{-1}} \Leftarrow  J^{\dagger}$) for PPP and PRMP. However, very little to no attention has been placed on the type of GI to guarantee and preserve unit-consistency and noise tolerance of the published methods. Recently, it has been shown that the Jacobian-based Inverse Kinematics solver using the Moore-Penrose GI often fails to preserve these same properties in the case of incommensurate robotic manipulators \cite{uhlmann2018generalized, zhang2019applying, zhang2020examining, demby2020use,dembys2023choosing}. As the reader probably knows, incommensurate concerns all sequential manipulators having a combination of prismatic (linear) and revolute (rotational) joints. Such manipulators may have its variables expressed in different units -- e.g. end-effector poses $\vec{D}$ expressed using meters, or centimeters, etc.; and prismatic joints, i.e. $\vec{Q} = [Q_1, Q_2, .., Q_i]$ where $Q_i = d_i$, may again be expressed in meters, or centimeters, etc. \cite{schwartz2002noncommensurate, schwartz2003non}. While these manipulators require consistency with respect to the choice of a single unit for all its variables (e.g. in the Denavit-Hartenberg, DH, representation), the use of some GI's -- often in the literature, the Moore-Penrose -- can violate the same unit consistency in the calculated inverse Jacobian \cite{uhlmann2018generalized, zhang2019applying, zhang2020examining, demby2020use,dembys2023choosing}. This problem can be understood as having an inverse Jacobian solution that exhibits unexpected sensitivity to the choice of unit, when in fact the system should not be affected by such choice \cite{schwartz2002noncommensurate, uhlmann2018generalized, zhang2019applying, zhang2020examining, demby2020use, dembys2023choosing}. The main question here is how significant this problem is for PPP and PRMP schemes, moreover in the presence of various noise types. As a motivating example, Figure \ref{fig:path-planning-7DoF-MP} shows the behavior of a 7DoF redundant incommensurate robot when following a simple circular path using the Moore-Penrose GI for the calculation of MVN (eq. \ref{eq:velocify-formulation}). We observe that the manipulator successfully follows the desired path when its prismatic joint is expressed in $m$, however, it fails to do so when the chosen unit is $mm$. This paper brings to light these undesired behaviors, often overlooked by the robotics community, and addresses the above question with the following main contributions:
\begin{enumerate}
    \item to investigate the effects of change of units in PPP and PRMP schemes;
    \item to review and compare well-established PPP and PRMP schemes under various noise types (e.g., zero, constant, time-varying) found in the literature; 
    \item to solve this problem by applying the Mixed Inverse (MX) which achieves unit-consistency in PPP and PRMP schemes;
    \item to demonstrate that even under various noise types, the MX-GI retains unit-consistent  properties for PPP and PRMP. That is, it allows for the design of reliable and robust (i.e. stable) schemes.
\end{enumerate}

\section{Background and Related Work}\label{sec:background}
As mentioned in section \ref{sec:intro}, over the year, many different schemes have been derived from the V-PPP and the A-PPP schemes. However, the Moore-Penrose GI (MP-GI) has been widely adopted for all such cases. Early work from Klein and Wang \cite{klein1983review} reviewed three pseudo-inverses (Left-Pseudo-inverse, Right-Pseudo-inverse, Moore-Penrose GI) used in PPP schemes. The authors proposed to supplement the MVN scheme with an additional homogeneous term based on an optimal criterion to be minimized subject to the optimal joint velocities. In this regard, several optimal criteria have been introduced \cite{zghal1990efficient, chan1995weighted, sciavicco2001modelling, siciliano2008springer}. De Luca \textit{et al} \cite{lanari1992control} started looking at necessary and sufficient conditions for achieving optimal repeatability and cyclicity in periodic PRMP tasks of redundant manipulators.  Recently,   Guo \textit{et al}. \cite{guo2017new} proposed a new velocity-level Proportional-Integral-Derivative-based (PID-based) scheme for the PPP tasks in the presence of various noise types (e.g., zero, constant, time-varying). Their scheme was designed to take into account the proportional, integral, and derivative information of the desired Cartesian end-effector trajectory. In \cite{guo2020repetitive}, Guo \textit{et al}. used their PID-bassed PPP scheme for PRMP tasks, hence achieving highly precise joint angle repeatability and end-effector motion. Moreover, Li \textit{et al}. \cite{li2017novel} built on previous work from  \cite{guo2017new} to design a P-based RMP scheme with noise suppression capabilities. In this work, the velocity-level PRMP scheme was also reformulated as a quadratic programming (QP) problem to guarantee the optimality of the solution found. Wang \textit{et al}. \cite{wang2019feedback} introduced an acceleration-level feedback-added PPP scheme based on a weighted combination of the MAN \cite{sciavicco2001modelling, siciliano2008springer} scheme and the Weighted MVN (WMVN) \cite{chan1995weighted, deo1997minimum} scheme reformulated up to the acceleration-level. In \cite{flacco2012motion, flacco2015control}, Flacco \textit{et al} introduced the Saturation in the Null Space (SNS) method for addressing online inversion of differential task kinematics for redundant manipulators, in the presence of hard limits of joint space motion. Their method integrates task scaling and task prioritization, and combine joint limits into hard constraints at specific robot configurations. All the aforementioned studies have the following points in common: (1) they mainly used the Moore-Penrose GI in their schemes, (2) they only focused on commensurate manipulators (made of only revolute joints),  and (3) they did not investigate the stability and robustness of their schemes in the case of incommensurate robots. More recently, Uhlmann \cite{uhlmann2018generalized} developed two GI's: UC and MX applicable to incommensurate and commensurate robots. In  \cite{zhang2019applying}, Zhang and Uhlmann used an incommensurate robot to demonstrate that the MP-GI was making the system unstable when the units were varied. With the UC GI, they achieved a stable IK solution. Similarly, Zhang and Uhlmann \cite{zhang2020examining} examined the MP, UC and MX-GI's with an incommensurate system consisting a robotic arm attached to a rover. They showed that the MP failed to preserve consistencies with respect to changes of units while the UC failed to preserve consistencies with respect to changes in rotation of the coordinate frame. Interestingly, a combination of these two GI's in the MX-GI  was able to achieve a unit-consistent behavior for the system. In these studies, path planning was not investigated and the robotic systems were assumed to be without noise. That is, in this paper, we employ the MX-GI rule of thumb developed in \cite{dembys2023choosing} to integrate the MX-GI in several PPP and PRMP schemes, and evaluate their unit-consistency properties and noise tolerance characteristics.

\begin{table*}[!ht] 
\footnotesize
\caption{\footnotesize Investigated Pseudo-inverse-based Path-Planning (PPP) and Repetitive Motion Planning (PRMP) schemes for robotic manipulators}\label{tab:surveyed-schemes-methodologies}
    \centering
    \begin{threeparttable}
         \begin{tabular}{|l|l|c|c|} 
         %\begin{tblr}{
         %       colspec = {|c|c|c|c|},
                %row{4} = {gray9},
                %row{12} = {gray9},
                %row{14} = {gray9},
                %row{22} = {gray9},
                %row{31} = {gray9},
                %row{32} = {gray9},
                %column{3} = {teal7},
                %cell{2}{3} = {yellow7},
        %      }
         \hline
          \textbf{Schemes} & \textbf{Equations} & \textbf{Robots Used} & \textbf{References}\\  
         \hline\hline
         \multicolumn{4}{|c|}{\textbf{Velocity-level schemes}}\\
         \hline
         PID-PPP & $\dot{Q} = J^{\dagger}\big(\dot{D}_{d} - \alpha(f(Q) - D_{d}) - \beta\int_{0}^{t}(f(Q) - D_{d})d\tau)\big)$ & 4R, 7R & \cite{guo2017new} \\
         \hline
         WMVN & $\dot{Q} = J^{\dagger}_{W}\dot{D_{d}} = W^{-1}J^{T}(JW^{-1}J^{T})^{-1}\dot{D_{d}}$ & 7R, 50R & \cite{chan1995weighted, deo1997minimum, sciavicco2001modelling}  \\
         \hline
         SNS-V & $\dot{Q} = (JW)^{\dagger}\dot{D}_{d} + (I - (JW)^{\dagger}J)\dot{Q}_{N} = \dot{Q}_{N} + (JW)^{\dagger}(\dot{D}_{d} - J\dot{Q}_{N}) $ & 7R & \cite{flacco2012motion, flacco2015control} \\
         \hline
         \multicolumn{4}{|c|}{\textbf{Acceleration-level schemes}}\\ 
         \hline
         MAN & $\ddot{Q} = J^{\dagger}(\ddot{D_{d}} - \dot{J}\dot{Q})$ & 4R, 7R & \cite{sciavicco2001modelling, siciliano2008springer}   \\
         \hline
         FPBM & $\ddot{Q} = \big(\alpha J^{\dagger}_{W} + (1-\alpha)J^{\dagger}\big)\big(\ddot{D_{d}} - \dot{J}\dot{Q} + k_{1}(\dot{D_{d}} - J\dot{Q}) + k_{2}(D_{d} -f(Q))\big) + \alpha(I - J^{\dagger}_{W}J)W^{-1}\dot{J}^{T}(JW^{-1}J^{T})^{-1}\dot{D_{d}}$ & 4R & \cite{wang2019feedback} \\
         \hline
         SNS-A & $\ddot{Q} = (JW)^{\dagger}(\ddot{D}_{d} - \dot{J}\dot{Q}) + (I - (JW)^{\dagger}J)\ddot{Q}_{N} = \ddot{Q}_{N} + (JW)^{\dagger}(\ddot{D}_{d} - \dot{J}\dot{Q} - J\ddot{Q}_{N}) $ & 7R, 50R & \cite{flacco2012motion, flacco2015control} \\
         \hline
        %\end{tblr}
        \end{tabular}
        % Note under the table
        \begin{tablenotes}
        \item
        \end{tablenotes}
    \end{threeparttable}
\vspace{-4mm}
\end{table*}

\begin{table}[!htb] %!htb
\scriptsize
    \caption{\footnotesize D-H Parameters of the serial robots used in the experiments. The angles $\theta$ and $\alpha$ are expressed in degrees. The variables $d$ and $a$ are shown in millimeters ($mm$) – but they were changed in the implementation to match the different choices of units.}
    \begin{subtable}{.5\linewidth}
      \centering
         \begin{tabular}{|c | c | c | c | c|} 
         \hline
         $i$ & $\theta$ & $d$ & $a$ & $\alpha$  \\  
         \hline\hline
         1 & $\theta_1$ & 0 & 1000 & 0\\ 
         \hline
         2 & $\theta_2$ & 0 & 1100 & 90 \\
         \hline
         3 & 0 & $d_3$ & 0 & 0 \\
         \hline
        \end{tabular}
        \medskip
        \caption{{\footnotesize 3 DoF Planar arm \cite{zhang2019applying}}}
        \label{tab:3DoF-DH-parameters}
    \end{subtable}%
    \medskip
    \begin{subtable}{.5\linewidth}
      \centering
         \begin{tabular}{|c | c | c | c | c|} 
         \hline
         $i$ & $\theta$ & $d$ & $a$ & $\alpha$  \\  
         \hline\hline
         1 & $\theta_1$ & 0 & 0 & 90\\ 
         \hline
         2 & $\theta_2$ & 0 & 250 & 90 \\
         \hline
         3 & 0 & $d_3$ & 0 & 0 \\
         \hline
         4 & $\theta_4$ & 0 & 0 & 90 \\
         \hline
         5 & $\theta_5$ & 140 & 0 & 90 \\
         \hline
         6 & $\theta_6$ & 0 & 0 & 90 \\
         \hline
         7 & $\theta_7$ & 0 & 0 & 0 \\
         \hline
        \end{tabular}
        \medskip
        \caption{{\footnotesize  7 DoF GP66+1 arm}}
        \label{tab:7DoF-DH-parameters}
    \end{subtable} 

    \label{tab:DH-parameters}
\vspace{-4mm}
\end{table}

\begin{algorithm}[!t]
\caption{\footnotesize Moore-Penrose Generalized Inverse (MP-GI)}\label{alg:MP}
\footnotesize
\begin{algorithmic}[1] 
\Procedure{MP-GI}{$J$}
\State $\text{\textbf{Step 1}: Compute Singular Valued Decomposition (SVD) of } J$ 
\State $[U,S,V] = SVD(J)$ 
\State $\text{\textbf{Step 2}: Compute MP-GI (} J^{-MP} \text{) of } J$ 
\State $J^{-MP} = (USV^{*})^{-MP} = VS^{-1}U^{*}$ 
\State \textbf{return} $J^{-MP}$ 
\EndProcedure 
\end{algorithmic}
\label{alg:MP-1}
\end{algorithm}

\begin{algorithm}[!t]
\caption{\footnotesize Unit-Consistent Generalized Inverse (UC-GI)}\label{alg:UC}
\footnotesize
\begin{algorithmic}[1] 
\Procedure{UC-GI}{$J$}
\State $\text{\textbf{Step 1}: Compute Scaling Decomposition (SD) of } J$ 
\State $[E,S_{UI},D] = SD(J)$ 
\State $\text{\textbf{Step 2}: Compute Singular Valued Decomposition (SVD) of } S_{UI}$ 
\State $[U_{S_{UI}},S_{S_{UI}},V_{S_{UI}}] = SVD(S_{UI})$ 
\State $\text{\textbf{Step 3}: Compute UC-GI (} J^{-UC} \text{) of } J$ 
\State $J^{-UC} = \big(E(U_{S_{UI}}S_{S_{UI}}V_{S_{UI}}^{*})D\big)^{-UC} = E^{-1}VS^{-1}U^{*}D^{-1}$ 
\State \textbf{return} $J^{-UC}$ 
\EndProcedure 
\end{algorithmic}
\label{alg:UC-1}
\end{algorithm}

\begin{algorithm}[!t]
\caption{\footnotesize Mixed Generalized Inverse (MX-GI)}\label{alg:MX}
\footnotesize
\begin{algorithmic}[1] 
\Procedure{MX-GI}{$J$}
\State $\text{\textbf{Step 1}: Partition } J \text{ into } A_{W}, A_{X}, A_{Y}, A_{Z} \text{ block matrices according to}$ 
\State $\text{the rule of thumb developed in \cite{dembys2023choosing} as expressed in Equation (\ref{eq:partition})}$ 
\State $\text{\textbf{Step 2}: Compute MX-GI (} J^{-MX} \text{) of } J \text{ based on Equation (\ref{eq:MX-inverse})}$
\State \textbf{return} $J^{-MX}$ 
\EndProcedure 
\end{algorithmic}
\label{alg:MX-1}
\end{algorithm}

\section{Proposed Methodology}
Table \ref{tab:surveyed-schemes-methodologies} summarizes the equations characterizing the velocity-level and acceleration-level PPP and PRMP schemes investigated in this paper alongside with their investigated commensurate robots and citations. As mentioned in Section \ref{sec:background}, all these algorithms were developed and evaluated using the MP-GI to calculate the inverse Jacobians ($J^{\dagger}$). Furthermore, their robustness and stability were not demonstrated on incommensurate robotic manipulators. In this paper, for each of the schemes presented in Table \ref{tab:surveyed-schemes-methodologies}, we replaced the MP-GI by the MX-GI ($J^{\dagger} \Leftarrow  J^{-MX}$) to achieve rotation and unit consistent properties of the PPP and PRMP (see Figure \ref{fig:path-planning-7DoF-MP}) for incommensurate robots. That is, the trajectories followed by the robot under all units are guaranteed to be exactly the same. The MX-GI is derived using the concept of block matrix inverse \cite{lu2002inverses}, where columns of the matrix related to variables requiring unit consistency (e.g. $\partial/\partial Q_{i}$, where $Q_{i}$ requires unit consistency) must be block-partitioned at the top left and bottom left of $J$, and the remaining columns for variables not requiring unit consistency, at the top and bottom right of $J$ \cite{uhlmann2018generalized, zhang2020examining, demby2020use, dembys2023choosing}. This partitioning is expressed as:
\begin{equation}\label{eq:partition}
    J = \begin{bmatrix}
        A_W & A_X\\
        A_Y & A_Z
        \end{bmatrix}
\end{equation}
where $A_W$ relates to the block of variables requiring unit-consistency, $A_Z$ relates the block of variables not requiring unit consistency, $A_X$ and $A_Y$ are remaining blocks after moving the columns of $J$. Once, the blocks of $J$ have been partitioned, the block-matrix inverse \cite{uhlmann2018generalized} is applied to $J$ to compute its MX-GI $J^{-MX}$ as follows:
\begin{equation} \label{eq:MX-inverse}
\begin{split}
    J^{-MX} &= \left[\begin{array}{cc}
    B_{11} & B_{12}\\
    B_{21} & B_{22}
    \end{array}\right] \\
    B_{11} &= (A_W-A_XA_Z^{-MP}A_Y)^{-UC} \\
    B_{12} &= -A_W^{-UC}A_{X}(A_Z-A_YA_W^{-UC}A_X)^{-MP} \\
    B_{21} &= -A_Z^{-MP}A_Y(A_W-A_XA_Z^{-MP}A_Y)^{-UC} \\
    B_{22} &= (A_Z-A_YA_W^{-UC}A_X)^{-MP}
\end{split}
\end{equation}   

%\footnotesize
%\begin{equation} \label{eq:MX-inverse}
%\begin{split}
%    J^{-M} &= \left[\begin{array}{cc}
%    (A_W-A_XA_Z^{-P}A_Y)^{-U} & -A_W^{-U}A_{X}(A_Z-A_YA_W^{-U}A_X)^{-P}\\
%    -A_Z^{-P}A_Y(A_W-A_XA_Z^{-P}A_Y)^{-U} & (A_Z-A_YA_W^{-U}A_X)^{-P}
%    \end{array}\right]
%\end{split}
%\end{equation}  
%{\small{}\noindent$B_{11} = (A_W-A_XA_Z^{-P}A_Y)^{-U}, \quad B_{12} = -A_W^{-U}A_{X}(A_Z-A_YA_W^{-U}A_X)^{-P},$\small{}}
%{\small{}\noindent$B_{21} = -A_Z^{-P}A_Y(A_W-A_XA_Z^{-P}A_Y)^{-U},  \quad B_{22} = (A_Z-A_YA_W^{-U}A_X)^{-P}.$\small{}}  
\noindent where $(.)^{-MP} \Leftarrow  (.)^{\dagger}$ is the MP-GI \cite{penrose1955generalized} and $(.)^{-UC}$ the UC-GI \cite{uhlmann2018generalized}. In a nutshell, the MX-GI is a combination of the MP-GI and UC-GI based on the concept of block matrix inverse. We refer the readers to \cite{uhlmann2018generalized} for all the mathematical proofs related to the derivation of the UC-GI. Algorithms \ref{alg:MP},  \ref{alg:UC}, and \ref{alg:MX} provide the main steps for computing the MP-, UC-, and MX-GI's, respectively.

The fundamental challenge in the use of the MX-GI remains the block-partitioning of the matrix to invert. In that regard, we employ the rule of thumb for the use of the MX-GI as established in \cite{demby2020use, dembys2023choosing}: all revolute joints appearing before a prismatic joint of interest whose $Z-axis$  are not parallel need to be included in $A_W$. That is because if the $Z-axis$ of a revolute joint prior to a prismatic joint are not parallel (i.e. causing a projection of the $Z-axis$ of the revolute joint to the $Z-axis$ of the following prismatic joint), a rotation caused by the revolute joint will affect the prismatic joint, which will violate unit-consistency unless they are placed in the $A_W$ block partitioning where they can be handled by the UC-GI. On the other hand, if the two $Z-axis$ are parallel, the revolute joint will not affect the prismatic joint, and hence they need to be placed in $A_Z$, so that they can be handled by the MP-GI.

%Finally, a variable requiring unit consistency can be identified by the order of rotational and prismatic joints of the sequential manipulator (e.g. from the DH representation). In fact, whenever the Z axis of a rotational joint is not aligned (i.e. causes a projection of) to the Z axis of a subsequent prismatic joint, the latter becomes affected and hence must be regarded as requiring unit consistency \cite{demby2020use}.

\subsection{Application of the MX-GI to the 3DoF-2RP robot}
The Jacobian $J$ of the 3DoF robot is given by:

\begin{equation} \label{eq12}
{\small{} J = \begin{bmatrix}
\frac{\partial X}{\partial \theta_1} & \frac{\partial X}{\partial \theta_2} & \frac{\partial X}{\partial d_3}\\
\frac{\partial Y}{\partial \theta_1} & \frac{\partial Y}{\partial \theta_2} & \frac{\partial Y}{\partial d_3}
\end{bmatrix}{\small} = \begin{bmatrix}
A_W & A_X\\
A_Y & A_Z
\end{bmatrix}
}
\end{equation}
Based on the rule of thumb developed in \cite{dembys2023choosing, demby2020use}, the MX-GI $J^{-M}$ of $J$ is given by:
\begin{equation}\label{eq:MX-3DoF}
    J^{-M}= \begin{bmatrix}
            A_W^{-U} & 0\\
            0 & 0
        \end{bmatrix}
\end{equation}
where $A_W = J$ is the entire 2x3 matrix; $A_X = [0 \;\; 0]^T$ is a 2x1 matrix of zeros, $A_Y = [0 \;\; 0 \;\; 0]$ is 1x3 matrix of zeros and $A_Z = [0]$ is a 1x1 matrix of zeros. The resulting $J^{-M}$ inverse Jacobian matrix is a 4x3 matrix with one row and one column of zeros.

\subsection{Application of the MX-GI to the 7DoF-2RP4R robot}
The Jacobian $J$  of the 7DoF robot is given by:

\begin{equation} \label{eq14}
{\small{} J = \begin{bmatrix}
\frac{\partial X}{\partial \theta_1} & \frac{\partial X}{\partial \theta_2} & \frac{\partial X}{\partial d_3} & \frac{\partial X}{\partial \theta_4} & \frac{\partial X}{\partial \theta_5} & \frac{\partial X}{\partial \theta_6} & \frac{\partial X}{\partial \theta_7}\\
\frac{\partial Y}{\partial \theta_1} & \frac{\partial Y}{\partial \theta_2} & \frac{\partial Y}{\partial d_3} & \frac{\partial Y}{\partial \theta_4} & \frac{\partial Y}{\partial \theta_5} & \frac{\partial Y}{\partial \theta_6} & \frac{\partial Y}{\partial \theta_7}\\
\frac{\partial Z}{\partial \theta_1} & \frac{\partial Z}{\partial \theta_2} & \frac{\partial Z}{\partial d_3} & \frac{\partial Z}{\partial \theta_4} & \frac{\partial Z}{\partial \theta_5}  & \frac{\partial Z}{\partial \theta_6}  & \frac{\partial Z}{\partial \theta_7}\\
\frac{\partial Ro}{\partial \theta_1} & \frac{\partial Ro}{\partial \theta_2} & \frac{\partial Ro}{\partial d_3} & \frac{\partial Ro}{\partial \theta_4} & \frac{\partial Ro}{\partial \theta_5}  & \frac{\partial Ro}{\partial \theta_6}  & \frac{\partial Ro}{\partial \theta_7}\\
\frac{\partial Pi}{\partial \theta_1} & \frac{\partial Pi}{\partial \theta_2} & \frac{\partial Pi}{\partial d_3} & \frac{\partial Pi}{\partial \theta_4} & \frac{\partial Pi}{\partial \theta_5}  & \frac{\partial Pi}{\partial \theta_6}  & \frac{\partial Pi}{\partial \theta_7}\\
\frac{\partial Ya}{\partial \theta_1} & \frac{\partial Ya}{\partial \theta_2} & \frac{\partial Ya}{\partial d_3} & \frac{\partial Ya}{\partial \theta_4} & \frac{\partial Ya}{\partial \theta_5}  & \frac{\partial Ya}{\partial \theta_6}  & \frac{\partial Ya}{\partial \theta_7}
\end{bmatrix} = \begin{bmatrix}
A_W & A_X\\
A_Y & A_Z
\end{bmatrix}
\small{}}
\end{equation}

\noindent Here also, based on the rule of thumb developed in \cite{dembys2023choosing, demby2020use}, the MX-GI $J^{-M}$ of $J$ is given by equation \ref{eq:MX-inverse} based-on the following block-partitioning of $J$:

{\small{}\noindent$A_W = \left[\begin{array}{ccc}
\frac{\partial X}{\partial \theta_1} & \frac{\partial X}{\partial \theta_2} & \frac{\partial X}{\partial d_3}\\
\frac{\partial Y}{\partial \theta_1} & \frac{\partial Y}{\partial \theta_2} & \frac{\partial Y}{\partial d_3}\\
\frac{\partial Z}{\partial \theta_1} & \frac{\partial Z}{\partial \theta_2} & \frac{\partial Z}{\partial d_3}
\end{array}\right], \quad A_X = \left[\begin{array}{cccc}
\frac{\partial X}{\partial \theta_4} & \frac{\partial X}{\partial \theta_5} & \frac{\partial X}{\partial \theta_6} & \frac{\partial X}{\partial \theta_7}\\
\frac{\partial Y}{\partial \theta_4} & \frac{\partial Y}{\partial \theta_5} & \frac{\partial Y}{\partial \theta_6} & \frac{\partial Y}{\partial \theta_7} \\
\frac{\partial Z}{\partial \theta_4} & \frac{\partial Z}{\partial \theta_5} & \frac{\partial Z}{\partial \theta_6} & \frac{\partial Z}{\partial \theta_7}
\end{array}\right],$\small{}}

{\small{}\noindent$A_Y = \left[\begin{array}{ccc}
\frac{\partial Ro}{\partial \theta_1} & \frac{\partial Ro}{\partial \theta_2} & \frac{\partial Ro}{\partial d_3}\\
\frac{\partial Pi}{\partial \theta_1} & \frac{\partial Pi}{\partial \theta_2} & \frac{\partial Pi}{\partial d_3}\\
\frac{\partial Ya}{\partial \theta_1} & \frac{\partial Ya}{\partial \theta_2} & \frac{\partial Ya}{\partial d_3}
\end{array}\right], \quad A_Z = \left[\begin{array}{cccc}
\frac{\partial Ro}{\partial \theta_4} & \frac{\partial Ro}{\partial \theta_5} & \frac{\partial Ro}{\partial \theta_6}  & \frac{\partial Ro}{\partial \theta_7}\\
\frac{\partial Pi}{\partial \theta_4} & \frac{\partial Pi}{\partial \theta_5} & \frac{\partial Pi}{\partial \theta_6}  & \frac{\partial Pi}{\partial \theta_7}\\
\frac{\partial Ya}{\partial \theta_4} & \frac{\partial Ya}{\partial \theta_5} & \frac{\partial Ya}{\partial \theta_6}  & \frac{\partial Ya}{\partial \theta_7}
\end{array}\right]$\small{}}

\subsection{Unit-consistency in the presence of various noise types}

In practice, the noise is often encountered in the form of truncated errors, rounding errors, scheme uncertainty, and external disturbance \cite{li2017novel}. Moreover, the noise can be constant or varying during the path planning \cite{li2017novel, guo2017new, li2018new}. To evaluate the noise suppression capabilities of the PPP and PRMP schemes presented in Table \ref{tab:surveyed-schemes-methodologies} when using the MX-GI, the noise formulation in \cite{guo2017new} is adopted. For each of these schemes, the desired path velocities  $\dot{D}_{d} \in \mathbb{R}^{m}$ or desired path accelerations $\ddot{D}_{d} \in \mathbb{R}^{m}$ were contaminated by a vector-form noise  $\delta(t) \in \mathbb{R}^{m}$ to create noise-polluted versions of these paths. In other words, without the existence of noise, $\delta(t)=0 \in \mathbb{R}^{m}$, and otherwise (i.e. in the presence of noise) $\delta(t)= [c_{1}, c_{2}, \dots, c_{m}]^{T}  \in \mathbb{R}^{m}$ for constant noise or $\delta(t)= [f_{1}(t), f_{2}(t), \dots, f_{m}(t)]^{T}  \in \mathbb{R}^{m}$ for when noise is time-varying \cite{li2017novel, guo2017new, li2018new}.  We also tested with $\delta(t)= [r_{1}, r_{2}, \dots, r_{m}]^{T}  \in \mathbb{R}^{m}$ with each $r_{i}$ being a seeded random number between $[0,1]$ for when noise is random (e.g., different throughout the path). In this case, the random generator is seeded as the units are varied.

\begin{figure}[!t]
\centering
\begin{subfigure}[b]{.15\textwidth}
  \centering
  % include first image
  \includegraphics[width=\textwidth,trim={0cm 0.7cm 0cm 1cm},clip]{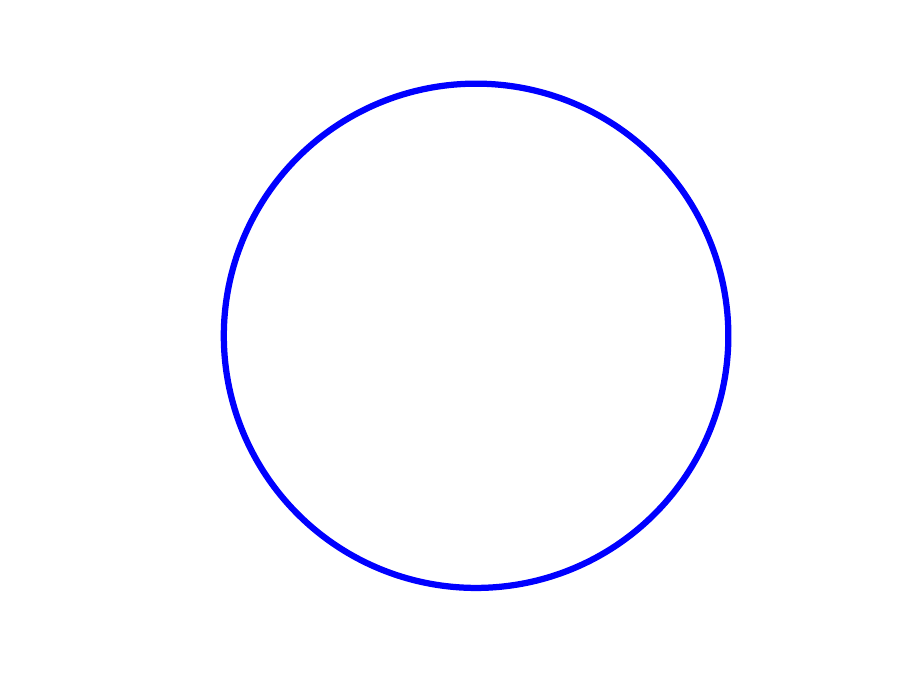} 
  %\caption{\scriptsize 3DoF - Path 1\\ \hspace{3cm} (Circle)}
  \caption[]{\scriptsize \begin{varwidth}[t]{\linewidth}3DoF - Path 1 \\ (Circle) \cite{guo2017new, li2018new}\end{varwidth}}
  \label{fig:3DoF-circle}
\end{subfigure}
\begin{subfigure}[b]{.15\textwidth}
  \centering
  % include second image
  \includegraphics[width=\textwidth,trim={0cm 0.7cm 0cm 1cm},clip]{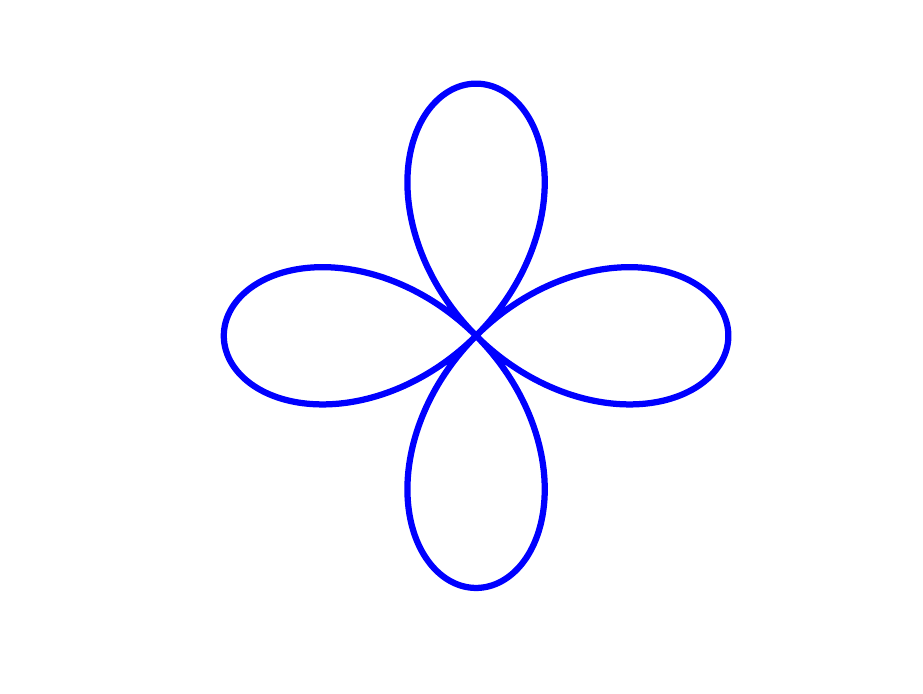} 
  %\caption{\scriptsize 3DoF - Path 2\\(Rhodonea)}
  \caption[]{\scriptsize \begin{varwidth}[t]{\linewidth}3DoF - Path 2 \\ (Rhodonea) \cite{guo2020repetitive}\end{varwidth}}
  \label{fig:3DoF-rhodonea}
\end{subfigure}
\begin{subfigure}[b]{.15\textwidth}
  \centering
  % include third image
  \includegraphics[width=\textwidth,trim={0cm 0.7cm 0cm 1cm},clip]{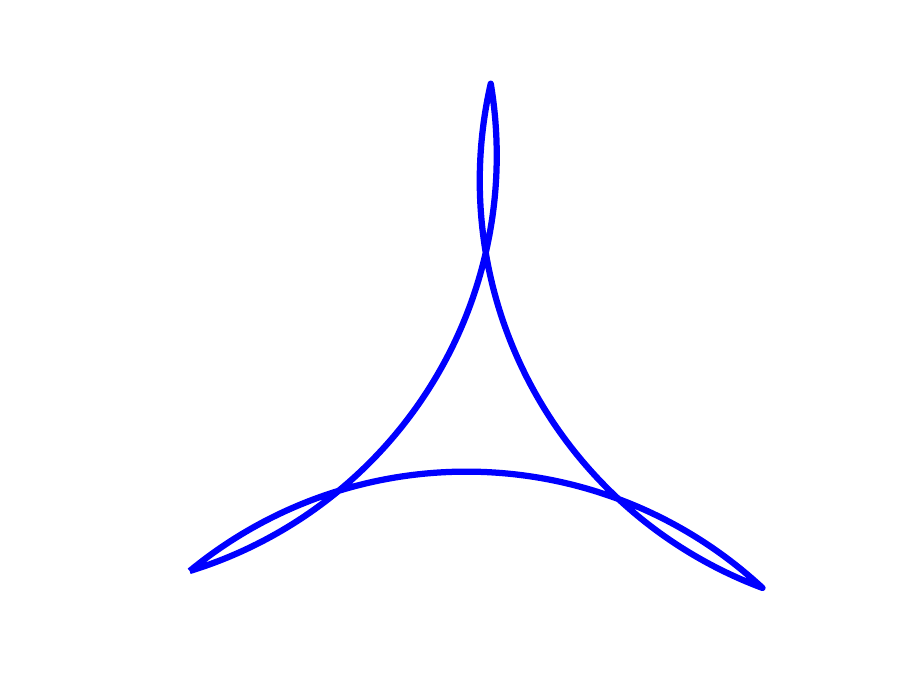} 
  %\caption{\scriptsize 3DoF - Path 3\\(Tricuspid)}
  \caption[]{\scriptsize \begin{varwidth}[t]{\linewidth}3DoF - Path 3 \\ (Tricuspid) \cite{wang2019feedback, guo2020repetitive}\end{varwidth}}
  \label{fig:3DoF-tricuspid}
\end{subfigure}

\begin{subfigure}[b]{.15\textwidth}
  \centering
  % include first image
  \includegraphics[width=\textwidth,trim={3cm 2.5cm 3cm 3cm},clip]{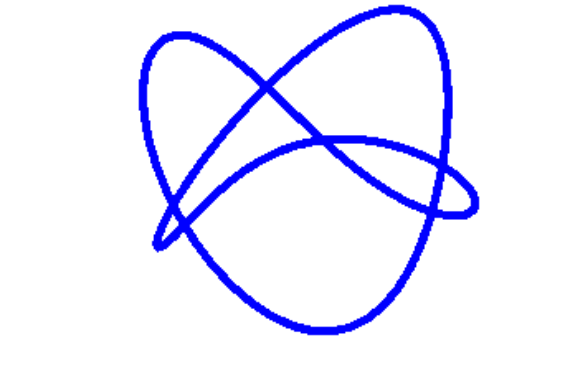} 
  %\caption{\scriptsize 7DoF - Path 1}
  \caption[]{\scriptsize \begin{varwidth}[t]{\linewidth}7DoF - Path 1 \\ (Interlaced Circle)\end{varwidth}}
  \label{fig:7DoF-entrelaced-circle}
\end{subfigure}
\begin{subfigure}[b]{.15\textwidth}
  \centering
  % include second image
  \includegraphics[width=\textwidth,trim={3cm 2cm 2cm 2cm},clip]{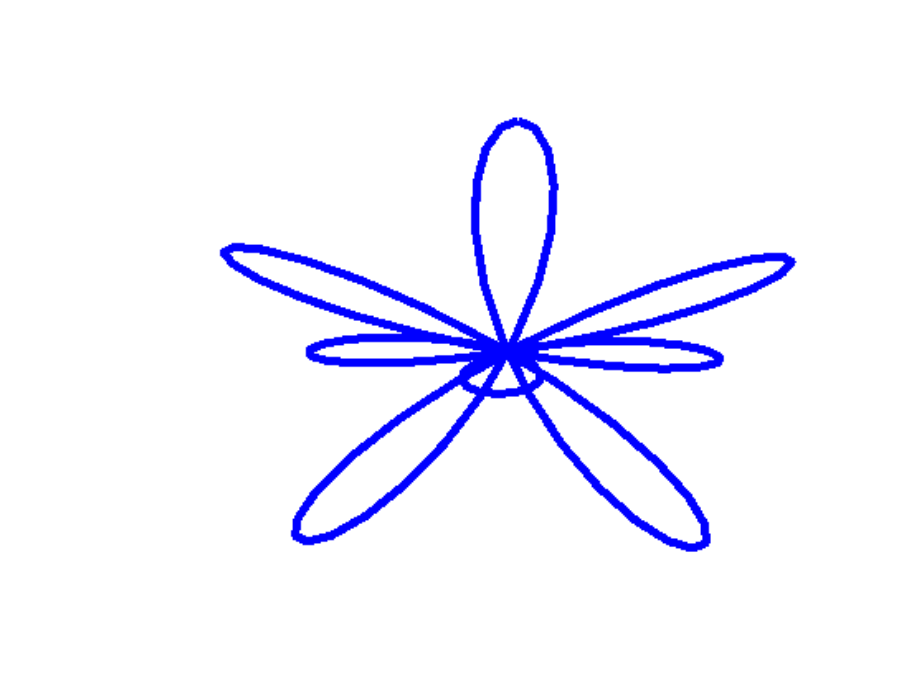} 
  %\caption{\scriptsize 7DoF - Path 2}
  \caption[]{\scriptsize \begin{varwidth}[t]{\linewidth}7DoF - Path 2 \\ (3D Rhodonea)\end{varwidth}}
  \label{fig:7DoF-entrelaced-rhodonea}
\end{subfigure}
\begin{subfigure}[b]{.15\textwidth}
  \centering
  % include third image
  \includegraphics[width=\textwidth,trim={5cm 2cm 2cm 3cm},clip]{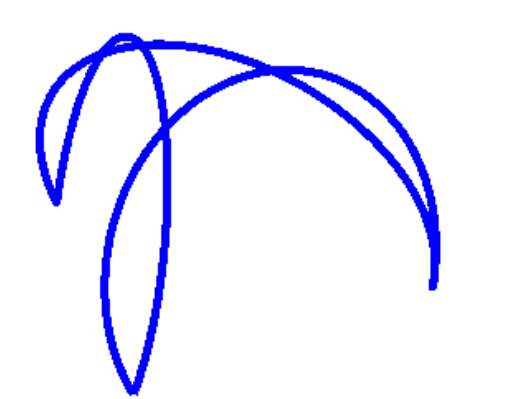} 
  %\caption{\scriptsize 7DoF - Path 3}
  \caption[]{\scriptsize \begin{varwidth}[t]{\linewidth}7DoF - Path 3 \\ (Bent Tricuspid)\end{varwidth}}
  \label{fig:7DoF-entrelaced-tricuspid}
\end{subfigure}
\caption{\footnotesize Investigated trajectories. For the 3DoF (2RP), these trajectories are 2-dimensional paths while for the 7DoF (2RP4R) they are designed to be 3-dimensional in the robot workspace.}
\label{fig:paths-investigated}
%\vspace{-4mm}
\end{figure}

\begin{figure}[!t]
\centering
\begin{subfigure}[b]{.24\textwidth}
  \centering
  % include first image
  \includegraphics[width=\textwidth,trim={0cm 0.9cm 0cm 0.8cm},clip]{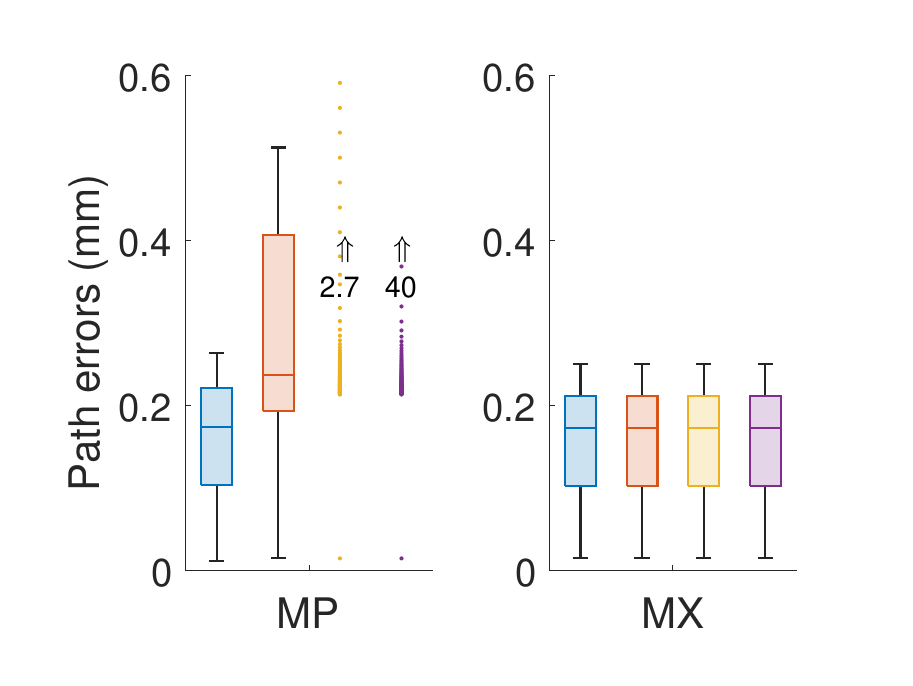} 
  \caption{\footnotesize WMVN / Zero noise / Path 1}
  \label{fig:3DoF-MVN-Circle}
\end{subfigure}
\begin{subfigure}[b]{.24\textwidth}
  \centering
  % include first image
  \includegraphics[width=\textwidth,trim={0cm 0.9cm 0cm 0.8cm},clip]{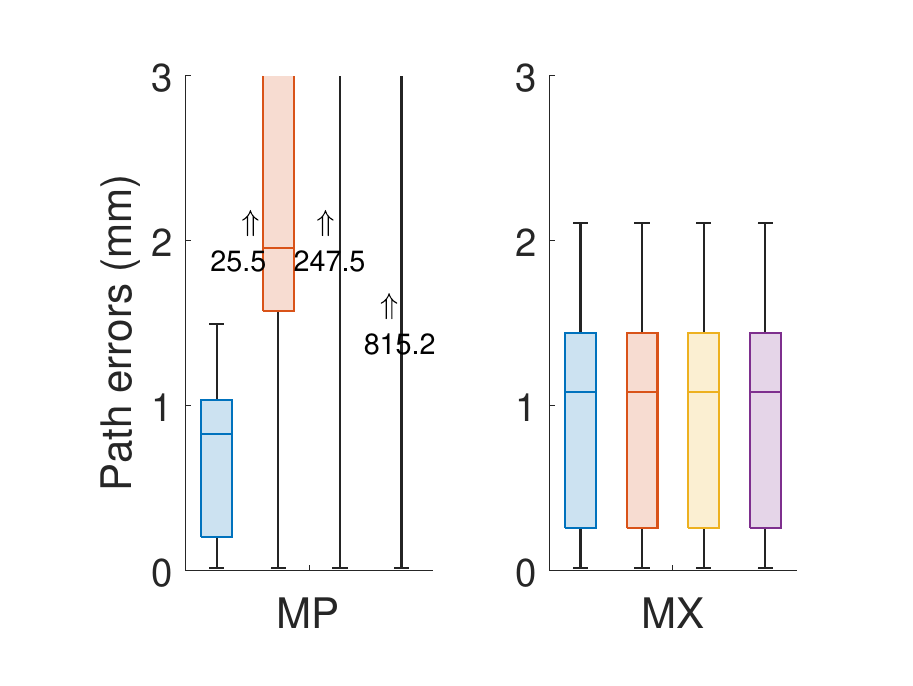} 
  \caption{\footnotesize MAN / Zero noise /  Path 1}
  \label{fig:3DoF-MAN-Circle}
\end{subfigure}
\begin{subfigure}[b]{.24\textwidth}
  \centering
  % include third image
  \includegraphics[width=\textwidth,trim={0cm 0.9cm 0cm 0cm},clip]{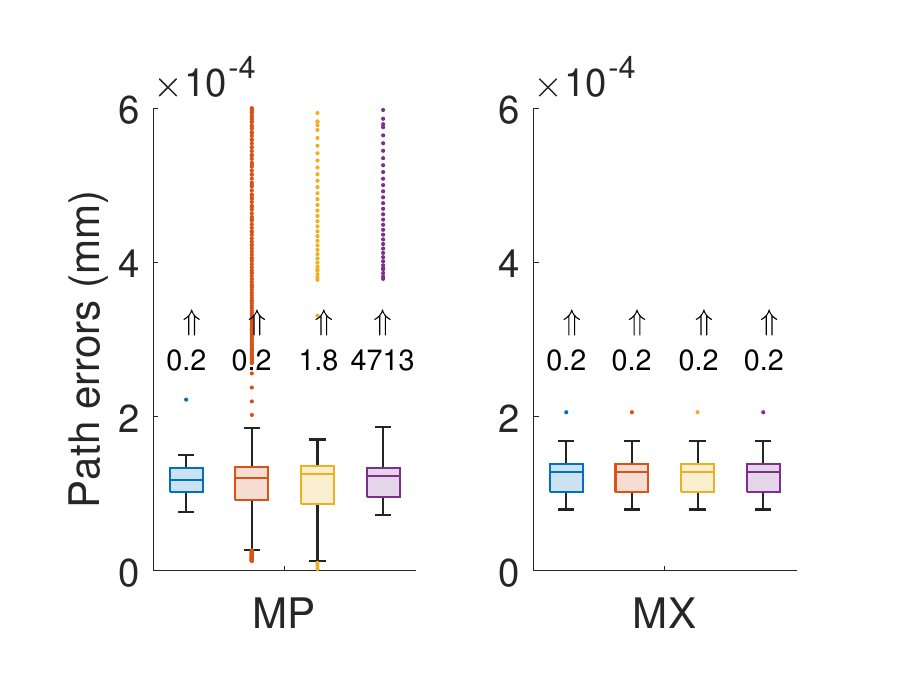} 
  \caption{\footnotesize PID-PPP / Zero noise /  Path 1}
  \label{fig:3DoF-PID-PPP-Circle}
\end{subfigure}
\begin{subfigure}[b]{.24\textwidth}
  \centering
  % include third image
  \includegraphics[width=\textwidth,trim={0cm 0.9cm 0cm 0cm},clip]{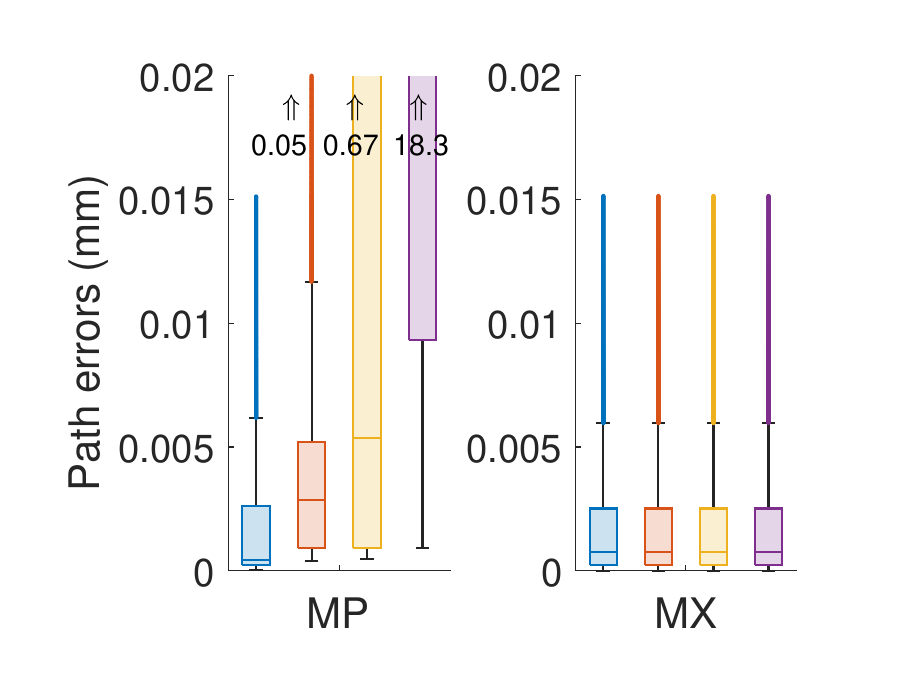} 
  \caption{\footnotesize FPBM / Zero noise /  Path 1}
  \label{fig:3DoF-FPBM-Circle}
\end{subfigure}
\begin{subfigure}[b]{.24\textwidth}
  \centering
  % include second image
  \includegraphics[width=\textwidth,trim={0cm 0.9cm 0cm 0cm},clip]{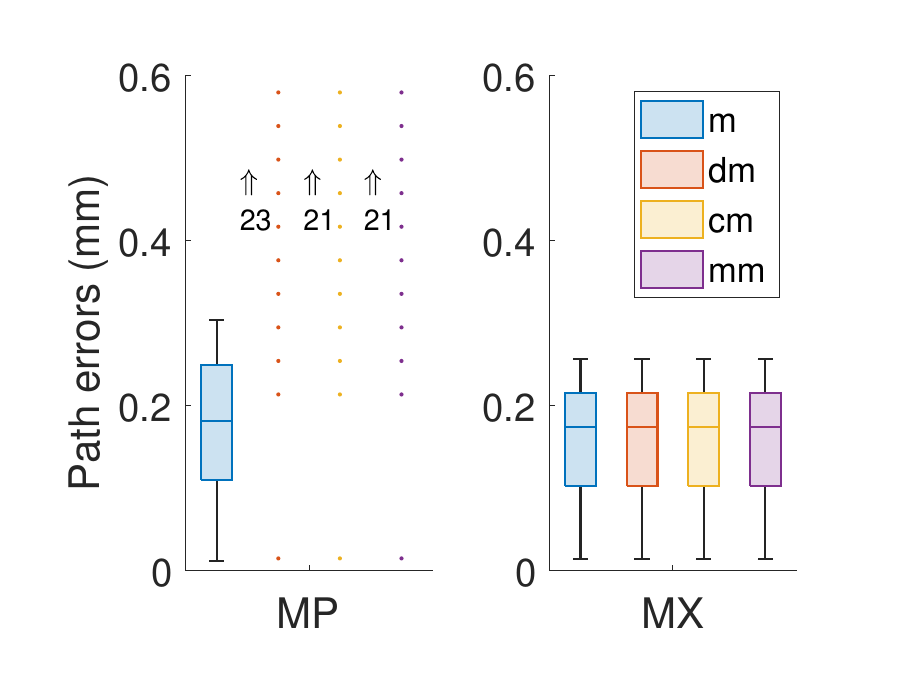} 
  \caption{\footnotesize V-SNS / Zero noise /  Path 1}
  \label{fig:3DoF-V-SNS-Circle}
\end{subfigure}
\begin{subfigure}[b]{.24\textwidth}
  \centering
  % include second image
  \includegraphics[width=\textwidth,trim={0cm 0.9cm 0cm 0cm},clip]{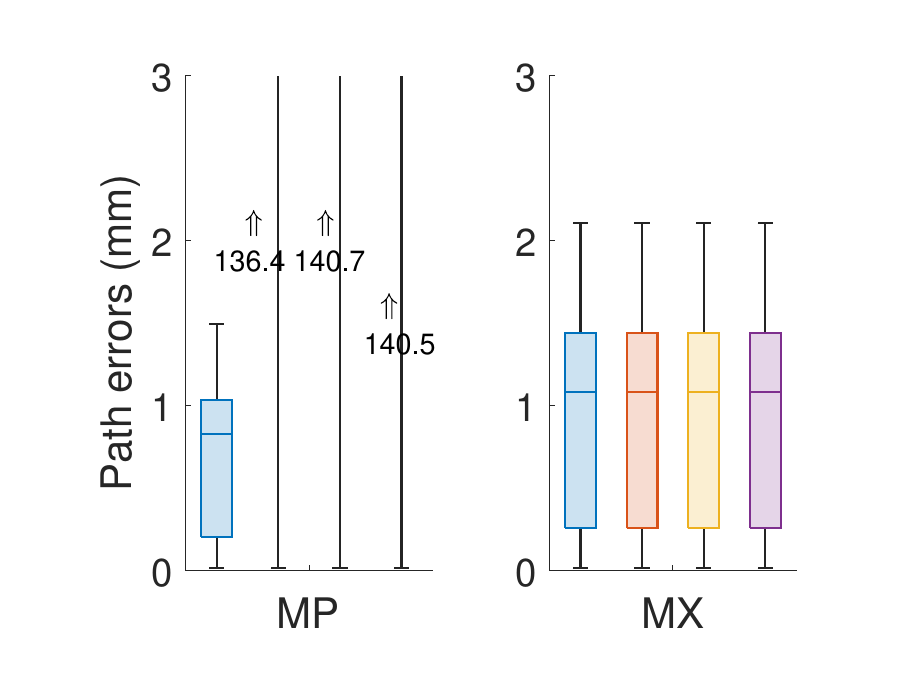} 
  \caption{\footnotesize A-SNS / Zero noise /  Path 1}
  \label{fig:3DoF-A-SNS-Circle}
\end{subfigure}
\caption{\footnotesize MP-GI versus MX-GI path errors while varying the units from $m$ to $mm$ for the WMVN, PID-PPP, V-SNS, MAN, FPBM and A-SNS schemes applied to the circle trajectory (path 1) of the 3DoF (2RP) manipulator. When a value exists next to a box plot, it indicates the maximum value (in $mm$) of the error distribution that has been zoomed in for better visualization.}
\label{fig:scheme-comparison-3DoF-Circle}
\vspace{-4mm}
\end{figure}

\begin{figure}[!t]
\centering
\begin{subfigure}[b]{.24\textwidth}
  \centering
  % include first image
  \includegraphics[width=\textwidth,trim={0cm 0.9cm 0cm 0.8cm},clip]{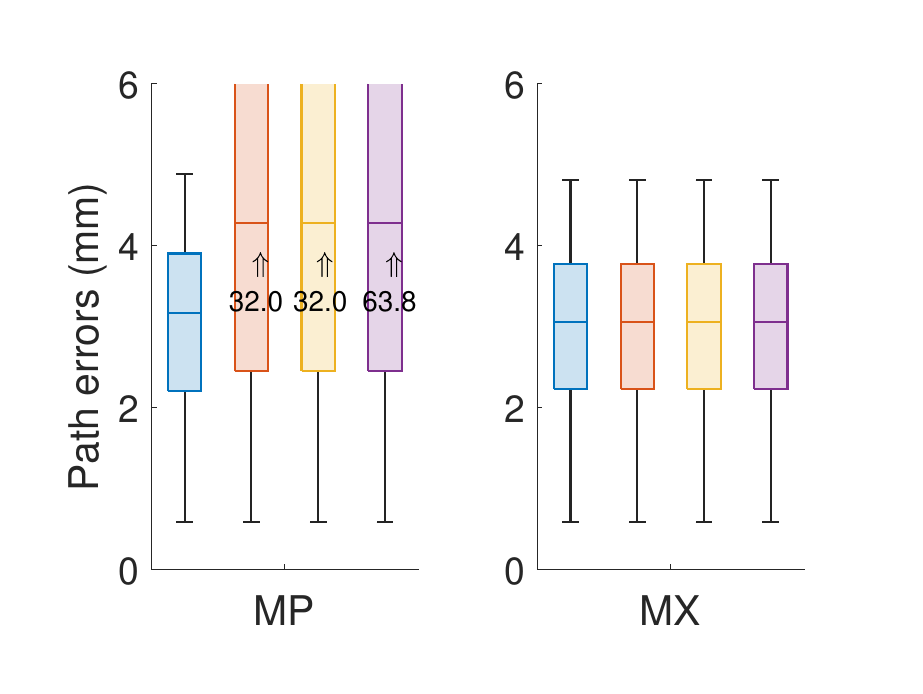} 
  \caption{\footnotesize WMVN / Zero noise /  Path 2}
  \label{fig:3DoF-MVN-Rhodonea}
\end{subfigure}
\begin{subfigure}[b]{.24\textwidth}
  \centering
  % include first image
  \includegraphics[width=\textwidth,trim={0cm 0.9cm 0cm 0.8cm},clip]{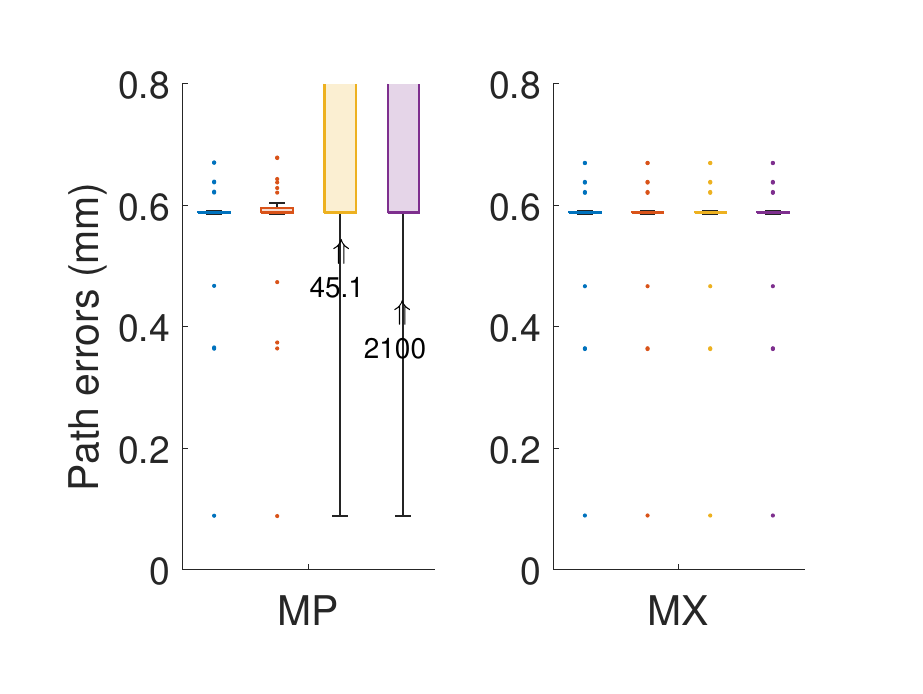} 
  \caption{\footnotesize MAN / Zero noise / Path 2}
  \label{fig:3DoF-MAN-Rhodonea}
\end{subfigure}
\begin{subfigure}[b]{.24\textwidth}
  \centering
  % include third image
  \includegraphics[width=\textwidth,trim={0cm 0.9cm 0cm 0cm},clip]{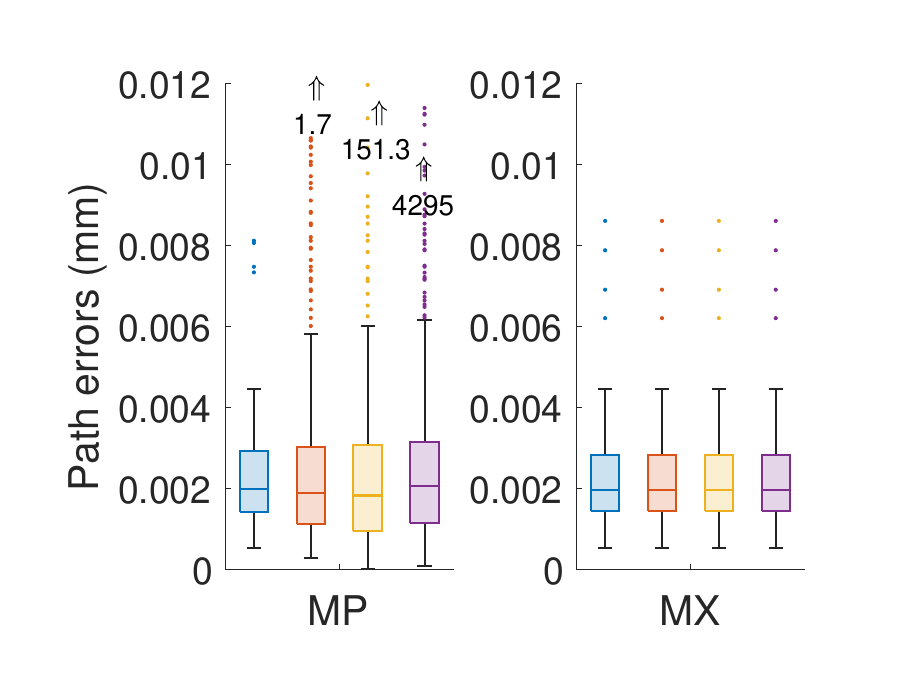} 
  \caption{\footnotesize PID-PPP / Zero noise / Path 2}
  \label{fig:3DoF-PID-PPP-Rhodonea}
\end{subfigure}
\begin{subfigure}[b]{.24\textwidth}
  \centering
  % include third image
  \includegraphics[width=\textwidth,trim={0cm 0.9cm 0cm 0cm},clip]{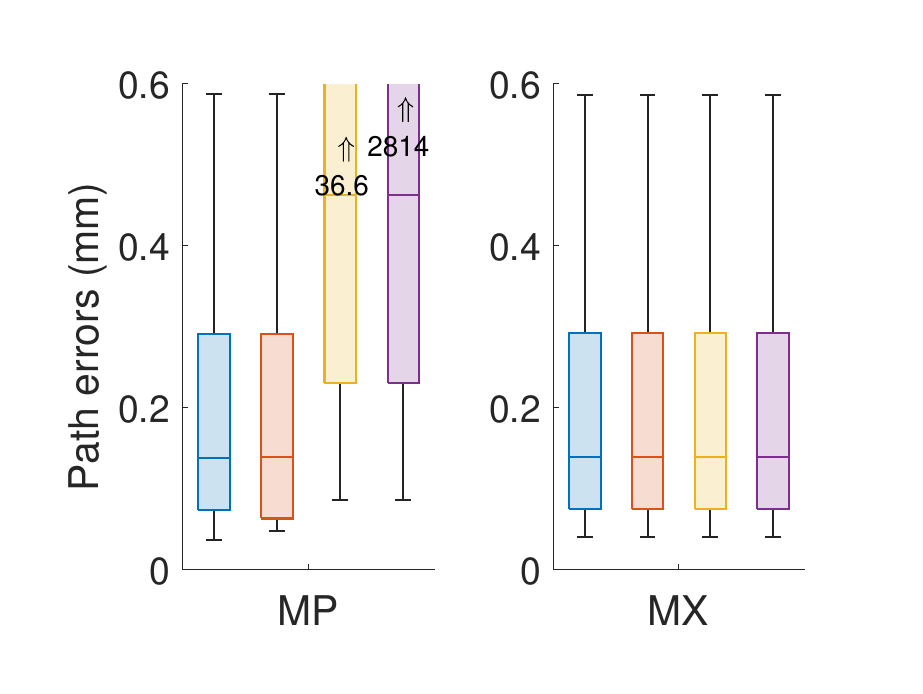} 
  \caption{\footnotesize FPBM / Zero noise / Path 2}
  \label{fig:3DoF-FPBM-Rhodonea}
\end{subfigure}
\begin{subfigure}[b]{.24\textwidth}
  \centering
  % include second image
  \includegraphics[width=\textwidth,trim={0cm 0.9cm 0cm 0cm},clip]{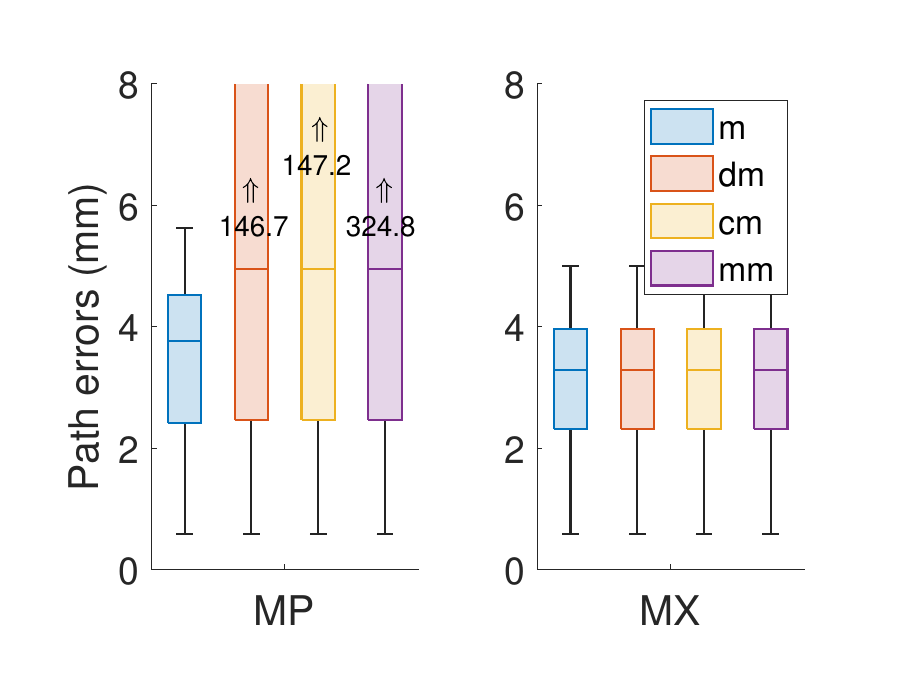} 
  \caption{\footnotesize V-SNS / Zero noise / Path 2}
  \label{fig:3DoF-V-SNS-Rhodonea}
\end{subfigure}
\begin{subfigure}[b]{.24\textwidth}
  \centering
  % include second image
  \includegraphics[width=\textwidth,trim={0cm 0.9cm 0cm 0cm},clip]{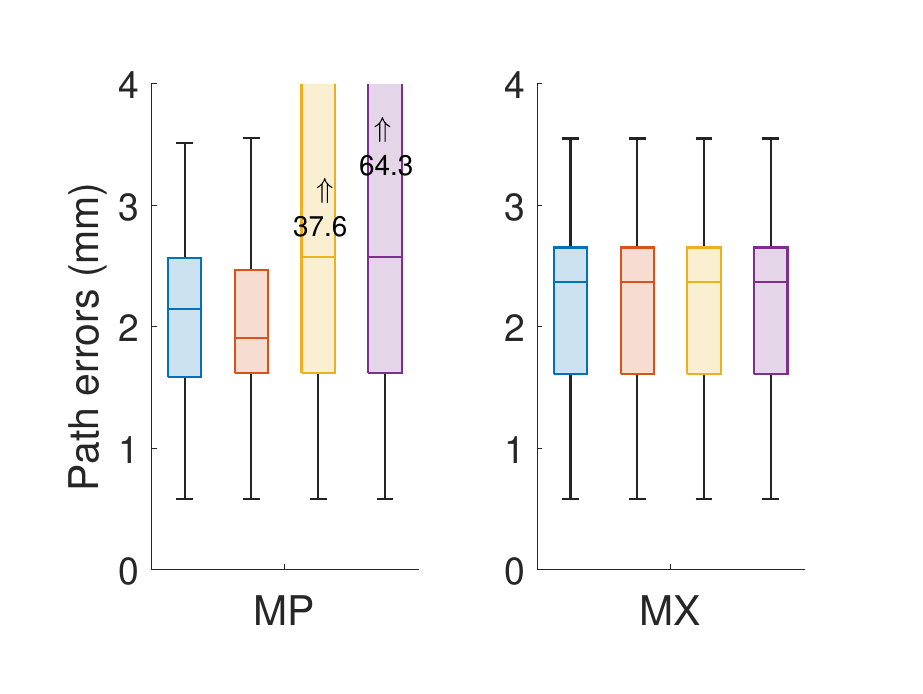} 
  \caption{\footnotesize A-SNS / Zero noise / Path 2}
  \label{fig:3DoF-A-SNS-Rhodonea}
\end{subfigure}
\caption{\footnotesize MP-GI versus MX-GI path errors while varying the units from $m$ to $mm$ for the WMVN, PID-PPP, V-SNS, MAN, FPBM and A-SNS schemes applied to the rhodonea trajectory (path 2) of the 3DoF (2RP) manipulator. When a value exists next to a box plot, it indicates the maximum value (in $mm$) of the error distribution that has been zoomed in for better visualization.}
\label{fig:scheme-comparison-3DoF-Rhodonea}
\vspace{-4mm}
\end{figure}

\begin{figure}[!t]
\centering
\begin{subfigure}[b]{.24\textwidth}
  \centering
  % include  image
  \includegraphics[width=\textwidth,trim={0cm 0.9cm 0cm 0.8cm},clip]{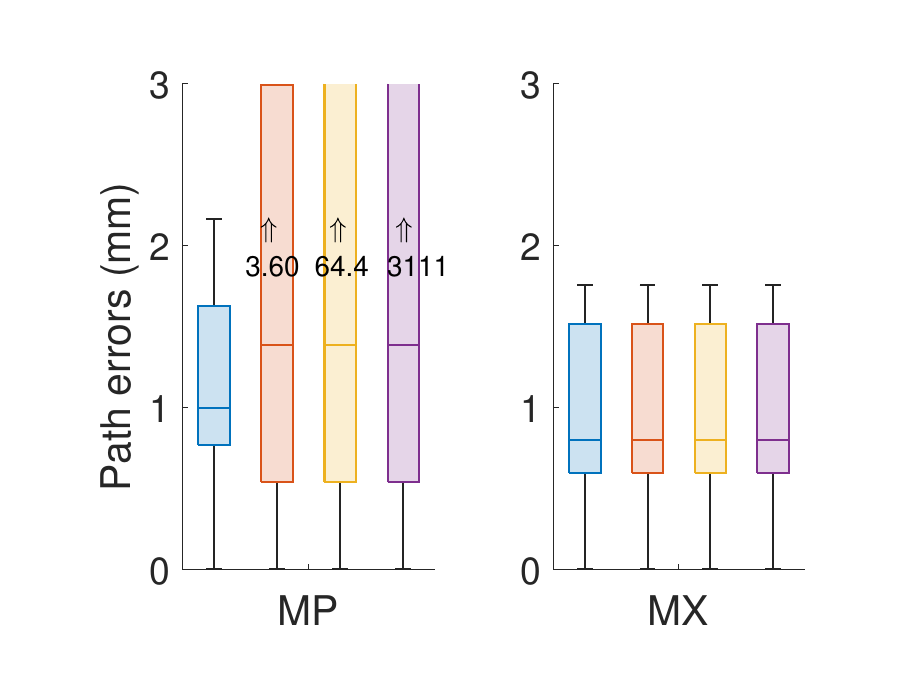} 
  \caption{\footnotesize WMVN / Zero noise / Path 3}
  \label{fig:3DoF-MVN-Tricuspid}
\end{subfigure}
\begin{subfigure}[b]{.24\textwidth}
  \centering
  % include  image
  \includegraphics[width=\textwidth,trim={0cm 0.9cm 0cm 0.8cm},clip]{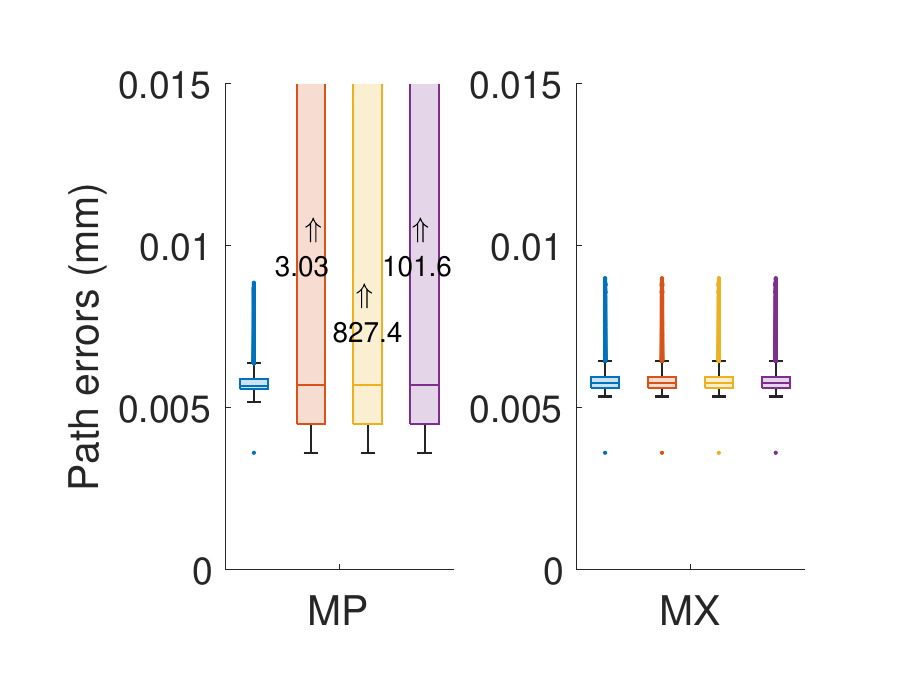} 
  \caption{\footnotesize MAN / Zero noise / Path 3}
  \label{fig:3DoF-MAN-Tricuspid}
\end{subfigure}
\begin{subfigure}[b]{.24\textwidth}
  \centering
  % include  image
  \includegraphics[width=\textwidth,trim={0cm 0.9cm 0cm 0cm},clip]{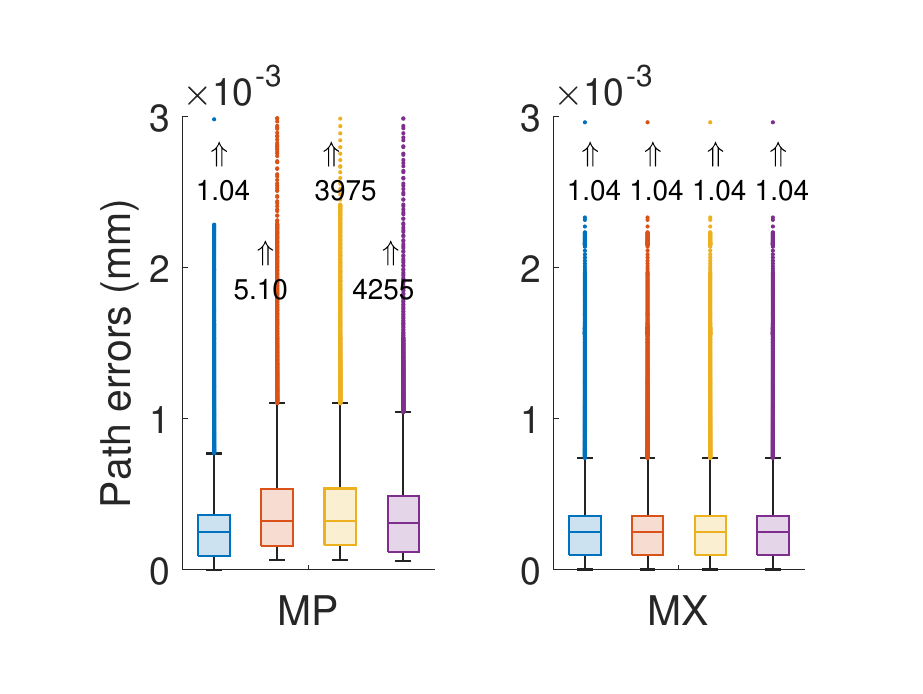} 
  \caption{\footnotesize PID-PPP / Zero noise / Path 3}
  \label{fig:3DoF-PID-PPP-Tricuspid}
\end{subfigure}
\begin{subfigure}[b]{.24\textwidth}
  \centering
  % include  image
  \includegraphics[width=\textwidth,trim={0cm 0.9cm 0cm 0cm},clip]{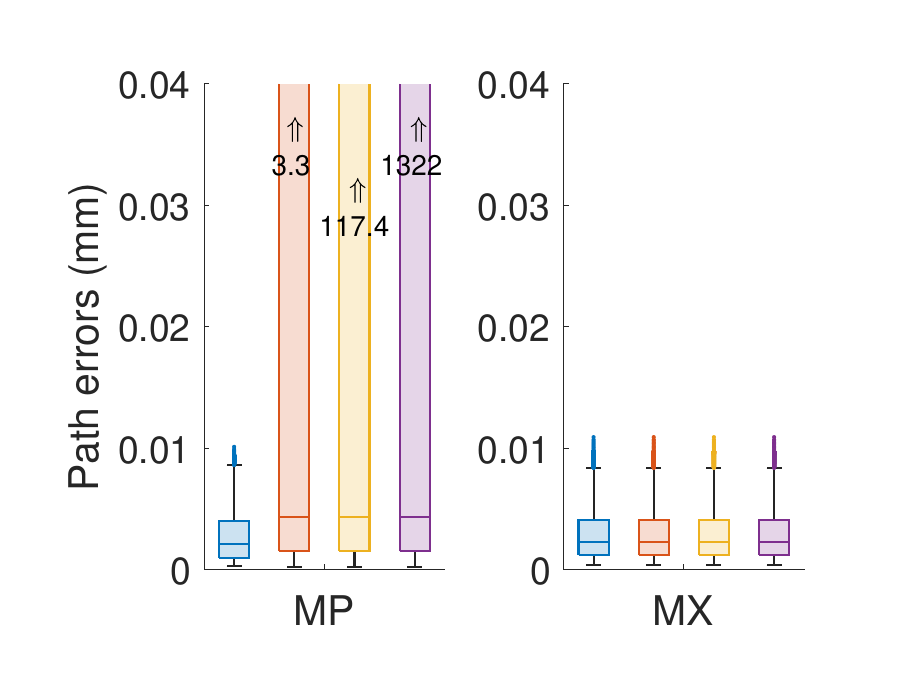} 
  \caption{\footnotesize FPBM / Zero noise / Path 3}
  \label{fig:3DoF-FPBM-Tricuspid}
\end{subfigure}
\begin{subfigure}[b]{.24\textwidth}
  \centering
  % include  image
  \includegraphics[width=\textwidth,trim={0cm 0.9cm 0cm 0cm},clip]{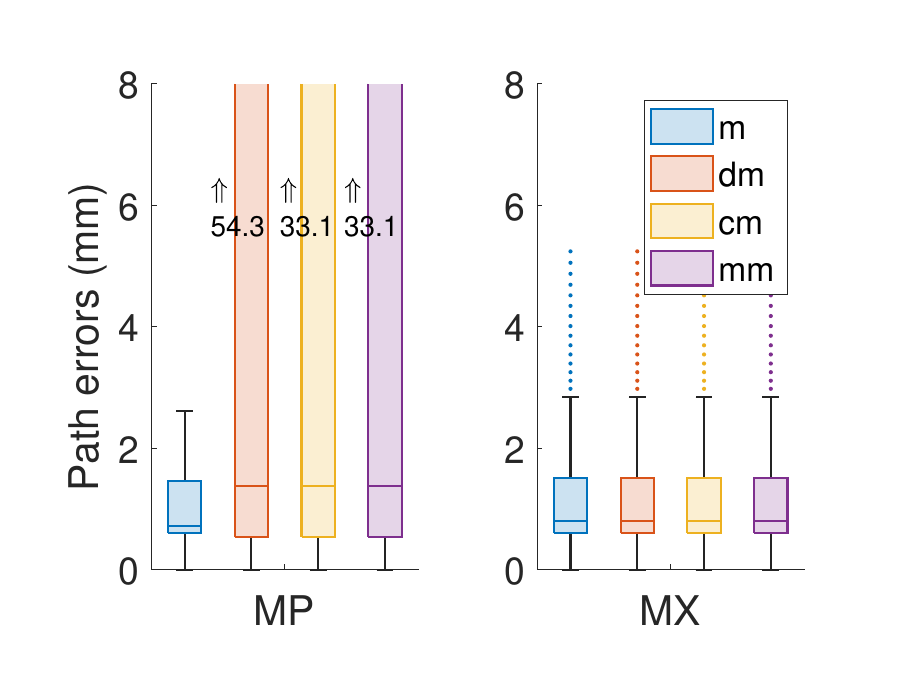} 
  \caption{\footnotesize V-SNS / Zero noise / Path 3}
  \label{fig:3DoF-V-SNS-Tricuspid}
\end{subfigure}
\begin{subfigure}[b]{.24\textwidth}
  \centering
  % include  image
  \includegraphics[width=\textwidth,trim={0cm 0.9cm 0cm 0cm},clip]{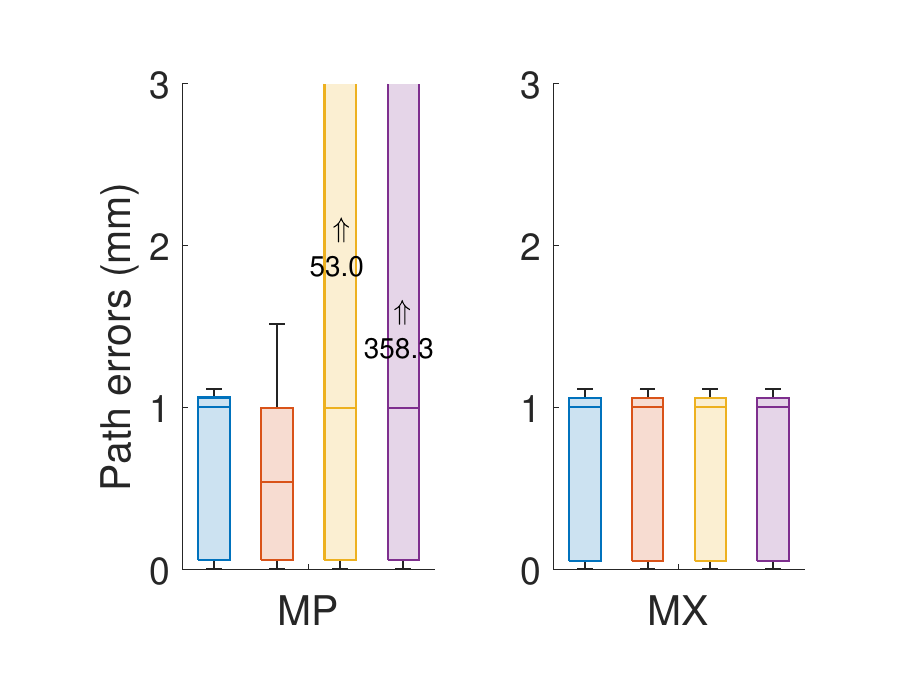} 
  \caption{\footnotesize A-SNS / Zero noise / Path 3}
  \label{fig:3DoF-A-SNS-Tricuspid}
\end{subfigure}
\caption{\footnotesize MP-GI versus MX-GI path errors while varying the units from $m$ to $mm$ for the WMVN, PID-PPP, V-SNS, MAN, FPBM, and A-SNS schemes applied to the tricuspid trajectory (path 3) of the 3DoF (2RP) manipulator. When a value exists next to a box plot, it indicates the maximum value (in $mm$) of the error distribution that has been zoomed in for better visualization.}
\label{fig:scheme-comparison-3DoF-Tricuspid}
\vspace{-6mm}
\end{figure}

\begin{table*}[!ht] 
\notsotinytwo
\caption{\footnotesize Average errors (in $mm$) between the desired and estimated paths using the investigated PPP schemes for the 3DoF (2RP) planar manipulator with zero, constant, time-varying, and random noises. The $dm$ results have been omitted for space consideration.}\label{tab:schemes-units-3DoF-zero}
    \centering
    \begin{threeparttable}
         \begin{tabular}{|p{1.2mm}|p{7.1mm}|c|c|c|c|c|c|c|c|c|c|c|c|c|c|c|c|c|c|} 
         %\begin{tblr}{
         %       colspec = {|c|c|c|c|},
                %row{4} = {gray9},
                %row{12} = {gray9},
                %row{14} = {gray9},
                %row{22} = {gray9},
                %row{31} = {gray9},
                %row{32} = {gray9},
                %column{3} = {teal7},
                %cell{2}{3} = {yellow7},
        %      }
         \hline
         %\multirow{3}{3.5em}{Schemes} & \multicolumn{6}{c|}{Zero Noise} \\
         %\cline{2-7} 
         1 & \multirow{3}{3.5em}{Schemes} & \multicolumn{6}{c|}{3DoF - Path 1 (Circle)} & \multicolumn{6}{c|}{3DoF - Path 2 (Rhodonea)} & \multicolumn{6}{c|}{3DoF - Path 3 (Tricuspid)} \\ 
         \cline{3-20}         
         2 & & \multicolumn{3}{c|}{Original MP-GI} & \multicolumn{3}{c|}{Proposed MX-GI} & \multicolumn{3}{c|}{Original MP-GI} & \multicolumn{3}{c|}{Proposed MX-GI} & \multicolumn{3}{c|}{Original MP-GI} & \multicolumn{3}{c|}{Proposed MX-GI}\\  
         \cline{3-20} 
         3 & & $m$ & $cm$ & $mm$ & $m$ & $cm$ & $mm$ & $m$ & $cm$ & $mm$ & $m$ & $cm$ & $mm$ & $m$ & $cm$ & $mm$ & $m$ & $cm$ & $mm$\\  
         \hline\hline
         4 & \multicolumn{19}{c|}{\textbf{Errors for Zero Noise in $mm$}}\\  
         \hline
         5 & PID-PPP & 0.00 & 0.01 & 913.4 & \cellcolor{gray9}0.00 & \cellcolor{gray9}0.00 & \cellcolor{gray9}0.00 & 0.00 & 0.23 & 21.1 & \cellcolor{gray9}0.00 & \cellcolor{gray9}0.00 & \cellcolor{gray9}0.00  & 0.00 & 11.9 & 17.4 & \cellcolor{gray9}0.00 & \cellcolor{gray9}0.00 & \cellcolor{gray9}0.00 \\
         \hline
         6 & WMVN & 0.16 & 2.54 & 38.41 & \cellcolor{gray9}0.16 & \cellcolor{gray9}0.16 & \cellcolor{gray9}0.16 &  3.07 & 14.3 & 27.7 & \cellcolor{gray9}2.99 & \cellcolor{gray9}2.99 & \cellcolor{gray9}2.99 & 1.14 & 22.4 & 967.8 & \cellcolor{gray9}0.98 & \cellcolor{gray9}0.98 & \cellcolor{gray9}0.98\\
         \hline
         7 & V-SNS & 0.18 & 16.08 & 16.42 & \cellcolor{gray9}0.16 & \cellcolor{gray9}0.16 & \cellcolor{gray9}0.16 & 3.55 & 64.9 & 79.4 & \cellcolor{gray9}3.17 & \cellcolor{gray9}3.17 & \cellcolor{gray9}3.17 & 0.94 & 8.94 & 8.95 & \cellcolor{gray9}0.99 & \cellcolor{gray9}0.99 & \cellcolor{gray9}0.99 \\ 
         \hline
         8 & MAN & 0.68 & 122.7 & 402.7 & \cellcolor{gray9}0.96 & \cellcolor{gray9}0.96 & \cellcolor{gray9}0.96 & 0.58 & 18.51 & 860.6 & \cellcolor{gray9}0.58 & \cellcolor{gray9}0.58 & \cellcolor{gray9}0.58 & 0.01 & 282.7 & 34.67 & \cellcolor{gray9}0.01 & \cellcolor{gray9}0.01 & \cellcolor{gray9}0.01 \\
         \hline
         9 & A-SNS & 0.68 & 68.58 & 68.45 & \cellcolor{gray9}0.96 & \cellcolor{gray9}0.96 & \cellcolor{gray9}0.96 & 2.05 & 16.3 & 25.8 & \cellcolor{gray9}2.15 & \cellcolor{gray9}2.15 & \cellcolor{gray9}2.15 & 0.71 & 18.25 & 116.84 & \cellcolor{gray9}0.71 & \cellcolor{gray9}0.71 & \cellcolor{gray9}0.71\\
         \hline
         10 & FPBM & 0.00 & 0.06 & 1.59 & \cellcolor{gray9}0.00 & \cellcolor{gray9}0.00 & \cellcolor{gray9}0.00 & 0.19 & 10.5 & 690.1 & \cellcolor{gray9}0.19 & \cellcolor{gray9}0.19 & \cellcolor{gray9}0.19 & 0.00 & 14.99 & 168.28 & \cellcolor{gray9}0.00 & \cellcolor{gray9}0.00 & \cellcolor{gray9}0.00 \\
         \hline
         \hline
         11 & \multicolumn{19}{c|}{\textbf{Errors for Constant Noise in $mm$}}\\ 
         \hline
         12 & PID-PPP & 0.58 & 0.58 & 28.30 & \cellcolor{gray9}0.58 & \cellcolor{gray9}0.58 & \cellcolor{gray9}0.58 & 0.55 & 0.73 & 23.94 & \cellcolor{gray9}0.55 & \cellcolor{gray9}0.55 & \cellcolor{gray9}0.55 & 0.52 & 18.9 & 8.22 & \cellcolor{gray9}0.52 & \cellcolor{gray9}0.52 & \cellcolor{gray9}0.52 \\
         \hline
         13 & WMVN & 1.8E3 & 1.5E3 & 434.7 & \cellcolor{red9}1.8E3 & \cellcolor{red9}1.8E3 & \cellcolor{red9}1.8E3 & 1.7E3 & 1.0E3 & 585.3 & \cellcolor{red9}1.7E3 & \cellcolor{red9}1.7E3 & \cellcolor{red9}1.7E3 & 2.9E3 & 1.4E3& 892.8 & \cellcolor{red9}2.9E3 & \cellcolor{red9}2.9E3 & \cellcolor{red9}2.9E3 \\
         \hline
         14 & V-SNS & 1.2E3 & 1.1E3 & 1.1E3 & \cellcolor{red9}1.2E3 & \cellcolor{red9}1.2E3 & \cellcolor{red9}1.2E3  & 1.1E3 & 1.2E3 & 1.2E3 & \cellcolor{red9}1.4E3 & \cellcolor{red9}1.4E3 & \cellcolor{red9}1.4E3 & 2.9E3 & 2.8E3 & 2.8E3 & \cellcolor{red9}2.9E3 & \cellcolor{red9}2.9E3 & \cellcolor{red9}2.9E3\\ 
         \hline
         15 & MAN & 9.7E3 & 9.4E3 & 2.4E3 & \cellcolor{red9}9.7E3 & \cellcolor{red9}9.7E3 & \cellcolor{red9}9.7E3 & 1.8E3 & 710.6 & 1.3E3 & \cellcolor{red9}1.8E3 & \cellcolor{red9}1.8E3 & \cellcolor{red9}1.8E3 & 964.4 & 2.4E3 & 3.1E19 & \cellcolor{red9}964.4 & \cellcolor{red9}964.4 & \cellcolor{red9}964.4 \\
         \hline
         16 & A-SNS & 9.7E3 & 7.7E3 & 7.6E3 & \cellcolor{red9}9.7E3 & \cellcolor{red9}9.7E3 & \cellcolor{red9}9.7E3  & 1.8E3 & 595.1 & 654.1 & \cellcolor{red9}1.8E3 & \cellcolor{red9}1.8E3 & \cellcolor{red9}1.8E3 & 2.1E3 & 847.9 & 787.7 & \cellcolor{red9}2.9E3 & \cellcolor{red9}2.9E3 & \cellcolor{red9}2.9E3 \\
         \hline
         17 & FPBM & 0.53 & 0.52 & 5.63 & \cellcolor{gray9}0.53 & \cellcolor{gray9}0.53 & \cellcolor{gray9}0.53 & 0.16 & 11.17 & 29.16 & \cellcolor{gray9}0.16 & \cellcolor{gray9}0.16 & \cellcolor{gray9}0.16 & 0.25 & 6.6 & 8.9 & \cellcolor{gray9}0.25 & \cellcolor{gray9}0.25 & \cellcolor{gray9}0.25 \\
         \hline
         \hline
         18 & \multicolumn{19}{c|}{\textbf{Errors for Time-varying Noise in $mm$}}\\ 
         \hline
         19 & PID-PPP & 0.30 & 0.30 & 752.21 & \cellcolor{gray9}0.30 & \cellcolor{gray9}0.30 & \cellcolor{gray9}0.30  & 0.29 & 0.46 & 144.4 & \cellcolor{gray9}0.29 & \cellcolor{gray9}0.29 & \cellcolor{gray9}0.29 & 0.22 & 12.47 & 14.92 &\cellcolor{gray9}0.22 & \cellcolor{gray9}0.22 & \cellcolor{gray9}0.22 \\
         \hline
         20 & WMVN & 190.9 & 163.7 & 204.3 & \cellcolor{red9}190.9 & \cellcolor{red9}190.9 & \cellcolor{red9}190.9  & 189.7 & 186.8 & 210.7 & \cellcolor{red9}189.7 & \cellcolor{red9}189.7 & \cellcolor{red9}189.7 & 185.1 & 185.4 & 573.77 & \cellcolor{red9}185.3 & \cellcolor{red9}185.3 & \cellcolor{red9}185.3 \\
         \hline
         21 & V-SNS & 312.9 & 307.3 & 305.5 & \cellcolor{red9}312.9 &  \cellcolor{red9}312.9 & \cellcolor{red9}312.9  & 193.7 & 248.4 & 248.8 & \cellcolor{red9}189.6 & \cellcolor{red9}189.6 & \cellcolor{red9}189.6 & 185.3 & 183.0 & 183.1 & \cellcolor{red9}185.3 & \cellcolor{red9}185.3 & \cellcolor{red9}185.3 \\ 
         \hline
         22 & MAN & 757.7 & 553.3 & 2E22 & \cellcolor{red9}757.8 & \cellcolor{red9}757.8 & \cellcolor{red9}757.8 &  190.7 & 193.2 & 209.5 & \cellcolor{red9}190.7 & \cellcolor{red9}190.7 & \cellcolor{red9}190.7 & 992.9 & 1.9E21 & 1.2E20 & \cellcolor{red9}89.38 & \cellcolor{red9}89.38 & \cellcolor{red9}89.38 \\
         \hline
         23 & A-SNS & 757.7 & 553.3 & 1.7E3 & \cellcolor{red9}757.7 & \cellcolor{red9}757.7 & \cellcolor{red9}757.7  & 190.2 & 196.5 & 201.8 & \cellcolor{red9}190.2 & \cellcolor{red9}190.2 & \cellcolor{red9}190.2 & 185.3 & 181.1 & 501.9 & \cellcolor{red9}185.3 & \cellcolor{red9}185.3 & \cellcolor{red9}185.3 \\
         \hline
         24 & FPBM & 0.13 & 0.36 & 3.72 & \cellcolor{gray9}0.13 & \cellcolor{gray9}0.13 & \cellcolor{gray9}0.13  & 0.19 & 11.8 & 305.8 & \cellcolor{gray9}0.19 & \cellcolor{gray9}0.19 & \cellcolor{gray9}0.19 & 0.21 & 6.34 & 8.01 & \cellcolor{gray9}0.23 & \cellcolor{gray9}0.23 & \cellcolor{gray9}0.23 \\
         \hline
         25 & \multicolumn{19}{c|}{\textbf{Errors for Random Noise in $mm$}}\\ 
         \hline
         26 & PID-PPP & 0.87 & 0.87 & 190.73 & \cellcolor{gray9}0.87 & \cellcolor{gray9}0.87& \cellcolor{gray9}0.87  & 0.76 & 1.02 & 90.9 & \cellcolor{gray9}0.76 & \cellcolor{gray9}0.76 & \cellcolor{gray9}0.76 & 0.71 & 9.26 & 7.34 & \cellcolor{gray9}0.71 & \cellcolor{gray9}0.71 & \cellcolor{gray9}0.71  \\
         \hline
         27 & WMVN & 3.5E3 & 2.5E3 & 256.1 & \cellcolor{red9}3.5E3 & \cellcolor{red9}3.5E3 & \cellcolor{red9}3.5E3 & 2.2E3 & 1.2E3 & 531.1 & \cellcolor{red9}2.2E3 & \cellcolor{red9}2.2E3 & \cellcolor{red9}2.2E3 &  3.6E3 & 1.5E3 & 927.9 & \cellcolor{red9} 3.6E3 &  \cellcolor{red9}3.6E3 & \cellcolor{red9}3.6E3\\
         \hline
         28 & V-SNS & 1.8E3 & 1.7E3 & 1.7E3 & \cellcolor{red9}1.8E3 &  \cellcolor{red9}1.8E3 & \cellcolor{red9}1.8E3 & 1.3E3 & 1.0E3 & 1.1E3 & \cellcolor{red9}1.6E3 & \cellcolor{red9}1.6E3 & \cellcolor{red9}1.6E3 &  3.4E3 & 3.2E3 & 3.5E3 & \cellcolor{red9}3.5E3 & \cellcolor{red9}3.5E3 & \cellcolor{red9}3.5E3 \\ 
         \hline
         29 & MAN & 1.2E4 & 1.1E4 & 2E17 & \cellcolor{red9}1.2E4 & \cellcolor{red9}1.2E4 & \cellcolor{red9}1.2E4  & 2.1E3 & 744.6 & 1.3E3 & \cellcolor{red9}2.1E3 & \cellcolor{red9}2.1E3 & \cellcolor{red9}2.1E3 & 1.2E3 & 3.1E3 & 1.3E21 & \cellcolor{red9}1.2E3 & \cellcolor{red9}1.2E3 & \cellcolor{red9}1.2E3 \\
         \hline
         30 & A-SNS & 1.2E4 & 9.1E3 & 9.1E3 &  \cellcolor{red9}1.2E4 & \cellcolor{red9}1.2E4 & \cellcolor{red9}1.2E4  & 2.1E3 & 606.0 & 477.3 & \cellcolor{red9}2.1E3 & \cellcolor{red9}2.1E3 & \cellcolor{red9}2.1E3 & 3.5E3 & 748.8 & 1.5E3 & \cellcolor{red9}3.5E3 & \cellcolor{red9}3.5E3 & \cellcolor{red9}3.5E3 \\
         \hline
         31 & FPBM & 0.64 & 7.13 & 22.06 & \cellcolor{gray9}0.64 & \cellcolor{gray9}0.64 & \cellcolor{gray9}0.64  & 0.16 & 11.85 & 521.8 & \cellcolor{gray9}0.16 & \cellcolor{gray9}0.16 & \cellcolor{gray9}0.16 & 0.41 & 66.96 & 11.15 & \cellcolor{gray9}0.26 & \cellcolor{gray9}0.26 & \cellcolor{gray9}0.26\\
         \hline
        %\end{tblr}
        \end{tabular}
        % Note under the table
        \begin{tablenotes}
        \small
        \item  
        \end{tablenotes}
    \end{threeparttable}
\vspace{-8mm}
\end{table*}

\begin{figure*}[!t]
\centering
\begin{subfigure}{.245\textwidth}
  \centering
  % include  image
  \includegraphics[width=\textwidth,trim={0cm 0.45cm 1cm 0.5cm},clip]{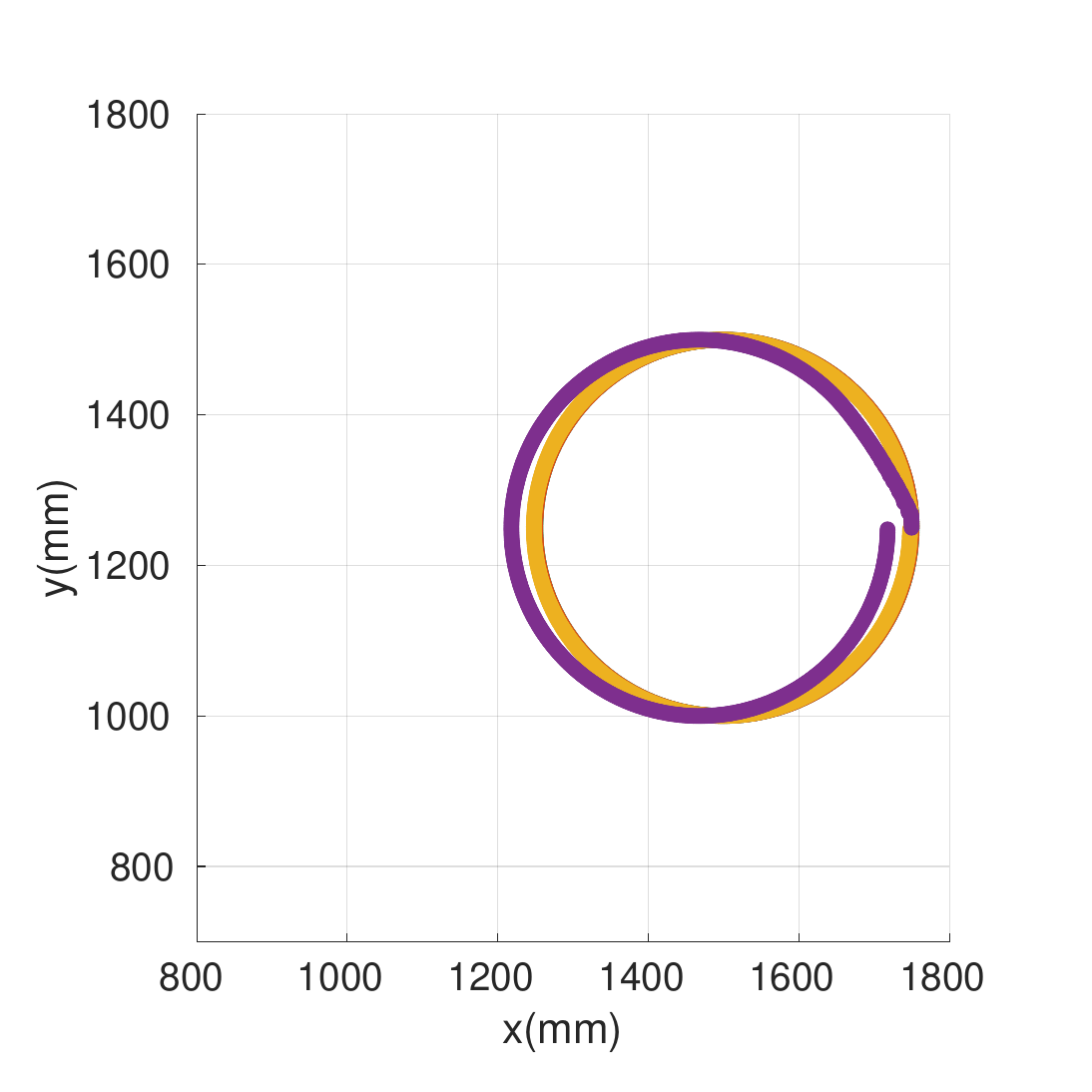}  
  \caption{\scriptsize WMVN / MP / Path 1 / Zero noise}
  \label{fig:3dof-MP-WMVN-Circle}
\end{subfigure}
%\hspace{5mm}
\begin{subfigure}{.245\textwidth}
  \centering
  % include  image
  \includegraphics[width=\textwidth,trim={0cm 0.45cm 1cm 0.5cm},clip]{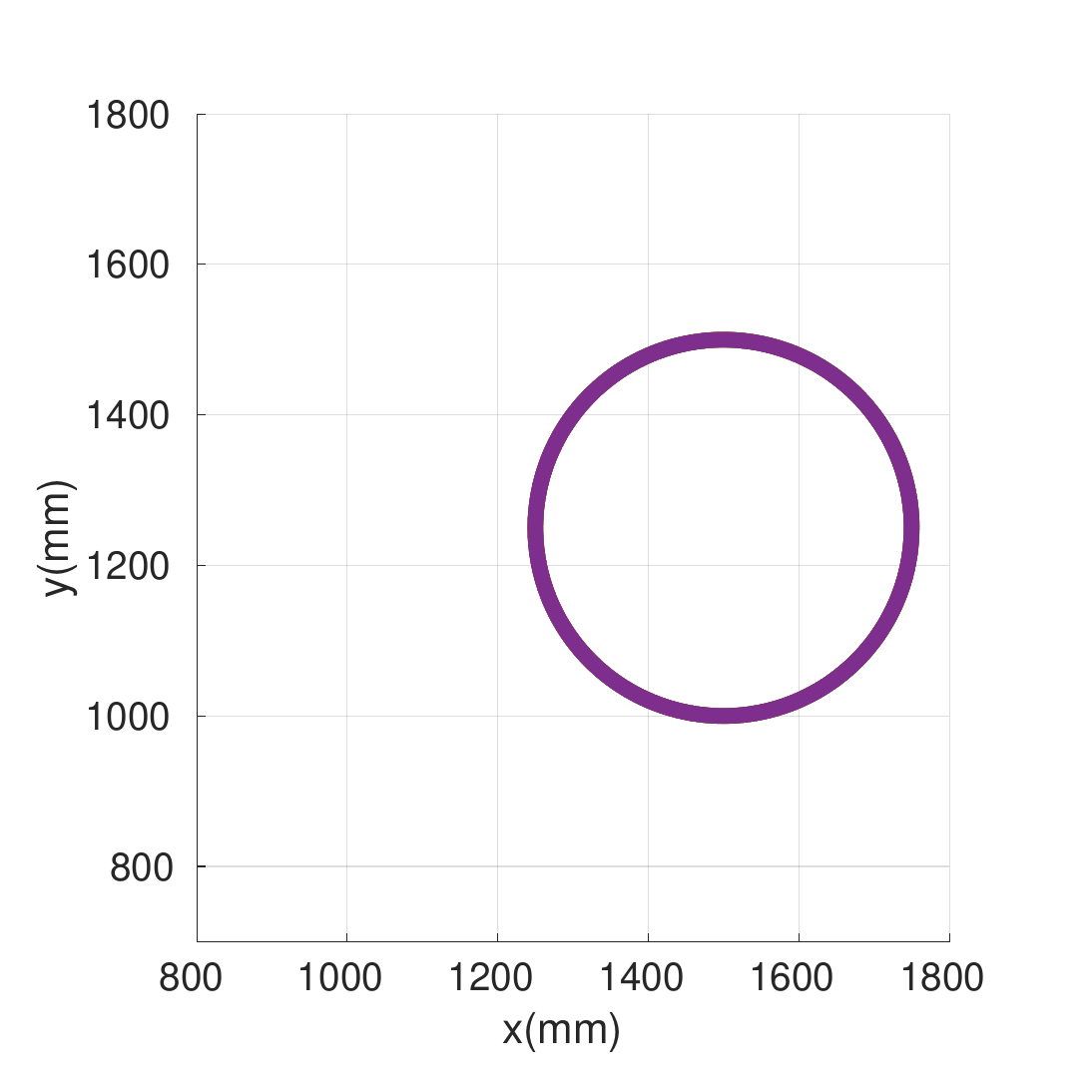}  
  \caption{\scriptsize WMVN / MX / Path 1 / Zero noise}
  \label{fig:3dof-MX-WMVN-Circle}
\end{subfigure}
%\hspace{5mm}
\begin{subfigure}{.245\textwidth}
  \centering
  % include  image
  \includegraphics[width=\textwidth,trim={0cm 0.45cm 1cm 0.5cm},clip]{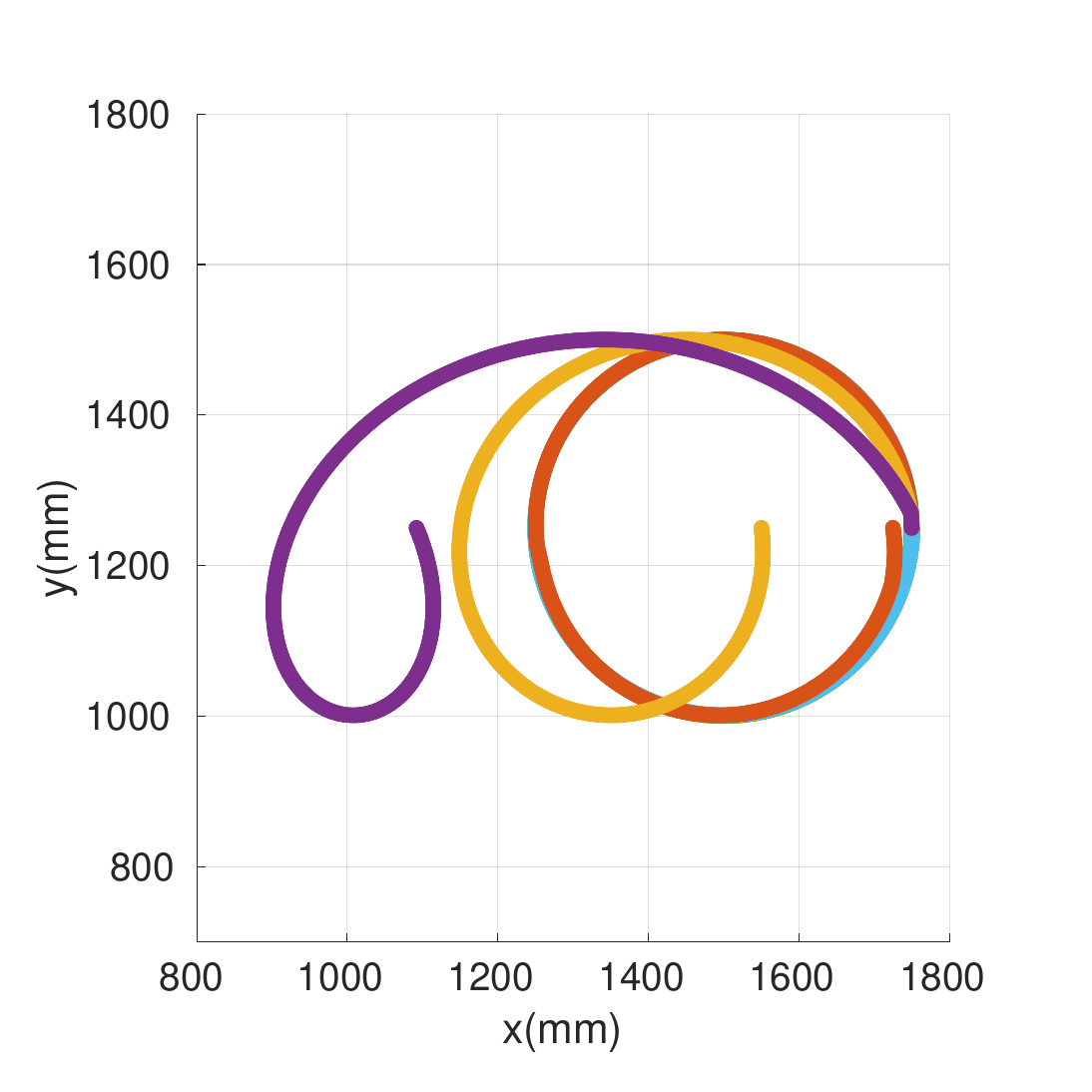}  
  \caption{\scriptsize MAN / MP / Path 1 / Zero noise}
  \label{fig:3dof-MP-MAN-Circle}
\end{subfigure}
%\hspace{5mm}
\begin{subfigure}{.245\textwidth}
  \centering
  % include  image
  \includegraphics[width=\textwidth,trim={0cm 0.45cm 1cm 0.5cm},clip]{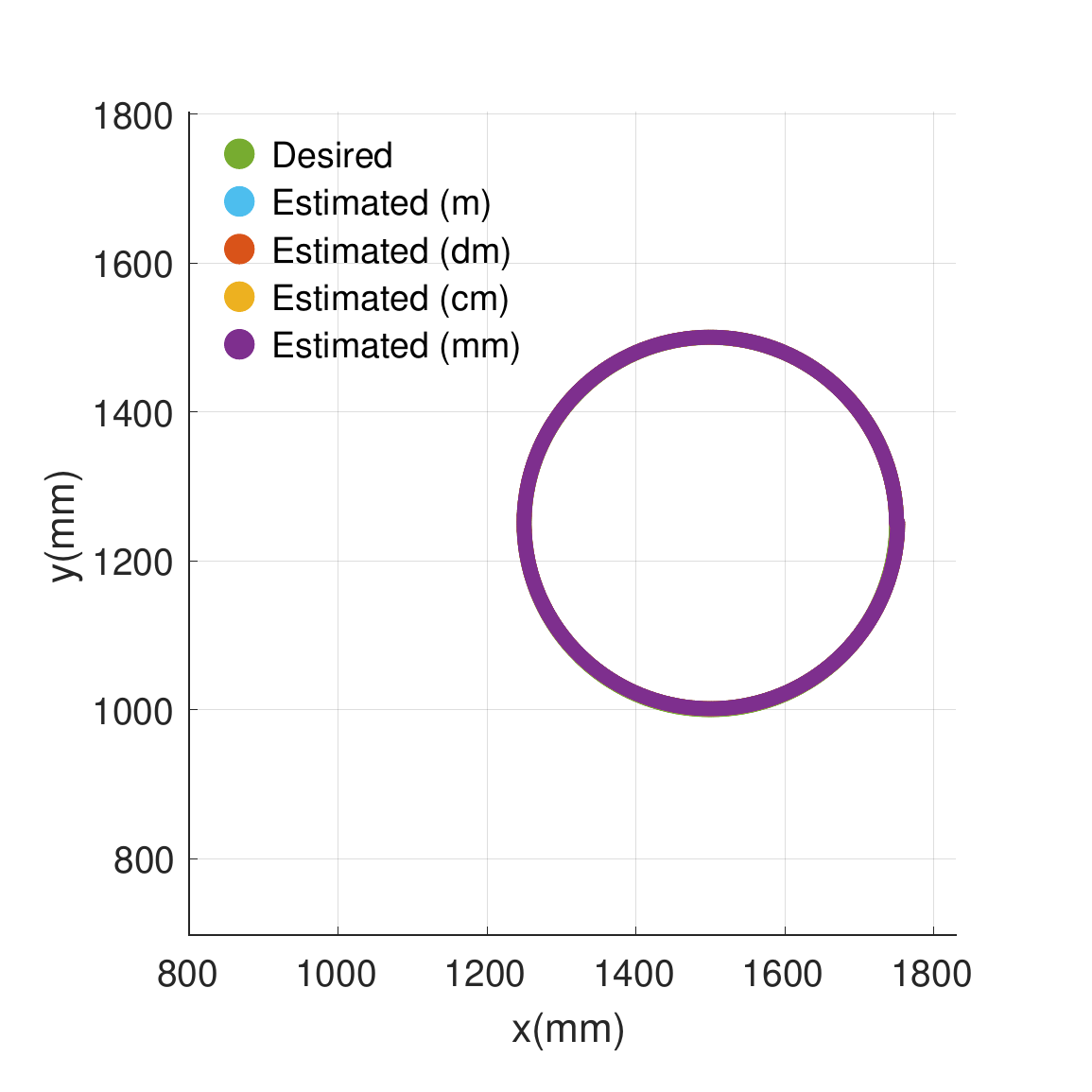}  
  \caption{\scriptsize MAN / MX / Path 1 / Zero noise}
  \label{fig:3dof-MX-MAN-Circle}
\end{subfigure}
\begin{subfigure}{.245\textwidth}
  \centering
  % include  image
  \includegraphics[width=\textwidth,trim={0cm 0.45cm 1cm 0.5cm},clip]{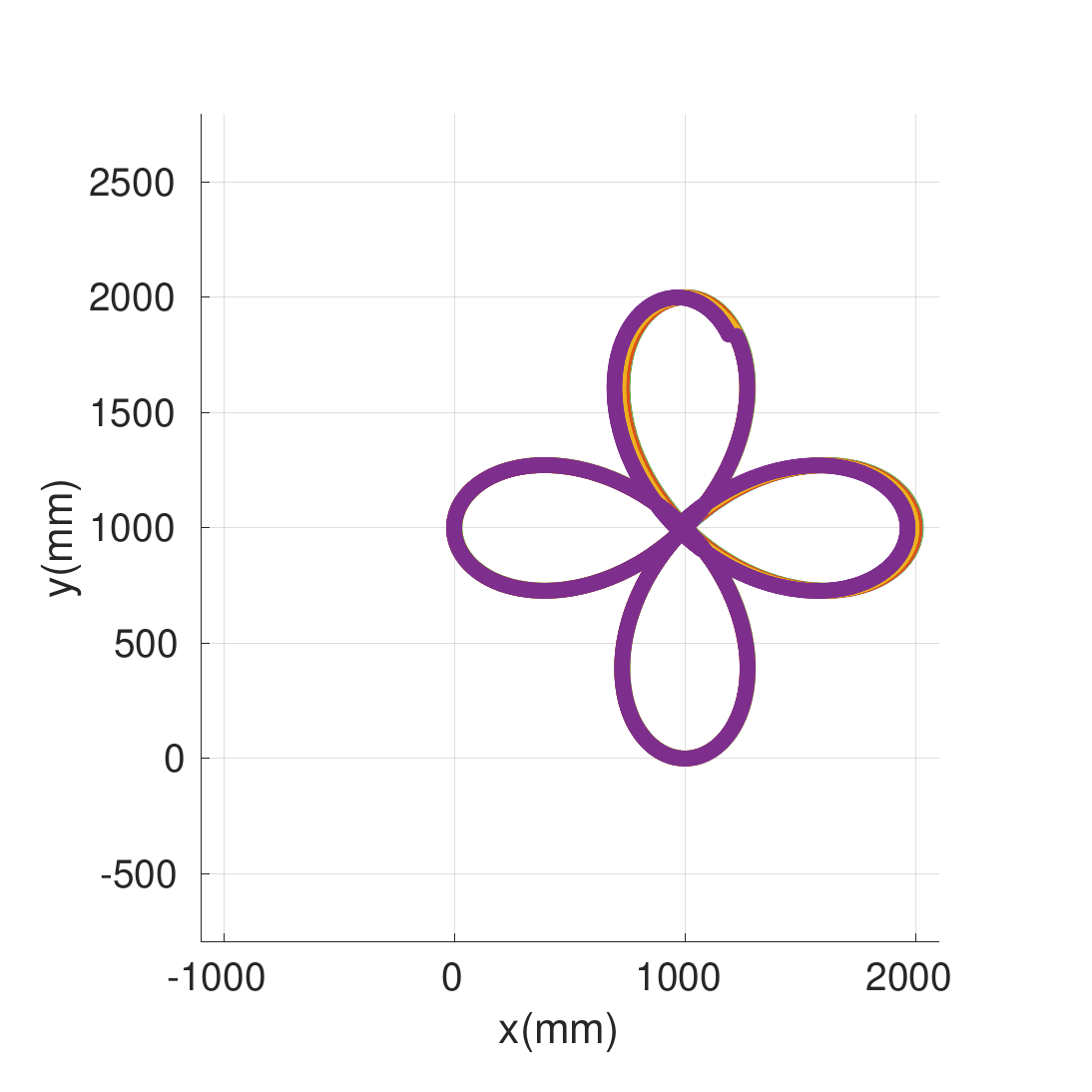}  
  \caption{\scriptsize WMVN / MP / Path 2 / Zero noise}
  \label{fig:3dof-MP-WMVN-Rhodonea}
\end{subfigure}
%\hspace{5mm}
\begin{subfigure}{.245\textwidth}
  \centering
  % include  image
  \includegraphics[width=\textwidth,trim={0cm 0.45cm 1cm 0.5cm},clip]{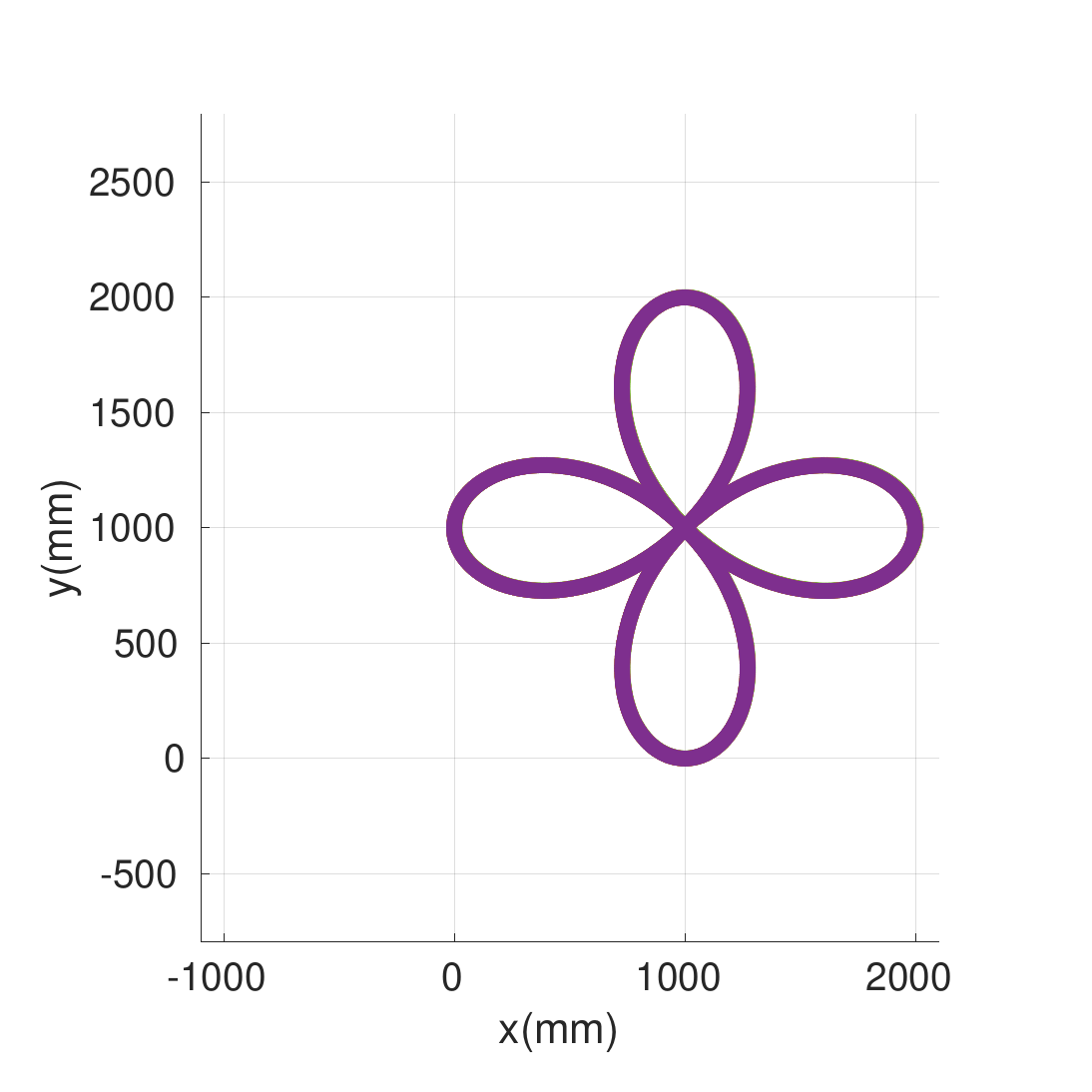}  
  \caption{\scriptsize WMVN / MX / Path 2 / Zero noise}
  \label{fig:3dof-MX-WMVN-Rhodonea}
\end{subfigure}
%\hspace{5mm}
\begin{subfigure}{.245\textwidth}
  \centering
  % include  image
  \includegraphics[width=\textwidth,trim={0cm 0.45cm 1cm 0.5cm},clip]{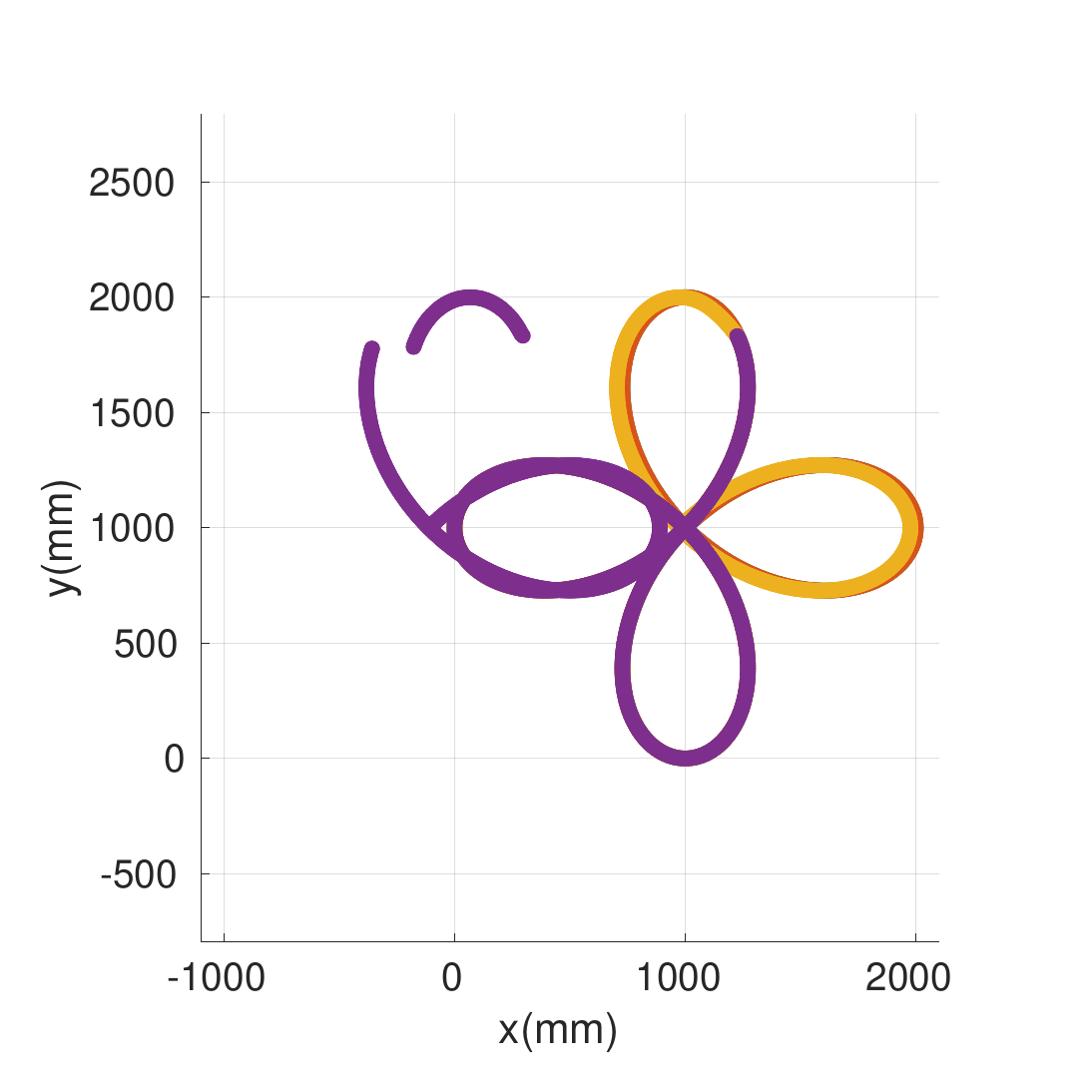}  
  \caption{\scriptsize MAN / MP / Path 2 / Zero noise}
  \label{fig:3dof-MP-MAN-Rhodonea}
\end{subfigure}
%\hspace{5mm}
\begin{subfigure}{.245\textwidth}
  \centering
  % include  image
  \includegraphics[width=\textwidth,trim={0cm 0.45cm 1cm 0.5cm},clip]{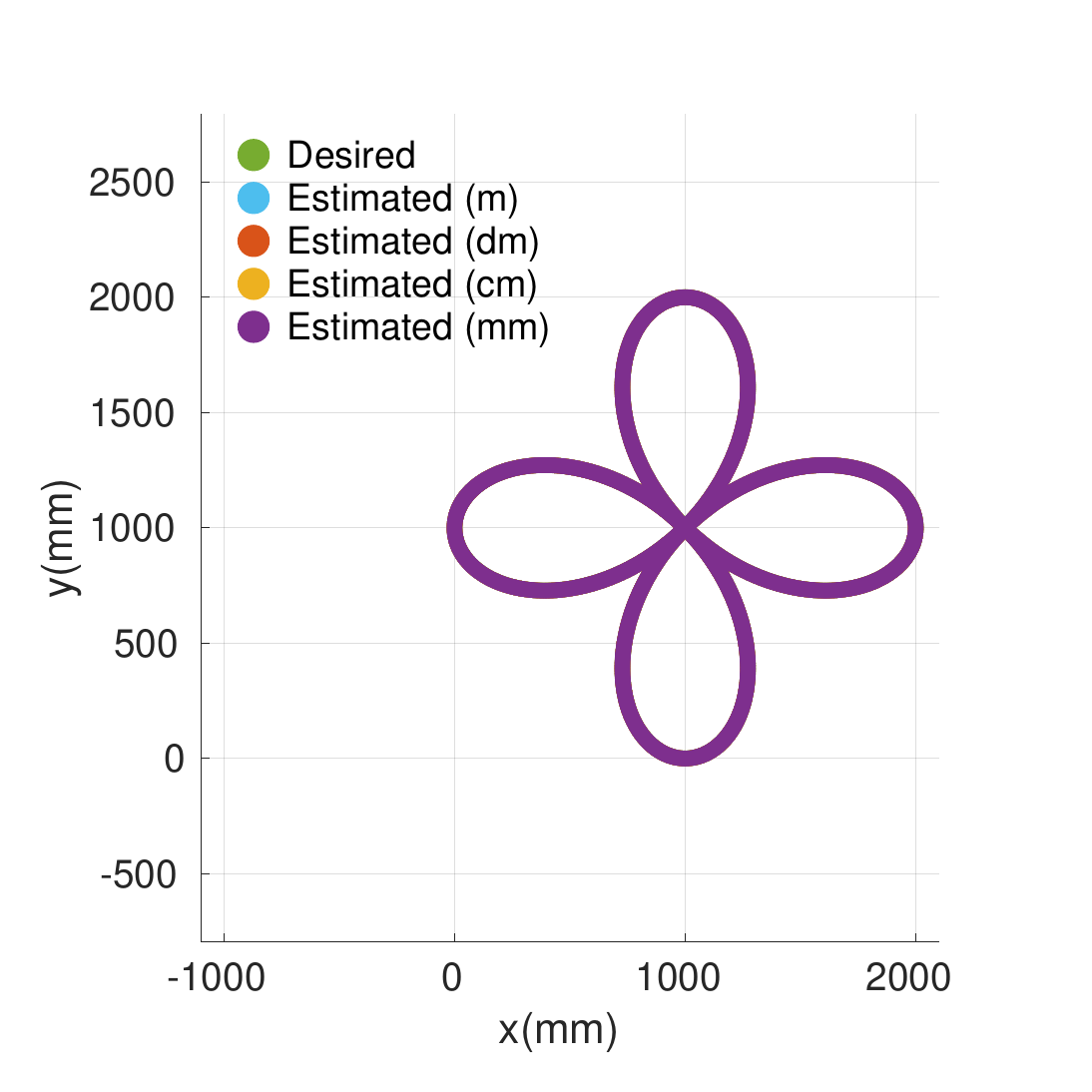}  
  \caption{\scriptsize MAN / MX / Path 2 / Zero noise}
  \label{fig:3dof-MX-MAN-Rhodonea}
\end{subfigure}
\begin{subfigure}{.245\textwidth}
  \centering
  % include  image
  \includegraphics[width=\textwidth,trim={0cm 0.45cm 1cm 0.5cm},clip]{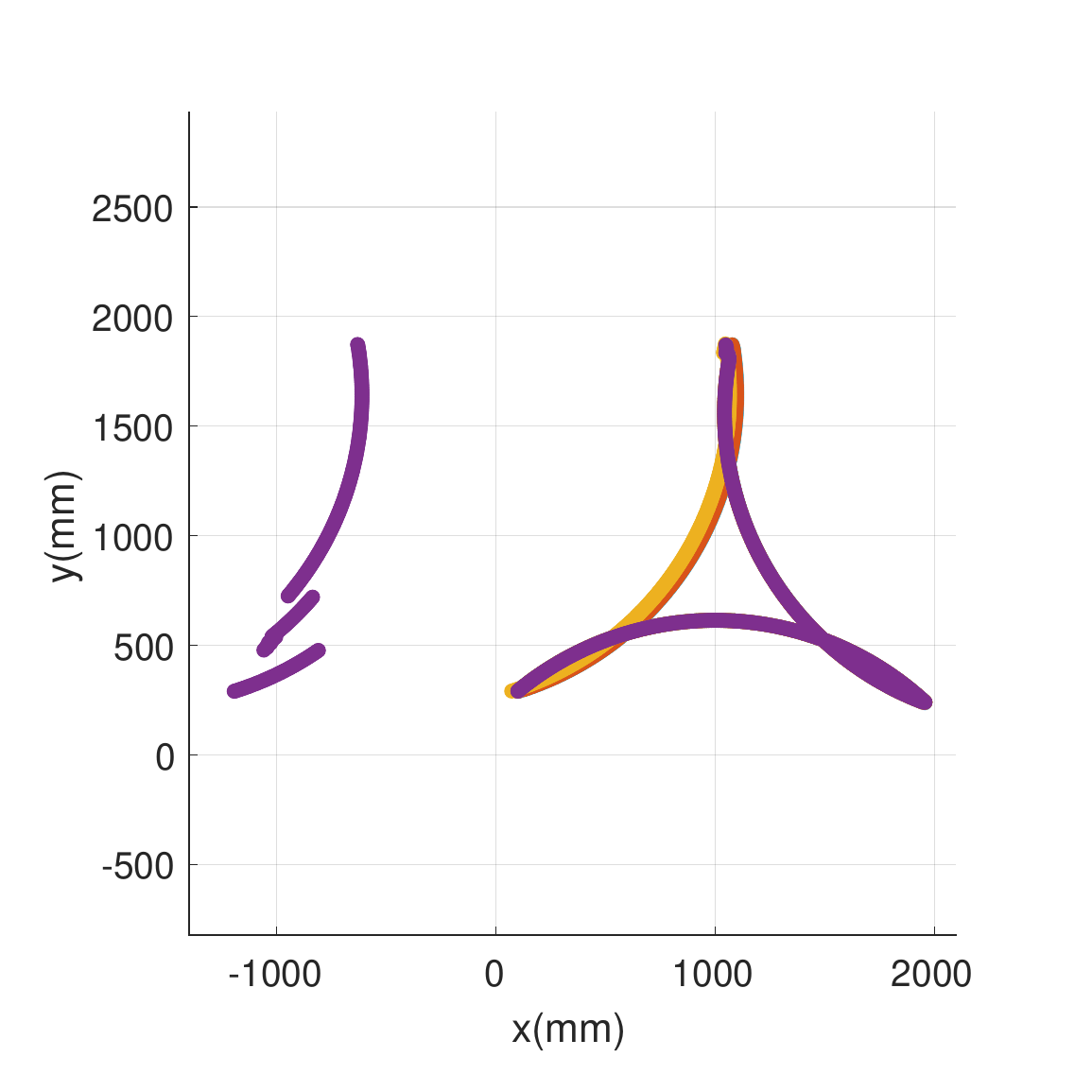}  
  \caption{\scriptsize WMVN / MP / Path 3 / Zero noise}
  \label{fig:3dof-MP-WMVN-Tricuspid}
\end{subfigure}
%\hspace{5mm}
\begin{subfigure}{.245\textwidth}
  \centering
  % include  image
  \includegraphics[width=\textwidth,trim={0cm 0.45cm 1cm 0.5cm},clip]{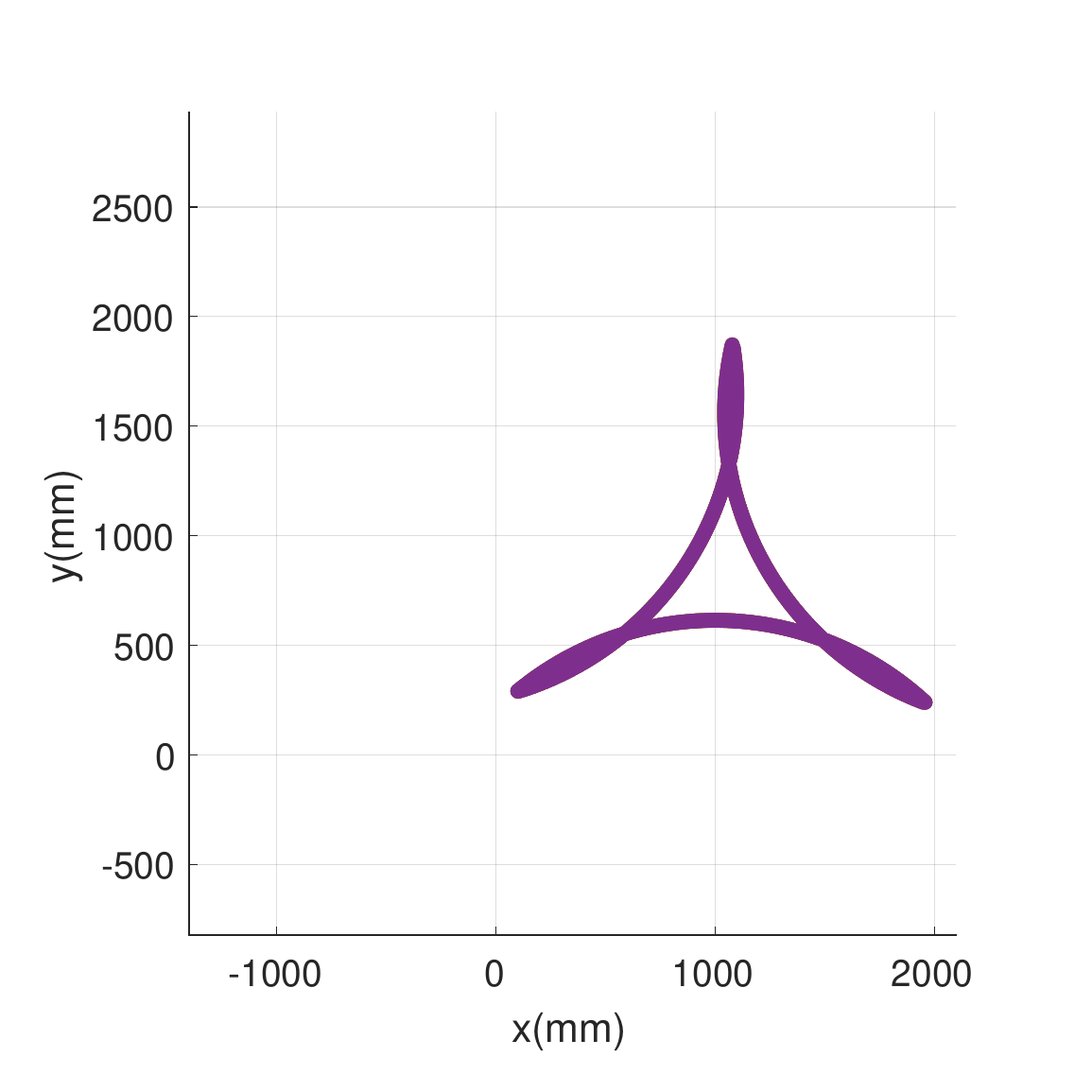}  
  \caption{\scriptsize WMVN / MX / Path 3 / Zero noise}
  \label{fig:3dof-MX-WMVN-Tricuspid}
\end{subfigure}
%\hspace{5mm}
\begin{subfigure}{.245\textwidth}
  \centering
  % include  image
  \includegraphics[width=\textwidth,trim={0cm 0.45cm 1cm 0.5cm},clip]{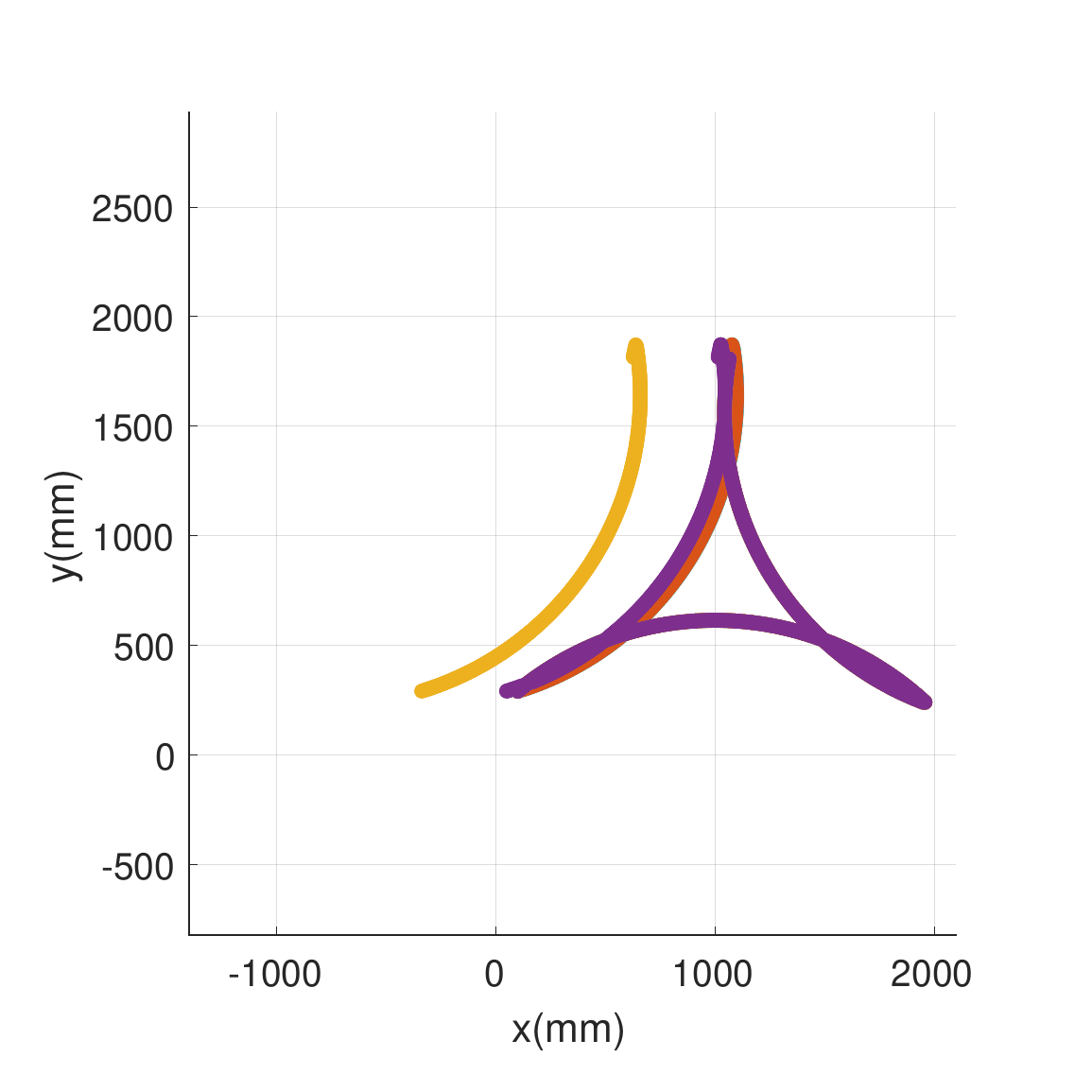}  
  \caption{\scriptsize MAN / MP / Path 3 / Zero noise}
  \label{fig:3dof-MP-MAN-Tricuspid}
\end{subfigure}
%\hspace{5mm}
\begin{subfigure}{.245\textwidth}
  \centering
  % include  image
  \includegraphics[width=\textwidth,trim={0cm 0.45cm 1cm 0.5cm},clip]{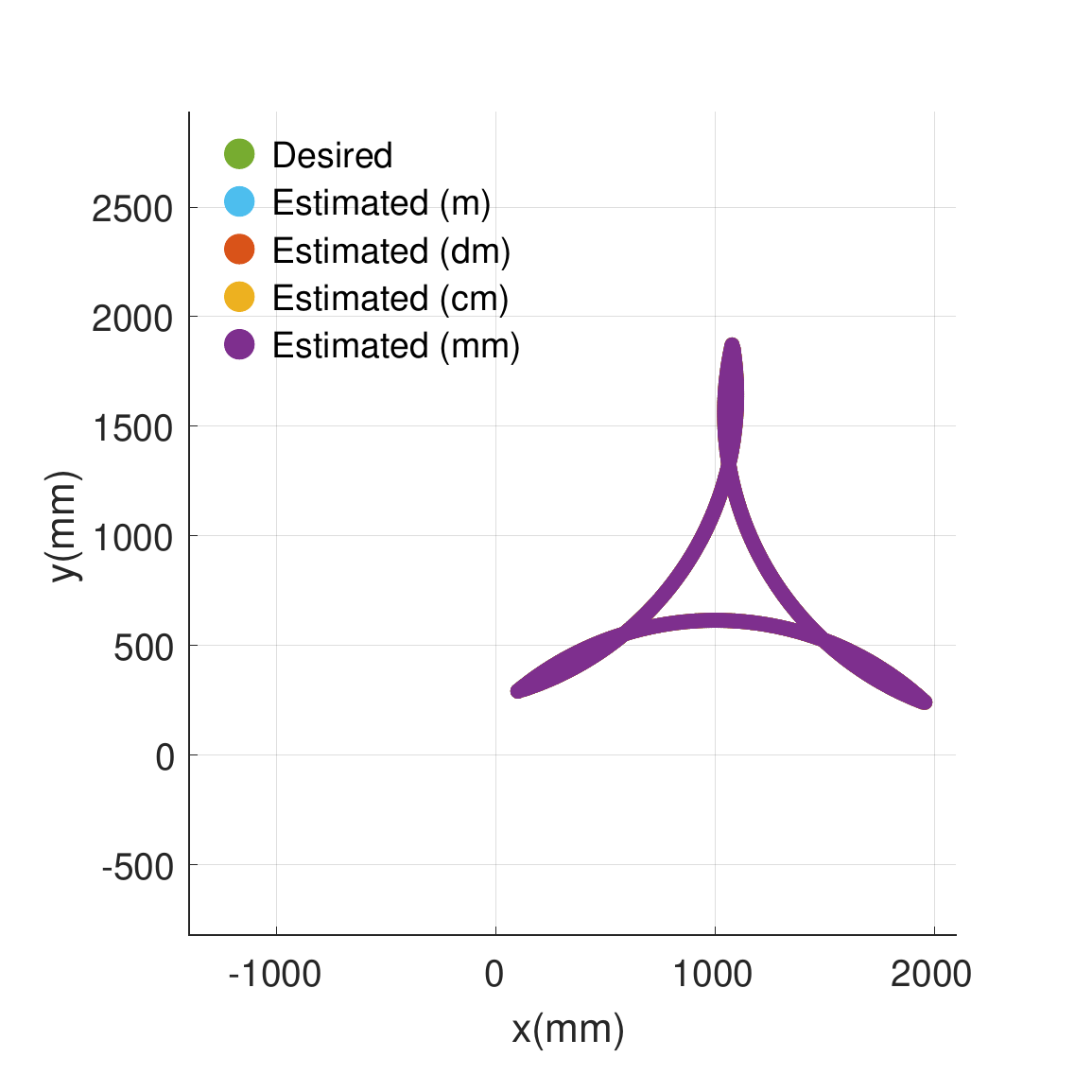}  
  \caption{\scriptsize MAN / MX / Path 3 / Zero noise}
  \label{fig:3dof-MX-MAN-Tricuspid}
\end{subfigure}
\caption{\footnotesize Desired path versus estimated paths while varying the units from $m$, $dm$, $cm$, to $mm$ for the 3DoF (2RP) using respectively the WMVN and MAN under zero noise with both the MP- and MX-GI's. Similar behaviors are observed with the other schemes.}
\label{fig:path-planning-3DoF}
\vspace{-4mm}
\end{figure*}

\section{Experimental Results and Discussion}
In this section, we present the comparative results obtained with the two aforementioned incommensurate redundant robots having their DH parameters in Table \ref{tab:DH-parameters}. As mentioned in section \ref{sec:numerical-example}, the issues related to the choice of units happen with all Jacobian types. Here, the Geometric Jacobian is employed in all the investigated schemes as it gets cumbersome to derive the Analytical Jacobian for higher DoF robots. Multiple experiments were run using various paths to show the unit- and noise-consistency properties of the proposed MX-GI PPP and MX-GI PRMP schemes. Figure \ref{fig:paths-investigated} depicts all the investigated paths for the 3DoF robot (circle, rhodonea, and tricuspid) and the 7DoF robot (interlaced circle, 3D rhodonea, and bent tricuspid). As in \cite{chan1995weighted, flacco2012motion, flacco2015control, guo2017new, wang2019feedback}, and since the rotational components of the end-effector pose (i.e. $RPY$ in $D_d(t)$) are not affected by unit choices (i.e. they lie on the bottom rows of the block-partitioned $J$), the experiments presented here focused only on reporting the end-effector positions ($XYZ$ in $D_d(t)$).

\subsection{In the case of the 3DoF-2RP redundant robot}

Table \ref{tab:schemes-units-3DoF-zero} presents the comparative results for all the investigated schemes when applied to the 3DoF incommensurate robot. In the absence of noise, rows 5 to 10 on the same Table, all schemes were able to achieve less than $1mm$ average error when the unit chosen was $m$, whether the MP and MX-GI's were employed for the circle and tricuspid paths. For the rhodonea path, the average errors were less than  $1mm$ for PID-PPP, MAN and FPBM schemes; and $3.07mm$, $3.55mm$ and $2.05mm$ respectively for the WMVN, V-SNS, and A-SNS when the unit chosen was $m$. However, when all variables were set to any of the remaining units ($cm$, $dm$ or $mm$) the average error was no longer consistent across all cases for the MP-GI, while the MX-GI consistently repeated the path followed in the $m$ case. As depicted in Figures \ref{fig:scheme-comparison-3DoF-Circle}, \ref{fig:scheme-comparison-3DoF-Rhodonea}, and \ref{fig:scheme-comparison-3DoF-Tricuspid}; the box-plots for the error vary with the unit choice when using the MP-GI, but they are the same when using the MX-GI for both V-PPP-derived schemes (i.e. WMVN, PID-PPP and V-SNS) and A-PPP-derived schemes (i.e. MAN, FPBM, and A-SNS). Figure \ref{fig:path-planning-3DoF} illustrates the trajectory followed by the end-effector with the WMVN and MAN schemes in the absence of noise. In Figures \ref{fig:3dof-MP-WMVN-Circle}, \ref{fig:3dof-MP-WMVN-Rhodonea}, and \ref{fig:3dof-MP-WMVN-Tricuspid}, we observe that the robot produces undesirable behaviors by failing to accurately and consistently follow the desired investigated paths under some of the units with the MP-GI-based WMVN scheme. However, Figures  \ref{fig:3dof-MX-WMVN-Circle}, \ref{fig:3dof-MX-WMVN-Rhodonea}, and \ref{fig:3dof-MX-WMVN-Tricuspid},  show that with the MX-GI-based WMVN scheme consistently succeeds for the same path with all units. The same observations can be made under the MAN scheme by looking at Figures \ref{fig:3dof-MP-MAN-Circle},  \ref{fig:3dof-MX-MAN-Circle},  \ref{fig:3dof-MP-MAN-Rhodonea},  \ref{fig:3dof-MX-MAN-Rhodonea}, \ref{fig:3dof-MP-MAN-Tricuspid}, and \ref{fig:3dof-MX-MAN-Tricuspid}. 

Finally, in the presence of noise, Table \ref{tab:schemes-units-3DoF-zero} present the results obtained with constant (rows 12 to 17), time-varying (rows 19 to 24), and random (rows 26 to 31)  noise contamination. Here, $\delta(t)$ was set in the following manner: (1) constant noise $\delta(t) = u*[0.3, 0.5]^{T}$ for velocity-level schemes \cite{guo2017new} and $\delta(t) = u*[0.03, 0.05]^{T}$ for acceleration-level schemes; (2) time-varying noise $\delta(t) = u*[0.3*sin(2t), 0.5*cos(2t)]^{T}$ for velocity-level schemes \cite{guo2017new} and $\delta(t) = u*[0.03*sin(2t), 0.05*cos(2t)]^{T}$ for acceleration-level schemes; (3) random noise $\delta(t)= u* [r_{1}, r_{2}, \dots, r_{m}]^{T}  \in \mathbb{R}^{m}$ with each $r_{i}$ being a seeded random number between $[0,1]$; where $t$ is the time and $u$ is either $1$, $10$, $100$, or $1000$ according to the choice of $m$, $dm$, $cm$, or $mm$ as unit. From rows 13 to 16, 20 to 23, and 27 to 31  on Table \ref{tab:schemes-units-3DoF-zero}, we observe that the WMVN, V-SNS, MAN, and A-SNS schemes, which do not have noise suppression capabilities, fail to follow the desired path across different units when using either the MP- and MX-GI's. However, the MX-GI is still capable of preserving its unit-consistency properties across different units even though the error is large. That is, the noise suppression capability of a scheme is not inherent to the choice of GI, but to its formulation, and even a consistent GI can not fix that. This observation is consolidated by the PID-PPP (rows 12, 19, 26) and FPBM (rows 17, 24, 31) schemes which succeed in their trajectory following tasks in the presence of noise because of their noise-suppression properties. Similar to \cite{guo2017new}, the feedback gains were set to $\alpha = \beta = 1000$ for the PID-PPP scheme. In the case of the FPBM scheme, the gains had to be empirically adjusted to $k_{1} = k_{2} = 1000$ from those used in the original MP-GI-based FPBM (\cite{wang2019feedback}) due to the use of a different GI and a different robot: here a MX-GI-based FPBM and a 3DoF-RRP.

\begin{table*}[!t] 
\notsotinythree
\caption{\footnotesize Average errors (in $mm$) between the desired and estimated circular using the investigated PPP schemes for the 7DoF (2RP4R) manipulator with zero, constant, time-varying, and random noises. The $dm$ results have been omitted for space consideration. When WC for "Without Completing" exists in a table cell, it indicates that the scheme did not complete the trajectory following task because MATLAB reached its highest precision.}\label{tab:schemes-units-7DoFP-zero}
    \centering
    \begin{threeparttable}
         \begin{tabular}{|p{1.2mm}|p{7.3mm}|c|c|c|c|c|c|c|c|c|c|c|c|c|c|c|c|c|c|} 
         %\begin{tblr}{
         %       colspec = {|c|c|c|c|},
                %row{4} = {gray9},
                %row{12} = {gray9},
                %row{14} = {gray9},
                %row{22} = {gray9},
                %row{31} = {gray9},
                %row{32} = {gray9},
                %column{3} = {teal7},
                %cell{2}{3} = {yellow7},
        %      }
         \hline
         %\multirow{3}{3.5em}{Schemes} & \multicolumn{6}{c|}{Zero Noise} \\
         %\cline{2-7} 
         1 & \multirow{3}{3.5em}{Schemes} & \multicolumn{6}{c|}{7DoF - Path 1 (Interlaced Circle)} & \multicolumn{6}{c|}{7DoF - Path 2 (3D Rhodonea)} & \multicolumn{6}{c|}{7DoF - Path 3 (Bent Tricuspid)} \\ 
         \cline{3-20}         
         2 & & \multicolumn{3}{c|}{Original MP-GI} & \multicolumn{3}{c|}{Proposed MX-GI} & \multicolumn{3}{c|}{Original MP-GI} & \multicolumn{3}{c|}{Proposed MX-GI} & \multicolumn{3}{c|}{Original MP-GI} & \multicolumn{3}{c|}{Proposed MX-GI}\\  
         \cline{3-20} 
         3 & & $m$ & $cm$ & $mm$ & $m$ & $cm$ & $mm$ & $m$ & $cm$ & $mm$ & $m$ & $cm$ & $mm$ & $m$ & $cm$ & $mm$ & $m$ & $cm$ & $mm$\\  
         \hline\hline 
         4 & \multicolumn{19}{c|}{\textbf{Errors for Zero Noise in $mm$}}\\ 
         \hline
         5 & PID-PPP & 0.00 & 0.01 & 978.1 & \cellcolor{gray9}0.00 & \cellcolor{gray9}0.00 & \cellcolor{gray9}0.00  & 0.00 &  0.02 & 148.5 & \cellcolor{gray9}0.00 &  \cellcolor{gray9}0.00 &  \cellcolor{gray9}0.00 & 0.00 & 0.00 & 663.6 & \cellcolor{gray9}0.00 & \cellcolor{gray9}0.00 & \cellcolor{gray9}0.00\\
         \hline
         6 & WMVN & 0.41 & 1.11 & 333.2 & \cellcolor{gray9}0.59 & \cellcolor{gray9}0.59 & \cellcolor{gray9}0.59 & 2.58 & 2.33 &  267.1 &    \cellcolor{gray9}0.80 & \cellcolor{gray9}0.80 & \cellcolor{gray9}0.80  & 0.72 & 1.56 & 522.4 & \cellcolor{gray9}0.69 & \cellcolor{gray9}0.69  & \cellcolor{gray9}0.69\\
         \hline
         7 & V-SNS & 0.43 & 27.9 & 28.0 & \cellcolor{gray9}0.96 & \cellcolor{gray9}0.96 & \cellcolor{gray9}0.96 & 1.84 &  2.85 &  48.08 &   \cellcolor{gray9}0.81 & \cellcolor{gray9}0.81 & \cellcolor{gray9}0.81 &  0.71 & 1.88 & 261.9 & \cellcolor{gray9}0.70 & \cellcolor{gray9}0.70 & \cellcolor{gray9}0.70  \\ 
         \hline
         8 & MAN & 0.17 & 0.16 & 807.6 & \cellcolor{gray9}0.15 & \cellcolor{gray9}0.15 & \cellcolor{gray9}0.15  & 0.04 &  9.53 & 310.7  &  \cellcolor{gray9}0.05 & \cellcolor{gray9}0.05 &  \cellcolor{gray9}0.05 & 0.04 & 0.27 & 576.9 & \cellcolor{gray9}0.04 & \cellcolor{gray9}0.04 & \cellcolor{gray9}0.04\\
         \hline
         9 & A-SNS & 1.66 & 103.8 & 816.2 & \cellcolor{gray9}1.01 & \cellcolor{gray9}1.01 & \cellcolor{gray9}1.01 & 6.01 & 59.1 & 433.6 & \cellcolor{gray9}0.29 & \cellcolor{gray9}0.29 & \cellcolor{gray9}0.29 & 0.48 & 2.65 & 537.4 & \cellcolor{gray9}0.47 & \cellcolor{gray9}0.47 & \cellcolor{gray9}0.47  \\
         \hline
         10 & FPBM & 0.05 & 2.03 & 675.4 & \cellcolor{gray9}0.05 & \cellcolor{gray9}0.05 & \cellcolor{gray9}0.05  & 0.12 & 4.19 & 315.5 &   \cellcolor{gray9}0.26  & \cellcolor{gray9}0.26  &  \cellcolor{gray9}0.26 & 0.00 & 0.16 & 309.9 & \cellcolor{gray9}0.00 & \cellcolor{gray9}0.00 & \cellcolor{gray9}0.00\\
         \hline
         \hline
         11 & \multicolumn{19}{c|}{\textbf{Errors for Constant Noise in $mm$}}\\
         \hline
         12 & PID-PPP & 0.66 & 0.66 & 965.7 & \cellcolor{gray9}0.66 & \cellcolor{gray9}0.66 & \cellcolor{gray9}0.66 & 0.64 & 0.68 & 183.5 & \cellcolor{gray9}0.64 & \cellcolor{gray9}0.64 & \cellcolor{gray9}0.64 & 1.31 & 1.31 & 298.6 & \cellcolor{gray9}1.31 & \cellcolor{gray9}1.31  & \cellcolor{gray9}1.31 \\
         \hline
         13 & WMVN &  2.0E3 & 2.1E3 & 1.7E3 & \cellcolor{red9}2.1E3 & \cellcolor{red9}2.1E3 & \cellcolor{red9}2.1E3 & 2.1E3 & 2.0E3 & 870.0 & \cellcolor{red9}2.0E3 & \cellcolor{red9}2.0E3 & \cellcolor{red9}2.0E3 & 6.5E3 & 6.6E3 & 1.2E3 & \cellcolor{red9}6.5E3 & \cellcolor{red9}6.5E3 & \cellcolor{red9}6.5E3 \\
         \hline
         14 & V-SNS &  1.2E3 & 1.4E3 & 1.2E3 & \cellcolor{red9}1.2E3 & \cellcolor{red9}1.2E3 & \cellcolor{red9}1.2E3  & 2.1E3 & 2.0E3 & 2.0E3 & \cellcolor{red9}2.0E3 & \cellcolor{red9}2.0E3 & \cellcolor{red9}2.0E3 & 6.6E3 & 6.6E3 & 6.5E3 & \cellcolor{red9}6.5E3 & \cellcolor{red9}6.5E3 & \cellcolor{red9}6.5E3 \\ 
         \hline
         15 & MAN &  2.1E3 & 2.1E3 & 1.4E3 & \cellcolor{red9}2.1E3 & \cellcolor{red9}2.1E3 & \cellcolor{red9}2.1E3  & 2.1E3 & 2.1E3  & 533.5 & \cellcolor{red9}2.1E3 & \cellcolor{red9}2.1E3 & \cellcolor{red9}2.1E3 & 6.7E3 & 6.5E3 & 660.9 & \cellcolor{red9}6.5E3 & \cellcolor{red9}6.5E3 & \cellcolor{red9}6.5E3  \\
         \hline
         16 & A-SNS &  2.1E3 & WC & WC & \cellcolor{red9}2.1E3 & \cellcolor{red9}2.1E3 & \cellcolor{red9}2.1E3  & 2.1E3 & 2.1E3 & 466.3 & \cellcolor{red9}2.1E3 & \cellcolor{red9}2.1E3 & \cellcolor{red9}2.1E3 & 6.5E3 & 6.3E3 & 594.2 & \cellcolor{red9}6.6E3 & \cellcolor{red9}6.6E3 & \cellcolor{red9}6.6E3 \\
         \hline
         17 & FPBM & 0.07 & 25.4 & 18.6 & \cellcolor{gray9}0.07 & \cellcolor{gray9}0.07 & \cellcolor{gray9}0.07  & 0.12 & 4.05 & 332.5 & \cellcolor{gray9}0.13 & \cellcolor{gray9}0.13 & \cellcolor{gray9}0.13 & 0.06 & 0.19 & 313.4 & \cellcolor{gray9}0.06 & \cellcolor{gray9}0.06 & \cellcolor{gray9}0.06 \\
         \hline
         \hline
         18 & \multicolumn{19}{c|}{\textbf{Errors for Time-varying Noise in $mm$}}\\
         \hline
         19 & PID-PPP & 0.36 & 0.36 & 1.0E3 & \cellcolor{gray9}0.36 & \cellcolor{gray9}0.36 & \cellcolor{gray9}0.36  & 0.45 & 0.47 & 183.7 & \cellcolor{gray9}0.45 & \cellcolor{gray9}0.45 & \cellcolor{gray9}0.45 & 0.73 & 0.73 & 683.1 & \cellcolor{gray9}0.73 & \cellcolor{gray9}0.73 & \cellcolor{gray9}0.73 \\
         \hline
         20 & WMVN &  245.6 & 246.4 & 788.6 & \cellcolor{red9}245.6 & \cellcolor{red9}245.6 & \cellcolor{red9}245.6  &  245.2 & 244.1 & 508.2 & \cellcolor{red9}245.1 & \cellcolor{red9}245.1 & \cellcolor{red9}245.1 & 241.8 & 240.5 & 510.8 & \cellcolor{red9}241.8 &  \cellcolor{red9}241.8 &  \cellcolor{red9}241.8  \\
         \hline
         21 & V-SNS &  545.6 & 546.3 & 546.7 & \cellcolor{red9}545.8 & \cellcolor{red9}545.8 & \cellcolor{red9}545.8  & 245.3 & 244.2 & 247.9 & \cellcolor{red9}245.0 & \cellcolor{red9}245.0 & \cellcolor{red9}245.0 & 241.8 & 235.7 & 250.4 & \cellcolor{red9}241.8 & \cellcolor{red9}241.8 &  \cellcolor{red9}241.8 \\  
         \hline
         22 & MAN &  245.3 & 245.4 & 743.5 & \cellcolor{red9}245.4 & \cellcolor{red9}245.4 & \cellcolor{red9}245.4  & 1.1E3 & 681.2 & WC & \cellcolor{red9}307.1 & \cellcolor{red9}307.1 & \cellcolor{red9}307.1 &  242.1 & 231.7 & 724.7 & \cellcolor{red9}242.1 & \cellcolor{red9}242.1 & \cellcolor{red9}242.1  \\
         \hline
         23 & A-SNS &  245.5 & WC & WC & \cellcolor{red9}245.5 & \cellcolor{red9}245.5 & \cellcolor{red9}245.5 & 411.8 & 2.3E3 & WC & \cellcolor{red9}344.5 & \cellcolor{red9}344.5 & \cellcolor{red9}344.5 & 242.0 & 240.4 & 747.9 & \cellcolor{red9}242.3 &  \cellcolor{red9}242.3 &  \cellcolor{red9}242.3\\
         \hline
         24 & FPBM & 0.03 & 26.3 & 83.1 & \cellcolor{gray9}0.04 & \cellcolor{gray9}0.04 & \cellcolor{gray9}0.04  & 0.11 & 4.21 & 310.4 & \cellcolor{gray9}0.11 & \cellcolor{gray9}0.11 & \cellcolor{gray9}0.11 &  0.02 & 0.16 & 315.9 & \cellcolor{gray9}0.02 &  \cellcolor{gray9}0.02 &  \cellcolor{gray9}0.02 \\
         \hline
         25 & \multicolumn{19}{|c|}{\textbf{Errors for Random Noise in $mm$}}\\ 
         \hline
         26 & PID-PPP & 0.94 & 0.94 & 898.5 & \cellcolor{gray9}0.94 & \cellcolor{gray9}0.94 & \cellcolor{gray9}0.94  & 0.91 & 0.98 & 155.0 & \cellcolor{gray9}0.91 & \cellcolor{gray9}0.91 & \cellcolor{gray9}0.91 & 2.2 & 2.2 & 441.6 & \cellcolor{gray9}2.2 & \cellcolor{gray9}2.2 & \cellcolor{gray9}2.2 \\
         \hline
         27 & WMVN & 2.7E3 & 2.7E3 & 2.5E3 & \cellcolor{red9}2.7E3 & \cellcolor{red9}2.7E3 & \cellcolor{red9}2.7E3  & 2.6E3 & 2.2E3 & 1.2E3 & \cellcolor{red9}2.6E3 & \cellcolor{red9}2.6E3 & \cellcolor{red9}2.6E3 & 8.6E3 & 8.7E3 & 996.6 & \cellcolor{red9}8.6E3 & \cellcolor{red9}8.6E3 & \cellcolor{red9}8.6E3 \\
         \hline
         28 & V-SNS &  1.5E3 & 1.5E3 & 1.4E3 & \cellcolor{red9}1.5E3 & \cellcolor{red9}1.5E3 & \cellcolor{red9}1.5E3  & 2.7E3 & 2.7E3 & 2.5E3 & \cellcolor{red9}2.7E3 & \cellcolor{red9}2.7E3 & \cellcolor{red9}2.7E3 & 8.6E3 & 8.7E3 & 8.7E3 & \cellcolor{red9}8.6E3 & \cellcolor{red9}8.6E3 & \cellcolor{red9}8.6E3   \\ 
         \hline
         29 & MAN & 784.7  & 522.7 & 8.6E3 & \cellcolor{red9}568.8 & \cellcolor{red9}568.8 & \cellcolor{red9}568.8 & 1.3E3 & 918.7 & WC & \cellcolor{red9}576.0 & \cellcolor{red9}576.0 & \cellcolor{red9}576.0 & 8.6E3 & 8.5E3 & 626.1 & \cellcolor{red9}8.6E3 & \cellcolor{red9}8.6E3 & \cellcolor{red9}8.6E3 \\
         \hline
         30 & A-SNS & 2.0E8  & WC & WC & \cellcolor{red9}2.0E8 & \cellcolor{red9}2.0E8 & \cellcolor{red9}2.0E8 & 415.9 & WC & WC & \cellcolor{red9}347.8 &  \cellcolor{red9}347.8 & \cellcolor{red9}347.8 & 8.6E3 & 8.3E3 & 691.0 & \cellcolor{red9}8.6E3 & \cellcolor{red9}8.6E3 & \cellcolor{red9}8.6E3 \\
         \hline
         31 & FPBM & 0.08 & 24.8 & 44.9 & \cellcolor{gray9}0.08 & \cellcolor{gray9}0.08 & \cellcolor{gray9}0.08  & 0.06 & 3.78 & 328.3 & \cellcolor{gray9}0.05 & \cellcolor{gray9}0.05 & \cellcolor{gray9}0.05 & 0.08 & 0.21 & 310.0 & \cellcolor{gray9}0.08 & \cellcolor{gray9}0.08 & \cellcolor{gray9}0.08 \\
         \hline
        %\end{tblr}
        \end{tabular}
        % Note under the table
        \begin{tablenotes}
        \small
        \item  
        \end{tablenotes}
    \end{threeparttable}
\vspace{-2mm}
\end{table*}

\begin{figure}[!t]
\centering
\begin{subfigure}[b]{.24\textwidth}
  \centering
  % include first image
  \includegraphics[width=\textwidth,trim={0cm 0.9cm 0cm 1cm},clip]{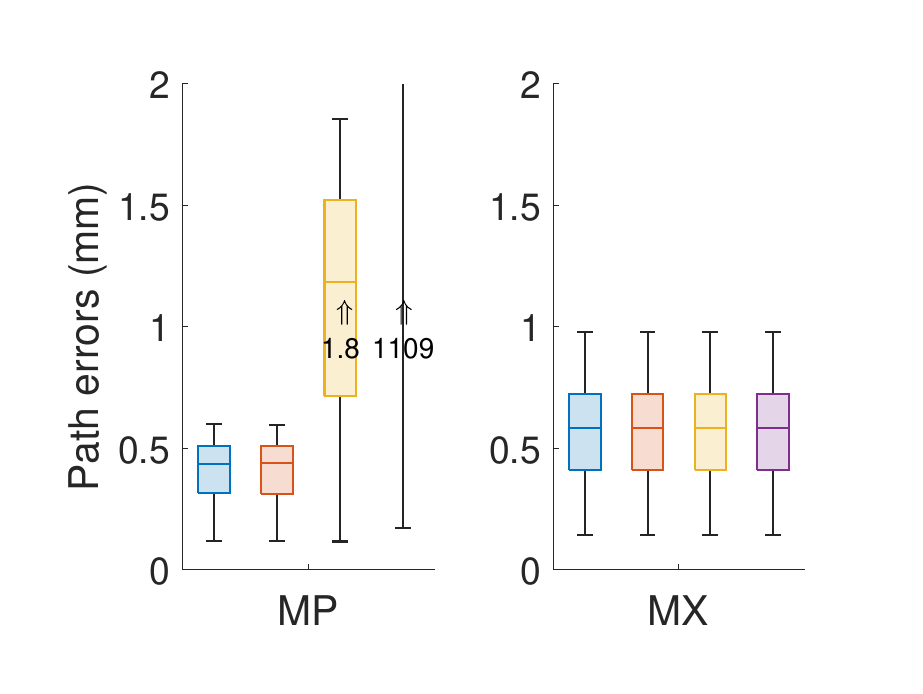} 
  \caption{\footnotesize WMVN / Zero noise / Path 1}
  \label{fig:7DoF-MVN-Circle}
\end{subfigure}
\begin{subfigure}[b]{.24\textwidth}
  \centering
  % include first image
  \includegraphics[width=\textwidth,trim={0cm 0.9cm 0cm 1cm},clip]{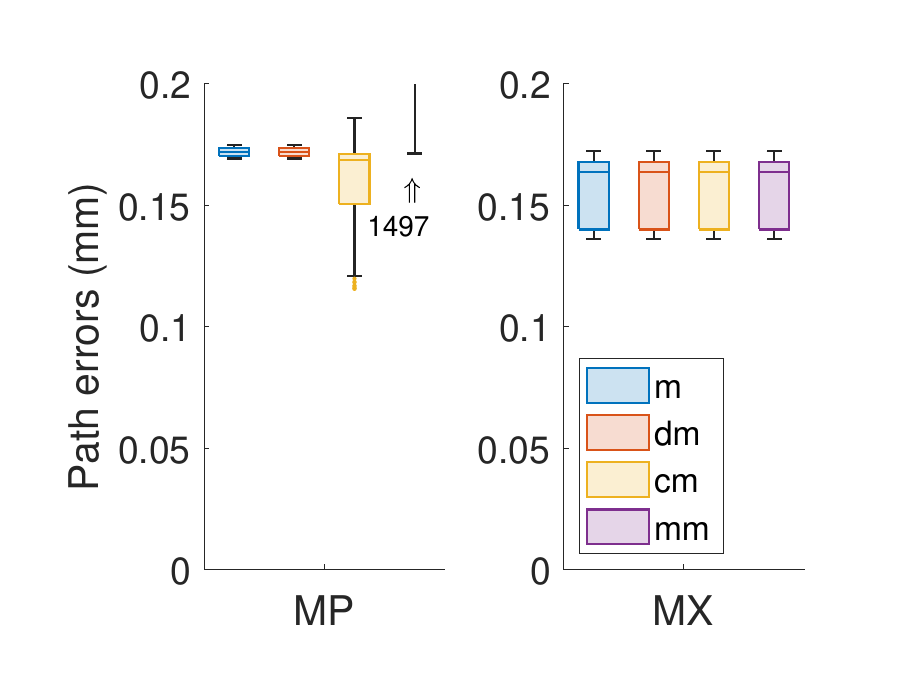} 
  \caption{\footnotesize MAN / Zero noise / Path 1}
  \label{fig:7DoF-MAN-Circle}
\end{subfigure}
\begin{subfigure}[b]{.24\textwidth}
  \centering
  % include third image
  \includegraphics[width=\textwidth,trim={0cm 0.9cm 0cm 1cm},clip]{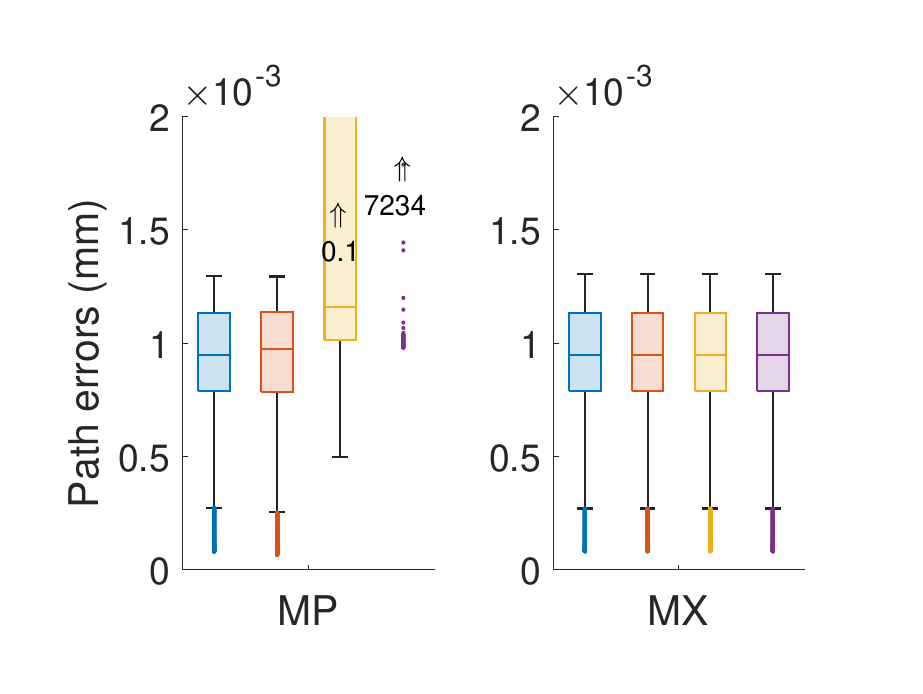} 
  \caption{\footnotesize PID-PPP / Zero noise / Path 1}
  \label{fig:7DoF-PID-PPP-Circle}
\end{subfigure}
\begin{subfigure}[b]{.24\textwidth}
  \centering
  % include third image
  \includegraphics[width=\textwidth,trim={0cm 0.9cm 0cm 1cm},clip]{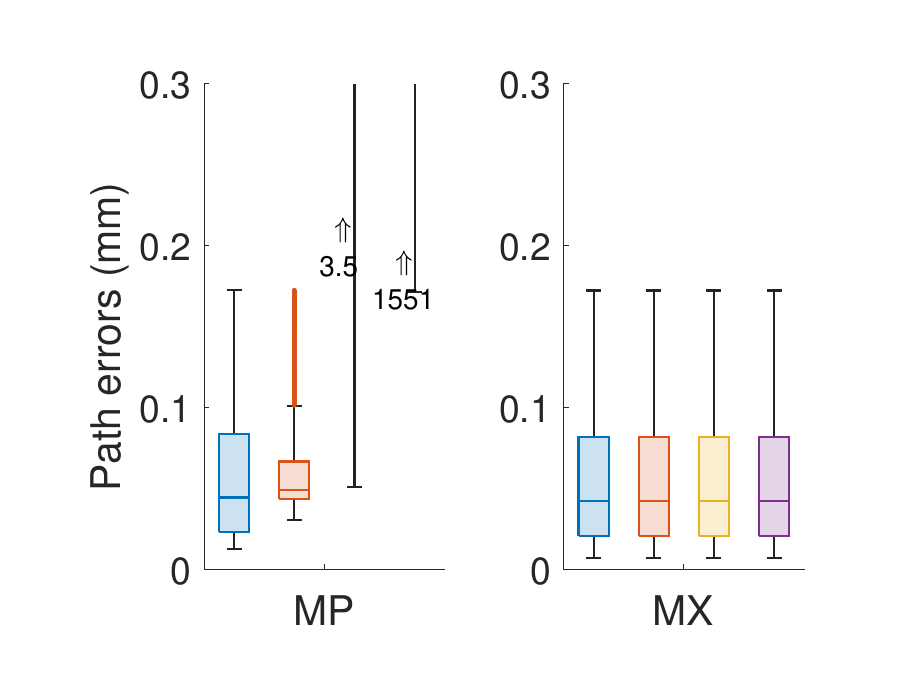} 
  \caption{\footnotesize FPBM / Zero noise / Path 1}
  \label{fig:7DoF-FPBM-Circle}
\end{subfigure}
\begin{subfigure}[b]{.24\textwidth}
  \centering
  % include second image
  \includegraphics[width=\textwidth,trim={0cm 0.9cm 0cm 1cm},clip]{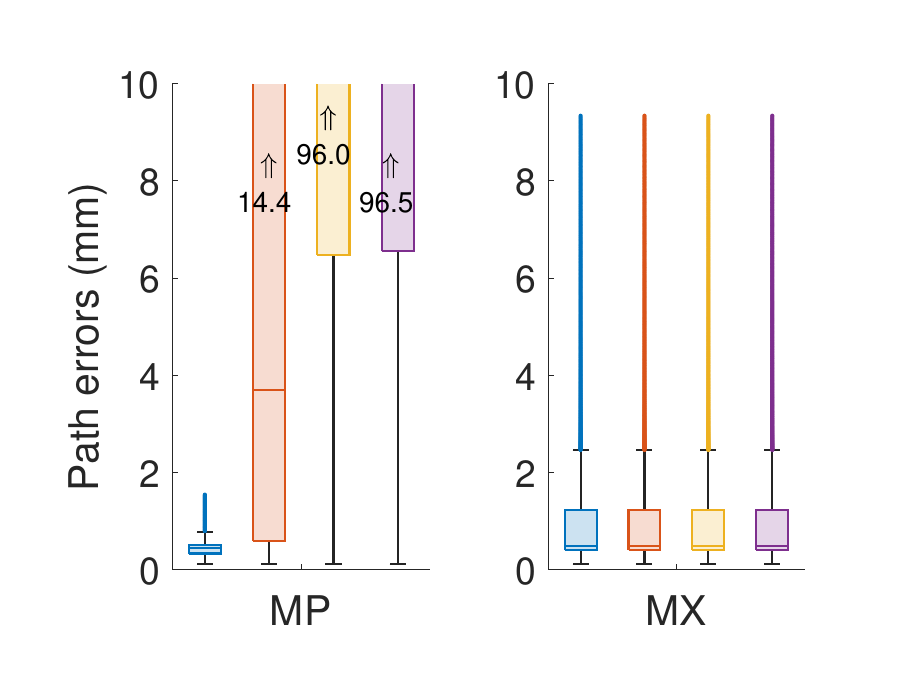} 
  \caption{\footnotesize V-SNS / Zero noise / Path 1}
  \label{fig:7DoF-V-SNS-Circle}
\end{subfigure}
\begin{subfigure}[b]{.24\textwidth}
  \centering
  % include second image
  \includegraphics[width=\textwidth,trim={0cm 0.9cm 0cm 1cm},clip]{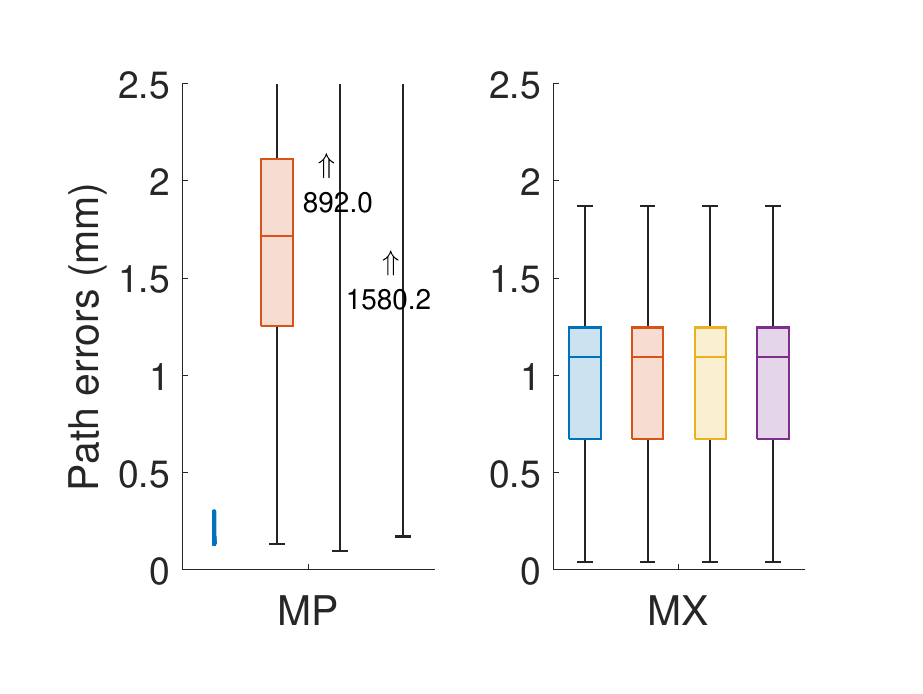} 
  \caption{\footnotesize A-SNS / Zero noise / Path 1}
  \label{fig:7DoF-A-SNS-Circle}
\end{subfigure}
\caption{\footnotesize MP-GI versus MX-GI path errors while varying the units of the prismatic joint from $m$ to $mm$ for the WMVN, PID-PPP, V-SNS, MAN, FPBM and A-SNS schemes applied to the interlaced circle trajectory (path 1) of  the 7DoF (2RP4R) manipulator. When a value exists next to a box plot, it indicates the maximum value (in $mm$) of the error distribution that has been zoomed in for better visualization.}
\label{fig:scheme-comparison-7DoF-Circle}
%\vspace{-4mm}
\end{figure}

\begin{figure}[!t]
\centering
\begin{subfigure}[b]{.24\textwidth}
  \centering
  % include first image
  \includegraphics[width=\textwidth,trim={0cm 0.9cm 0cm 1cm},clip]{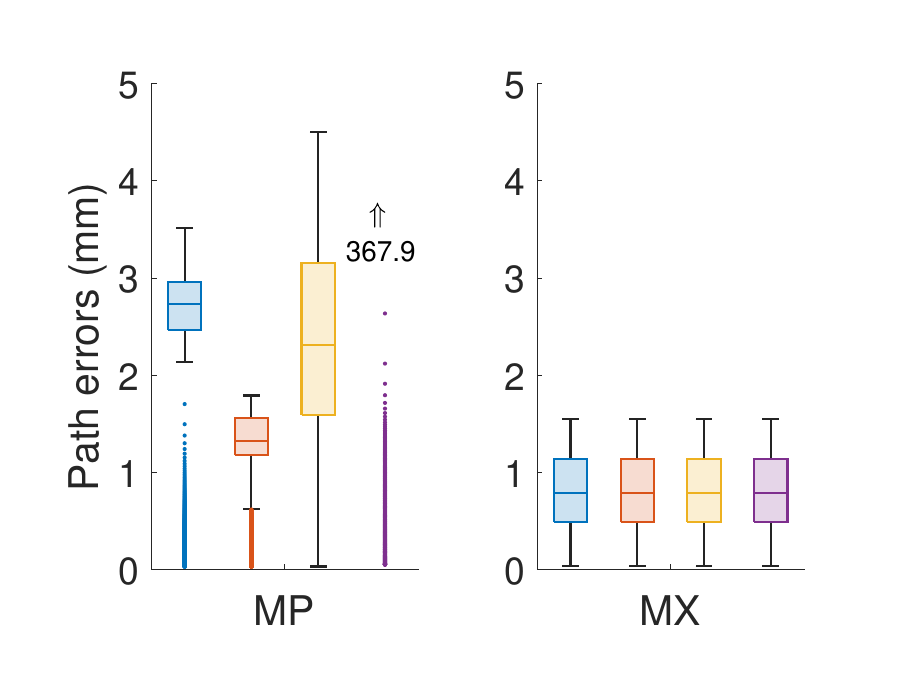} 
  \caption{\footnotesize WMVN / Zero noise / Path 2}
  \label{fig:7DoF-MVN-Rhodonea}
\end{subfigure}
\begin{subfigure}[b]{.24\textwidth}
  \centering
  % include first image
  \includegraphics[width=\textwidth,trim={0cm 0.9cm 0cm 1cm},clip]{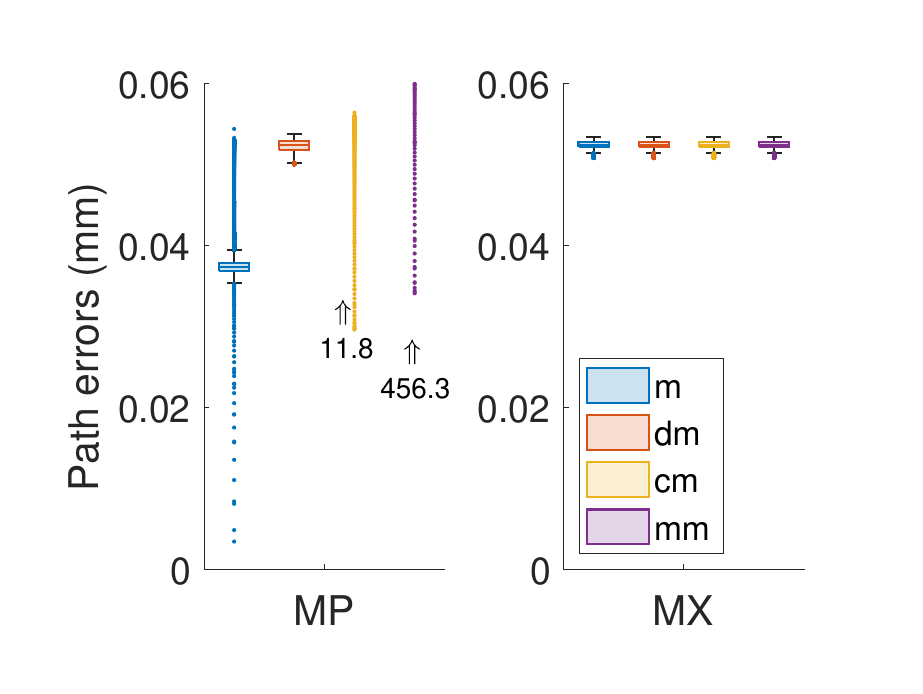} 
  \caption{\footnotesize MAN / Zero noise / Path 2}
  \label{fig:7DoF-MAN-Rhodonea}
\end{subfigure}
\begin{subfigure}[b]{.24\textwidth}
  \centering
  % include third image
  \includegraphics[width=\textwidth,trim={0cm 0.9cm 0cm 1cm},clip]{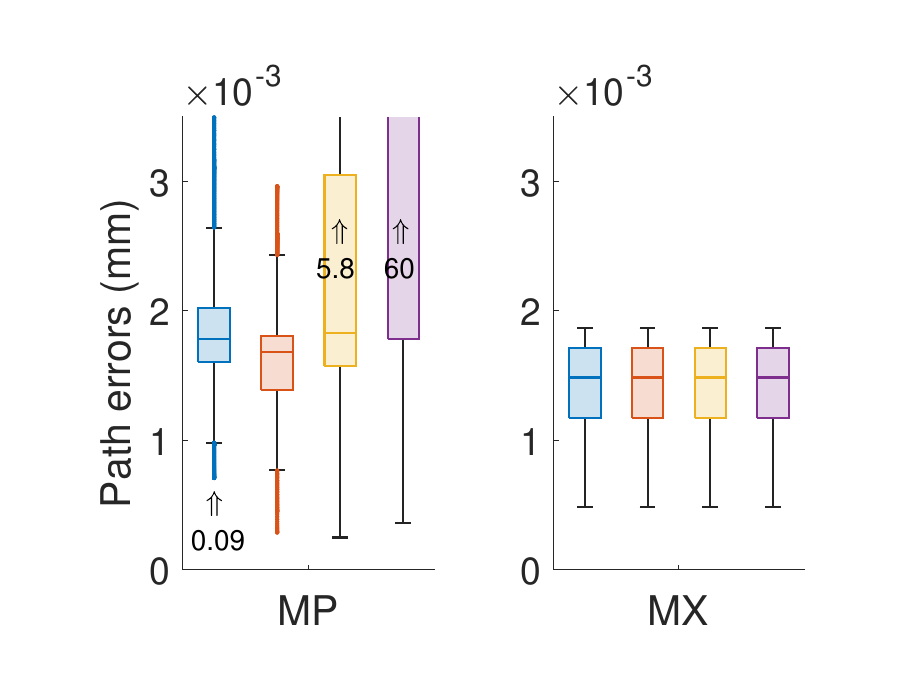} 
  \caption{\footnotesize PID-PPP / Zero noise / Path 2}
  \label{fig:7DoF-PID-PPP-Rhodonea}
\end{subfigure}
\begin{subfigure}[b]{.24\textwidth}
  \centering
  % include third image
  \includegraphics[width=\textwidth,trim={0cm 0.9cm 0cm 1cm},clip]{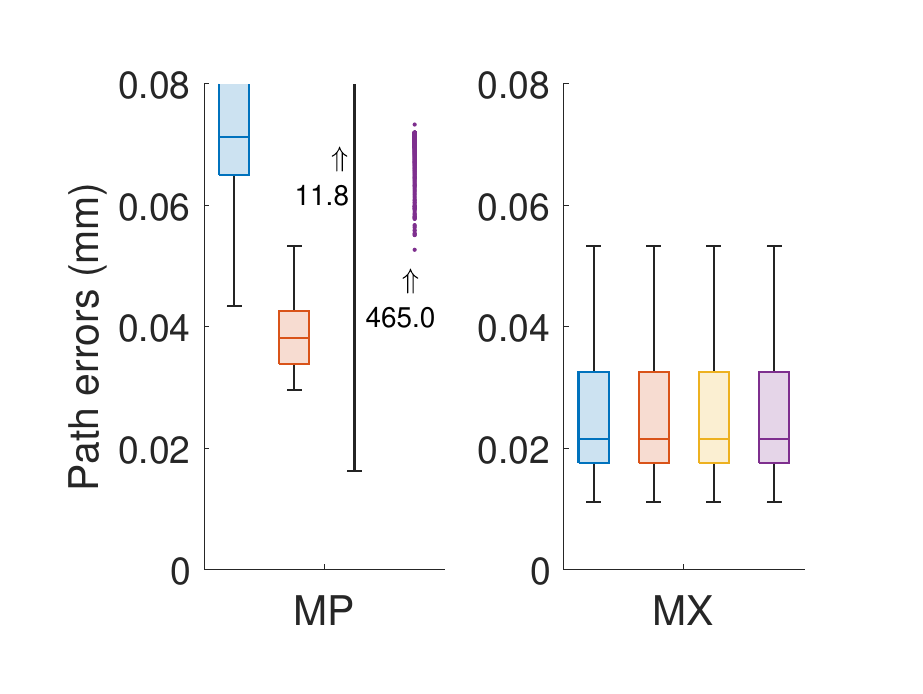} 
  \caption{\footnotesize FPBM / Zero noise / Path 2}
  \label{fig:7DoF-FPBM-Rhodonea}
\end{subfigure}
\begin{subfigure}[b]{.24\textwidth}
  \centering
  % include second image
  \includegraphics[width=\textwidth,trim={0cm 0.9cm 0cm 1cm},clip]{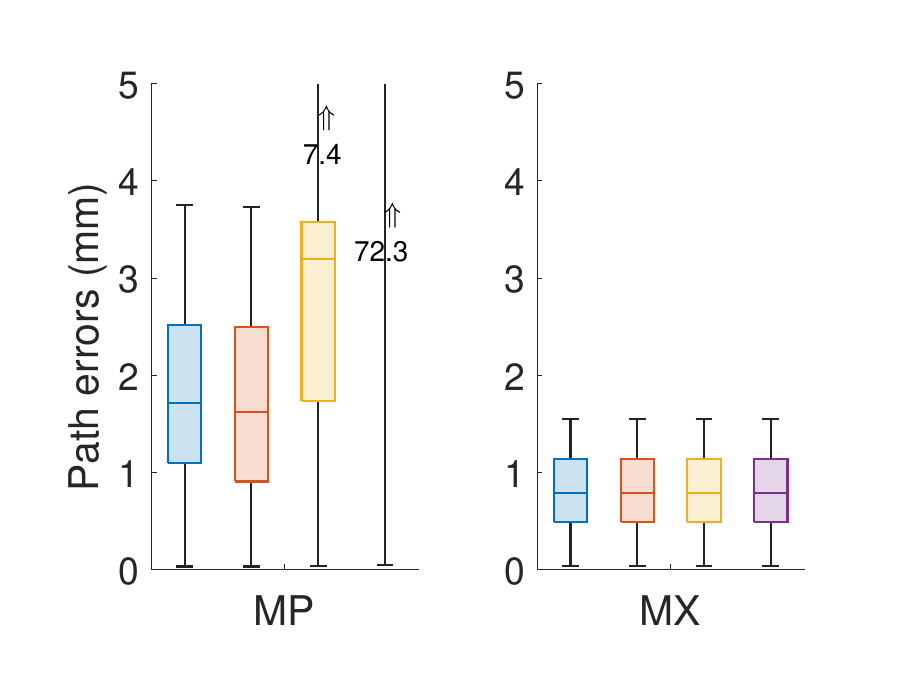} 
  \caption{\footnotesize V-SNS / Zero noise / Path 2}
  \label{fig:7DoF-V-SNS-Rhodonea}
\end{subfigure}
\begin{subfigure}[b]{.24\textwidth}
  \centering
  % include second image
  \includegraphics[width=\textwidth,trim={0cm 0.9cm 0cm 1cm},clip]{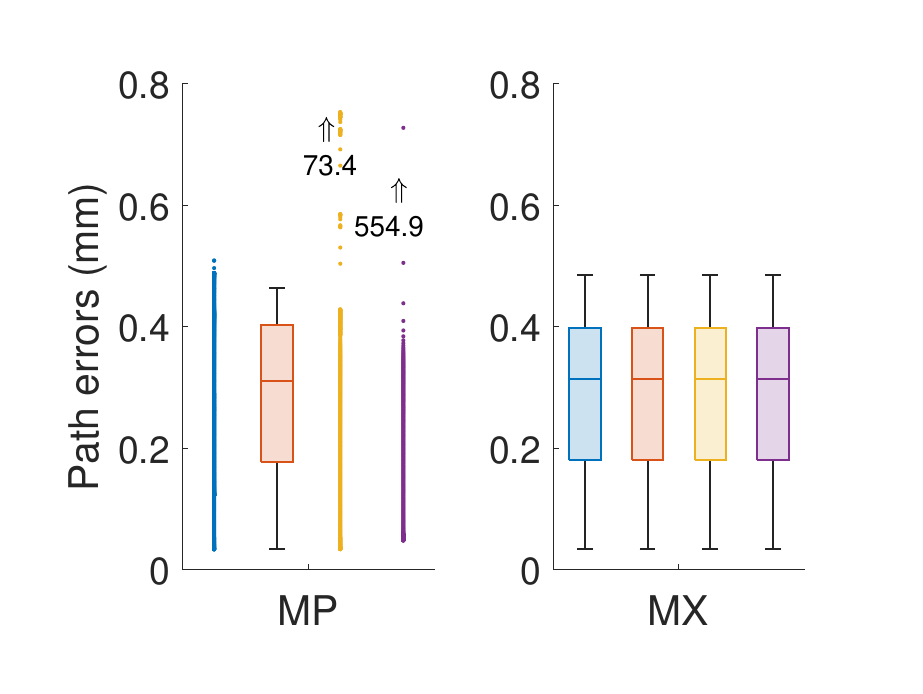} 
  \caption{\footnotesize A-SNS / Zero noise / Path 2}
  \label{fig:7DoF-A-SNS-Rhodonea}
\end{subfigure}
\caption{\footnotesize MP-GI versus MX-GI path errors while varying the units of the prismatic joint from $m$ to $mm$ for the WMVN, PID-PPP, V-SNS, MAN, FPBM and A-SNS schemes applied to the 3-dimensional rhodonea trajectory (path 2) of  the 7DoF (2RP4R) manipulator. When a value exists next to a box plot, it indicates the maximum value (in $mm$) of the error distribution that has been zoomed in for better visualization.}
\label{fig:scheme-comparison-7DoF-Rhodonea}
%\vspace{-4mm}
\end{figure}

\begin{figure}[!t]
\centering
\begin{subfigure}[b]{.24\textwidth}
  \centering
  % include first image
  \includegraphics[width=\textwidth,trim={0cm 0.9cm 0cm 1cm},clip]{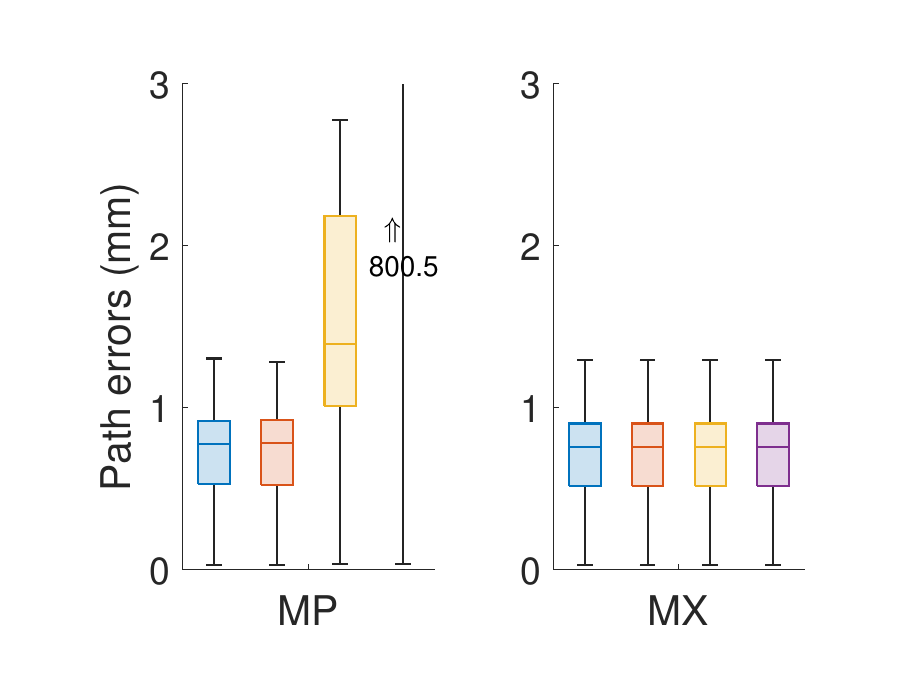} 
  \caption{\footnotesize WMVN / Zero noise / Path 3}
  \label{fig:7DoF-MVN-Tricuspid}
\end{subfigure}
\begin{subfigure}[b]{.24\textwidth}
  \centering
  % include first image
  \includegraphics[width=\textwidth,trim={0cm 0.9cm 0cm 1cm},clip]{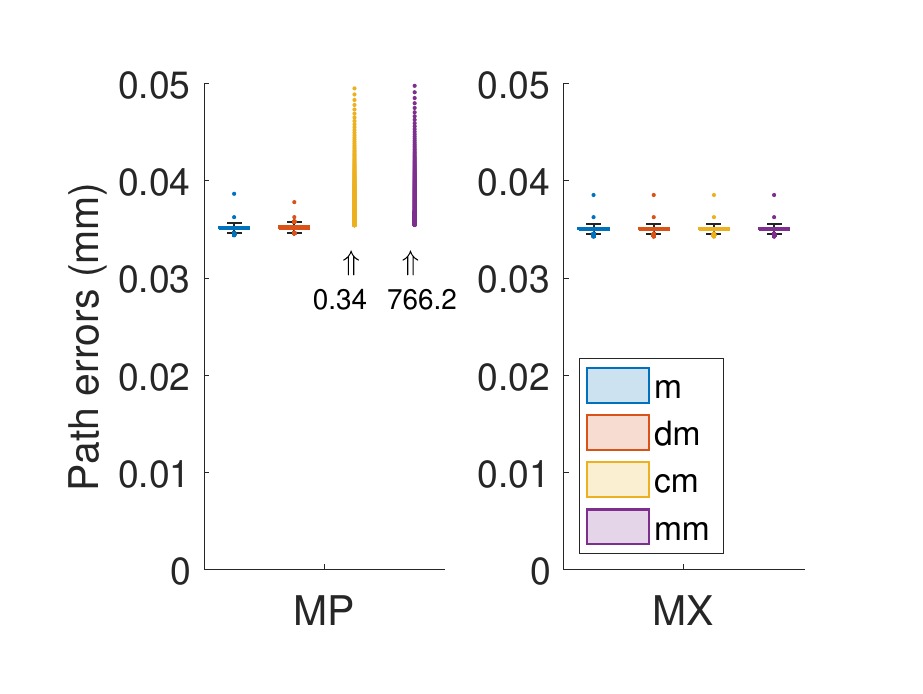} 
  \caption{\footnotesize MAN / Zero noise / Path 3}
  \label{fig:7DoF-MAN-Tricuspid}
\end{subfigure}
\begin{subfigure}[b]{.24\textwidth}
  \centering
  % include third image
  \includegraphics[width=\textwidth,trim={0cm 0.9cm 0cm 1cm},clip]{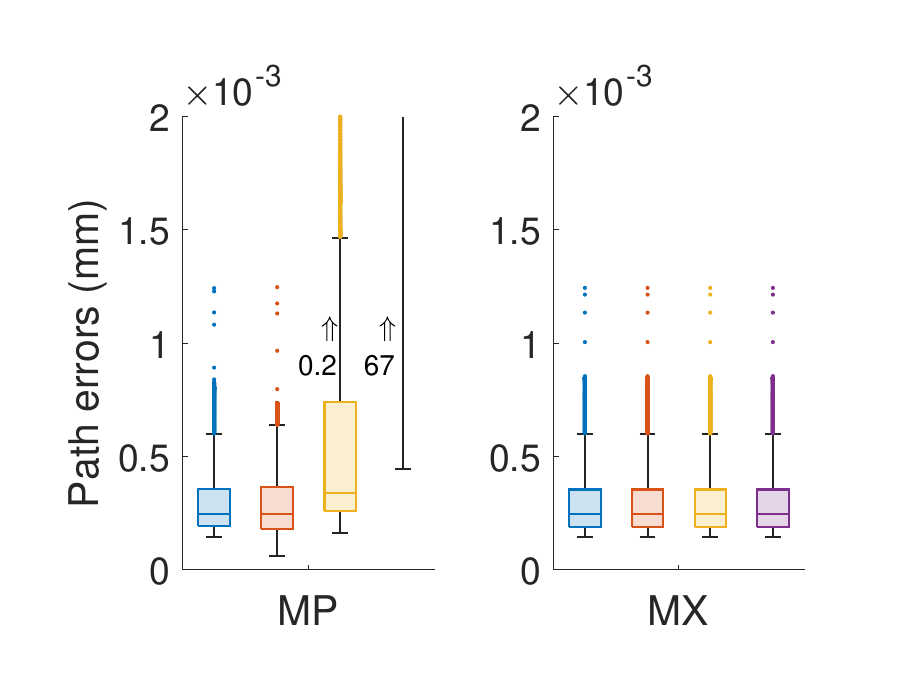} 
  \caption{\footnotesize PID-PPP / Zero noise / Path 3}
  \label{fig:7DoF-PID-PPP-Tricuspid}
\end{subfigure}
\begin{subfigure}[b]{.24\textwidth}
  \centering
  % include third image
  \includegraphics[width=\textwidth,trim={0cm 0.9cm 0cm 1cm},clip]{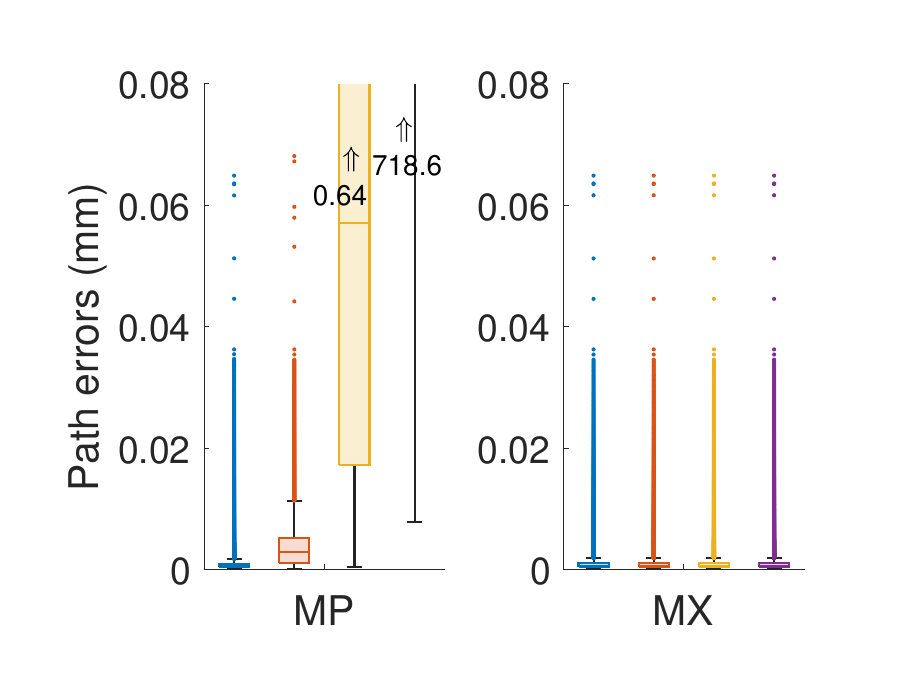} 
  \caption{\footnotesize FPBM / Zero noise / Path 3}
  \label{fig:7DoF-FPBM-Tricuspid}
\end{subfigure}
\begin{subfigure}[b]{.24\textwidth}
  \centering
  % include second image
  \includegraphics[width=\textwidth,trim={0cm 0.9cm 0cm 1cm},clip]{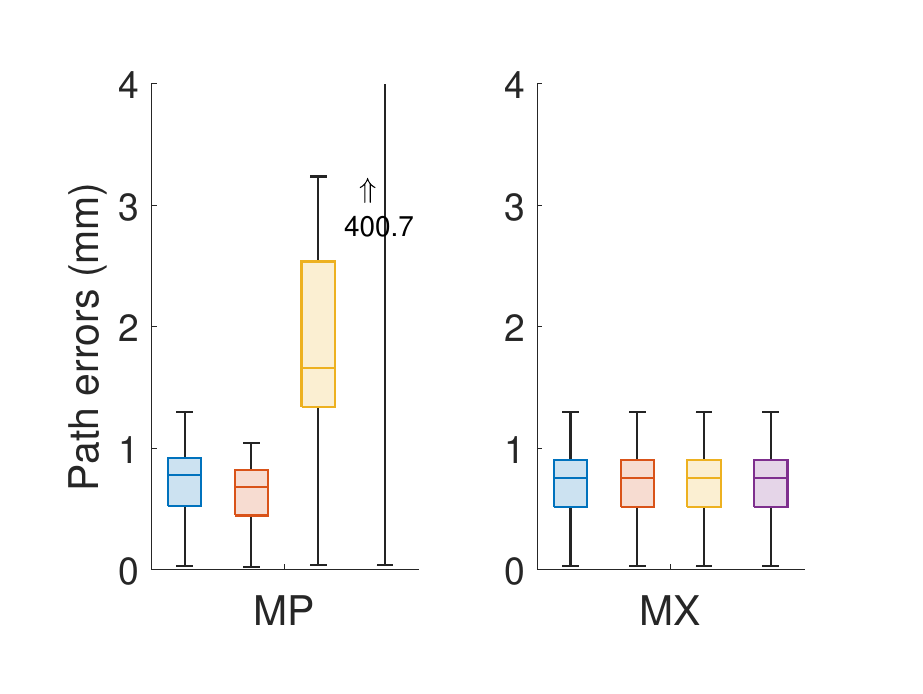} 
  \caption{\footnotesize V-SNS / Zero noise / Path 3}
  \label{fig:7DoF-V-SNS-Tricuspid}
\end{subfigure}
\begin{subfigure}[b]{.24\textwidth}
  \centering
  % include second image
  \includegraphics[width=\textwidth,trim={0cm 0.9cm 0cm 1cm},clip]{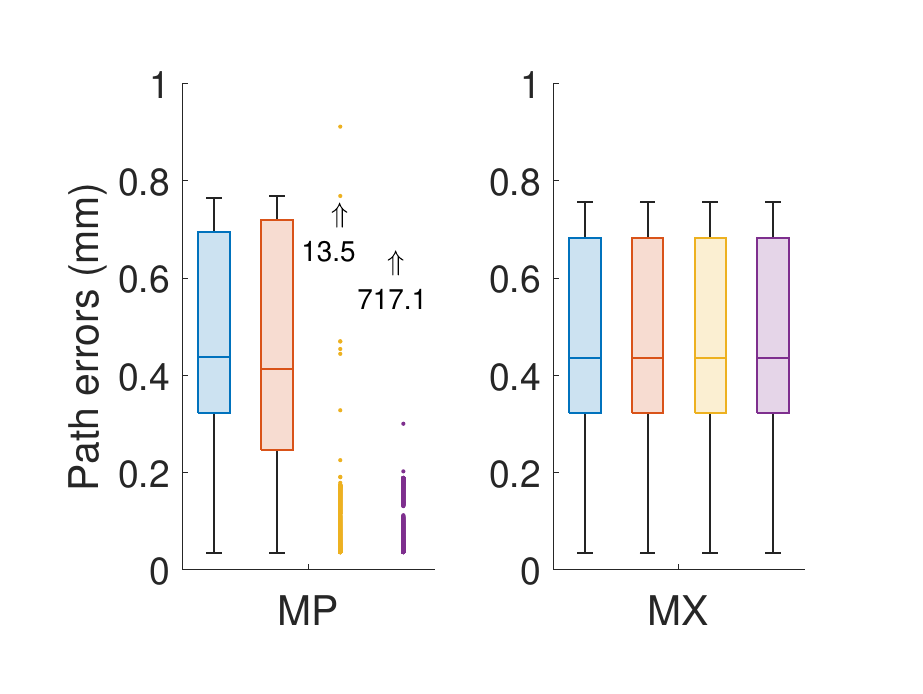} 
  \caption{\footnotesize A-SNS / Zero noise / Path 3}
  \label{fig:7DoF-A-SNS-Tricuspid}
\end{subfigure}
\caption{\footnotesize MP-GI versus MX-GI path errors while varying the units of the prismatic joint from $m$ to $mm$ for the WMVN, PID-PPP, V-SNS, MAN, FPBM and A-SNS schemes applied to the bent tricuspid trajectory (path 3) of  the 7DoF (2RP4R) manipulator. When a value exists next to a box plot, it indicates the maximum value (in $mm$) of the error distribution that has been zoomed in for better visualization.}
\label{fig:scheme-comparison-7DoF-Tricuspid}
%\vspace{-4mm}
\end{figure}

\begin{figure*}[!t]
\centering
\begin{subfigure}{.245\textwidth}
  \centering
  % include  image
  \includegraphics[width=\textwidth,trim={0.8cm 0.8cm 0.8cm 0.8cm},clip]{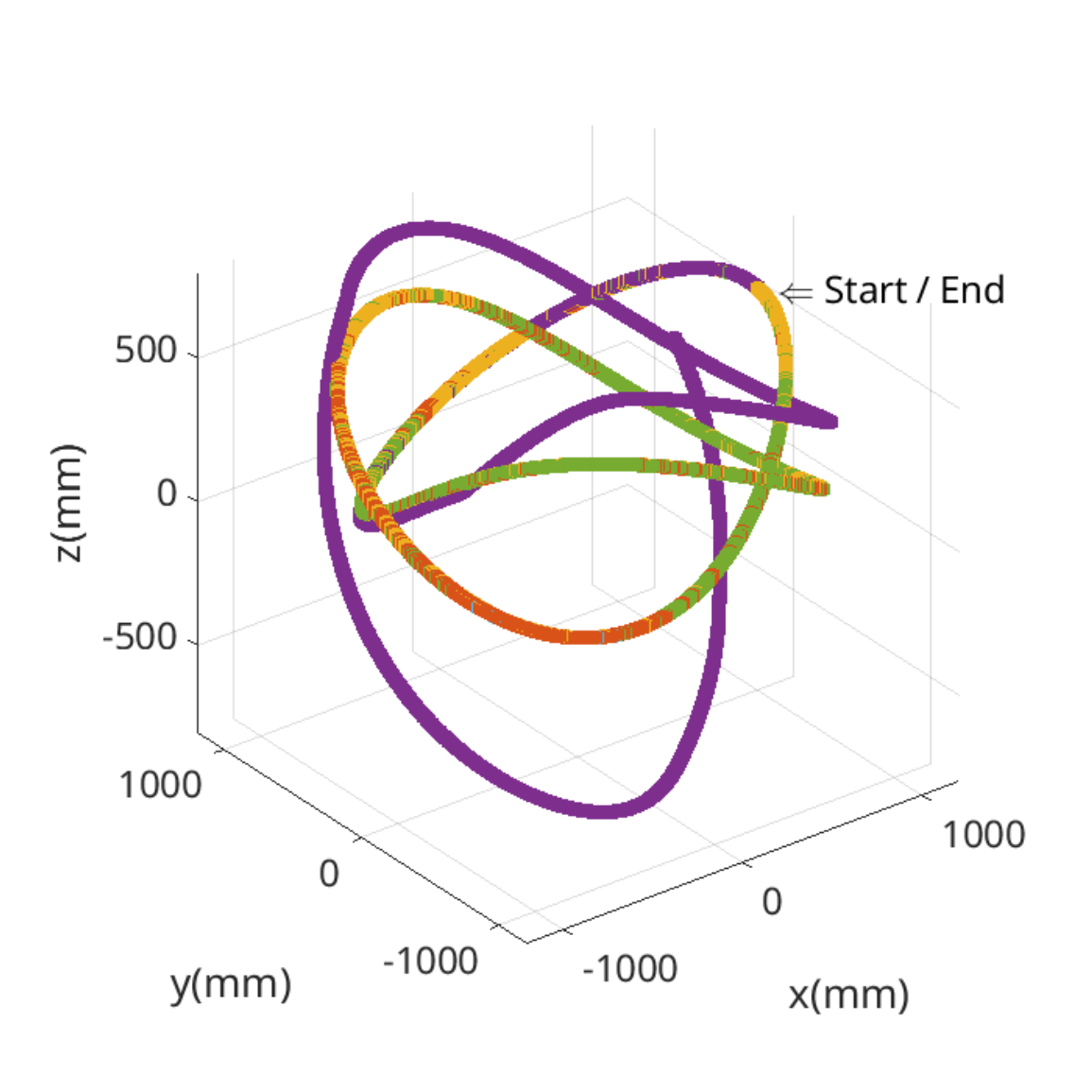}  
  \caption{\scriptsize WMVN / MP / Path 1 / Zero noise}
  \label{fig:7dof-MP-WMVN-Circle}
\end{subfigure}
%\hspace{5mm}
\begin{subfigure}{.245\textwidth}
  \centering
  % include  image
  \includegraphics[width=\textwidth,trim={0.8cm 0.8cm 0.8cm 0.8cm},clip]{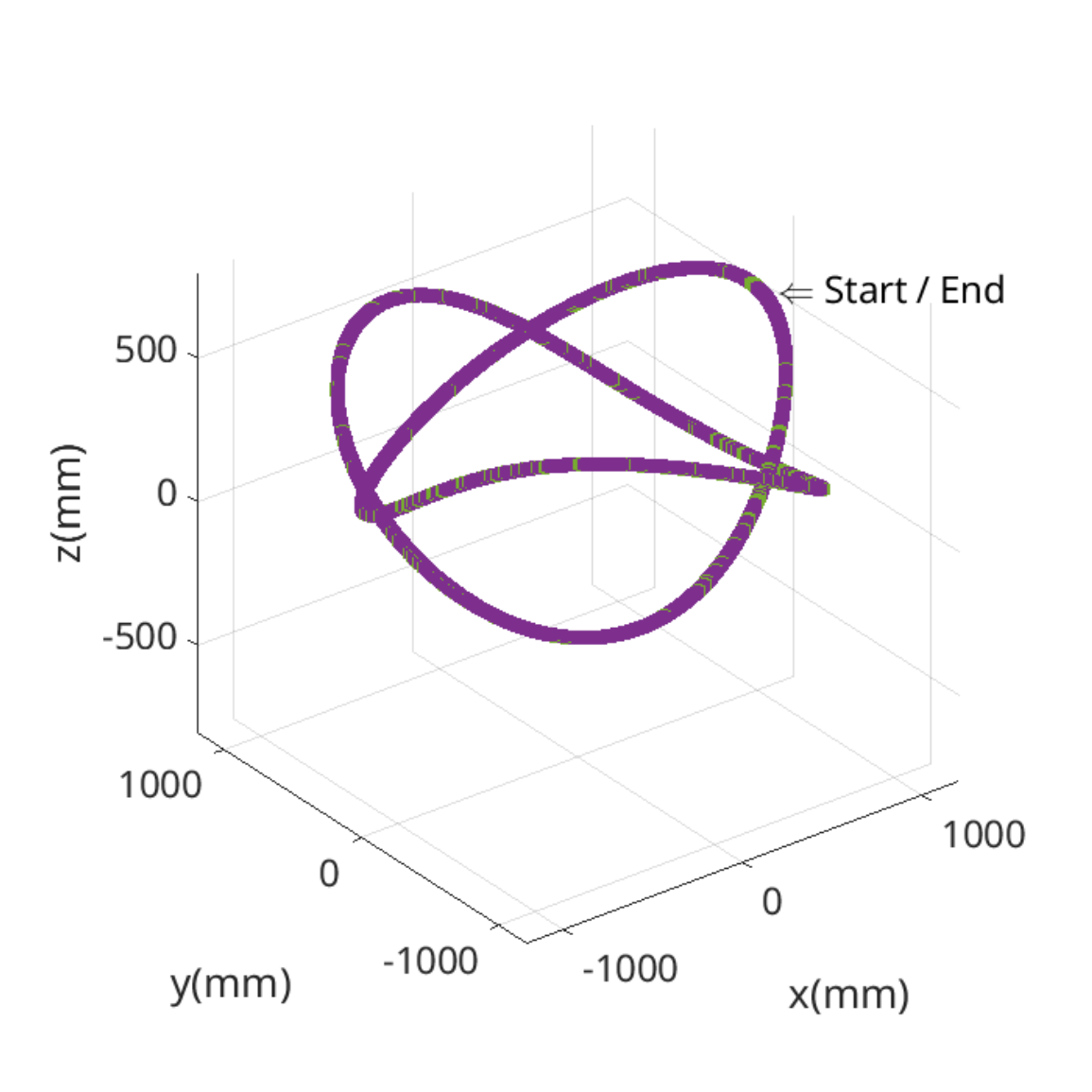}  
  \caption{\scriptsize WMVN / MX / Path 1 / Zero noise}
  \label{fig:7dof-MX-WMVN-Circle}
\end{subfigure}
\begin{subfigure}{.245\textwidth}
  \centering
  % include  image
  \includegraphics[width=\textwidth,trim={0.8cm 0.8cm 0.8cm 0.8cm},clip]{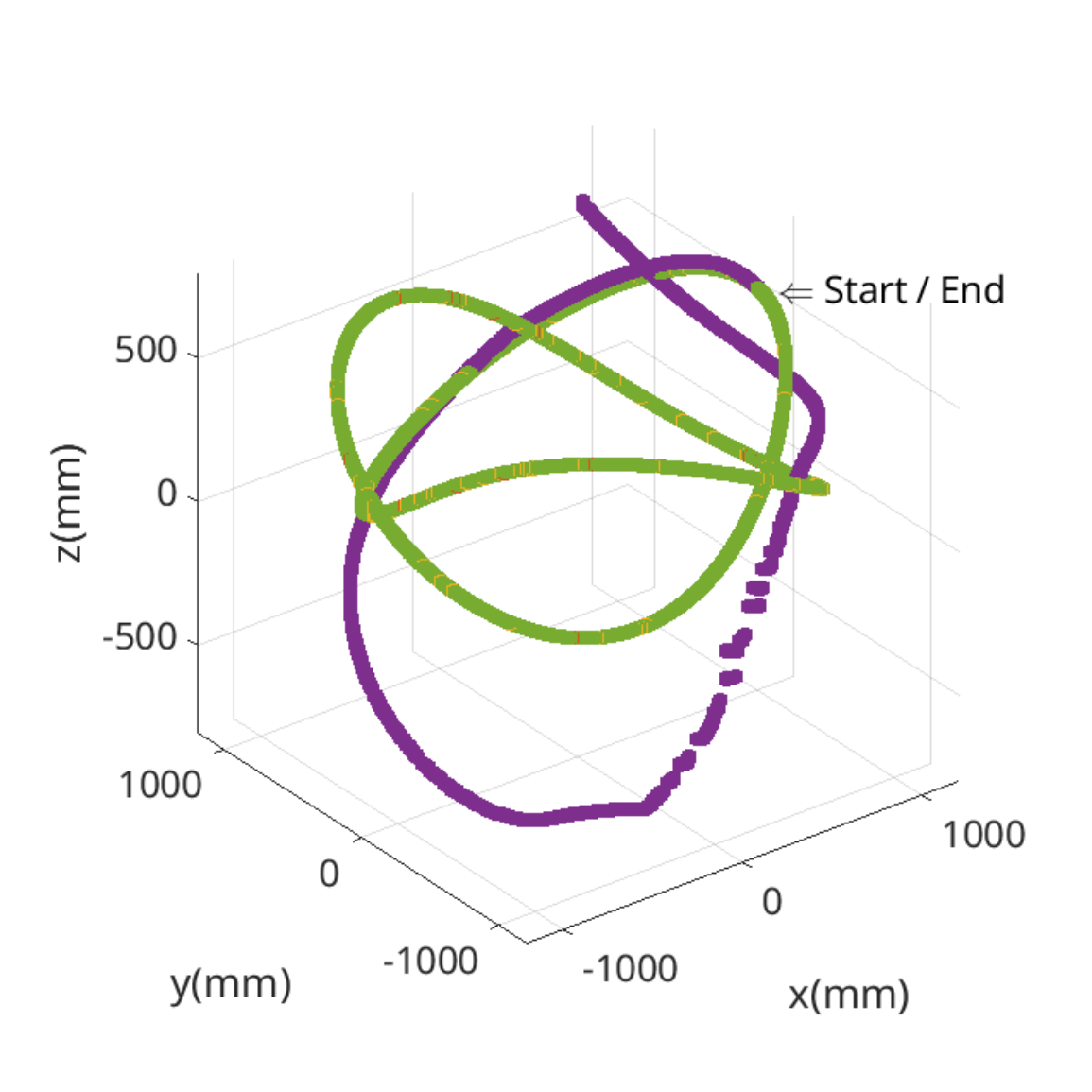}  
  \caption{\scriptsize MAN / MP / Path 1 / Zero noise}
  \label{fig:7dof-MP-MAN-Circle}
\end{subfigure}
%\hspace{5mm}
\begin{subfigure}{.245\textwidth}
  \centering
  % include  image
  \includegraphics[width=\textwidth,trim={0.8cm 0.8cm 0.8cm 0.8cm},clip]{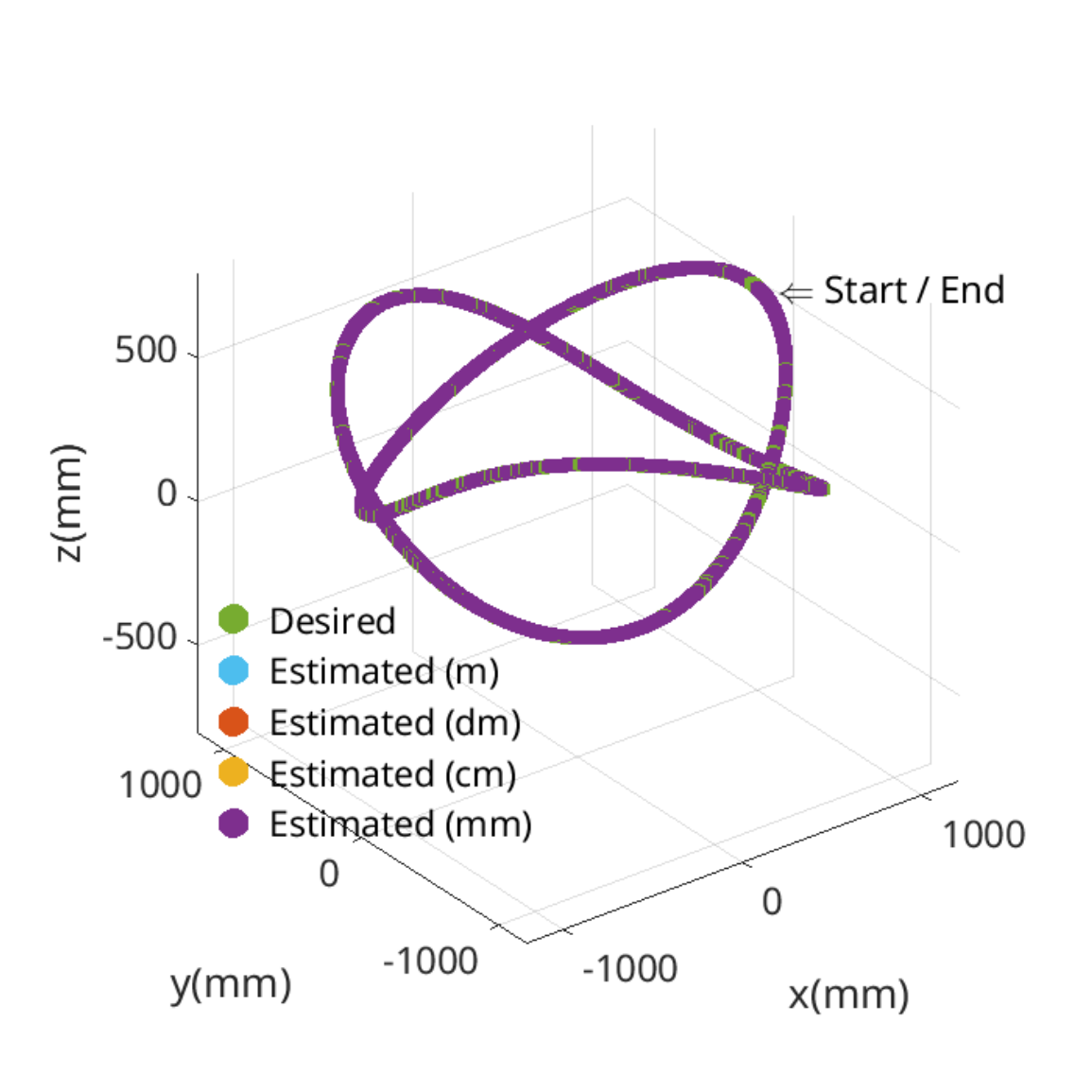}  
  \caption{\scriptsize MAN / MX / Path 1 / Zero noise}
  \label{fig:7dof-MX-MAN-Circle}
\end{subfigure}
\begin{subfigure}{.245\textwidth}
  \centering
  % include  image
  \includegraphics[width=\textwidth,trim={0.8cm 0.8cm 0.8cm 0.8cm},clip]{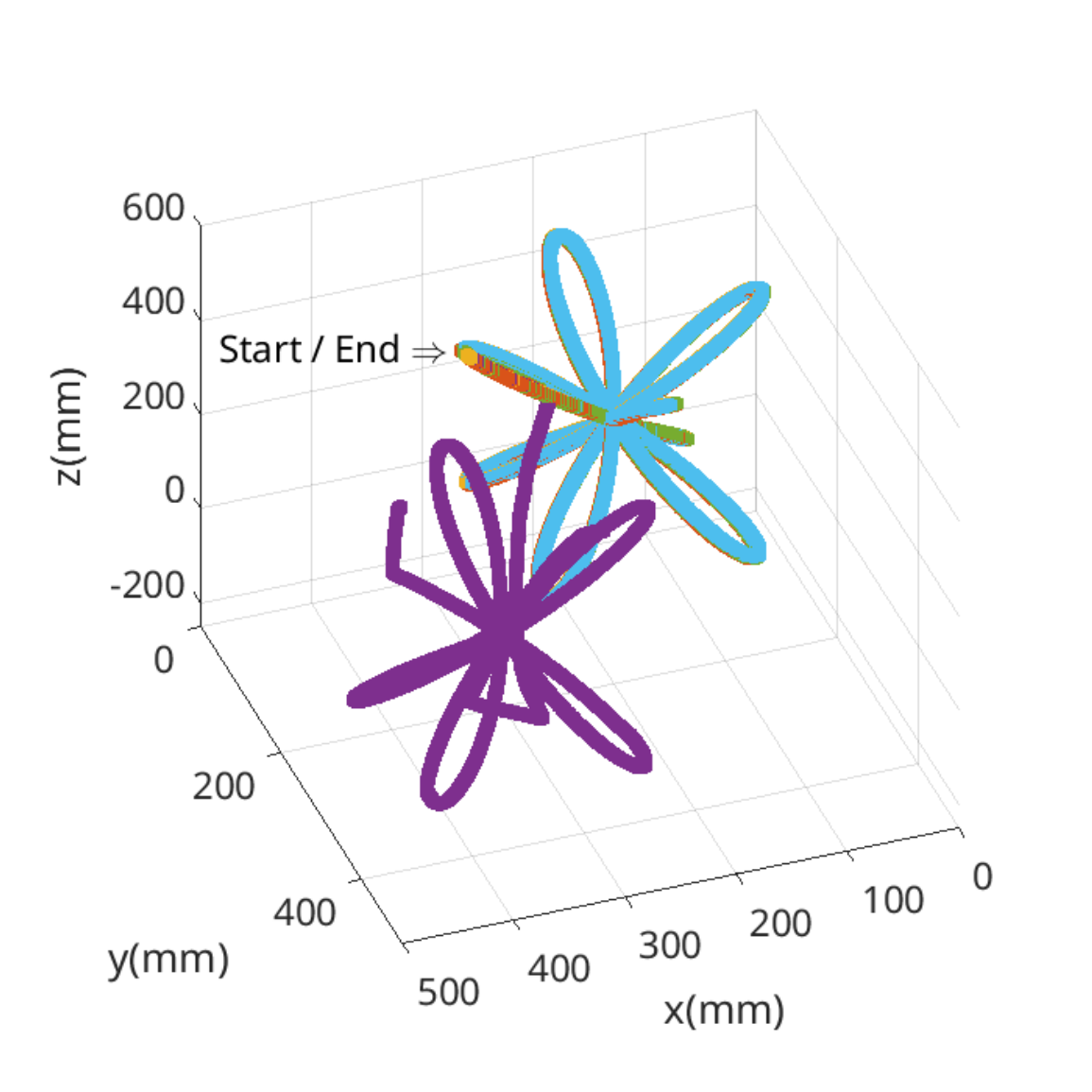}  
  \caption{\scriptsize WMVN / MP / Path 2 / Zero noise}
  \label{fig:7dof-MP-WMVN-Rhodonea}
\end{subfigure}
%\hspace{5mm}
\begin{subfigure}{.245\textwidth}
  \centering
  % include  image
  \includegraphics[width=\textwidth,trim={0.8cm 0.8cm 0.8cm 0.8cm},clip]{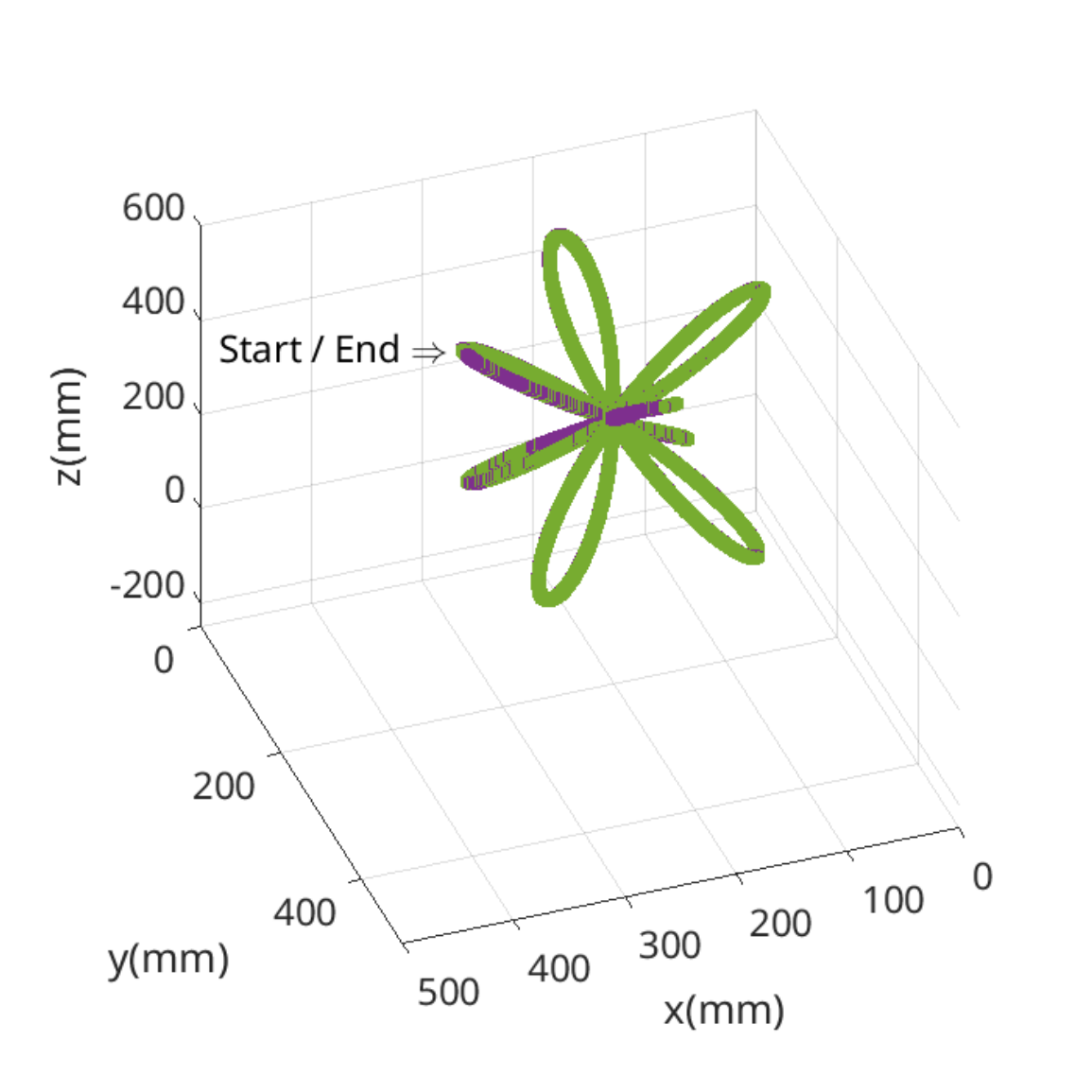}  
  \caption{\scriptsize WMVN / MX / Path 2 / Zero noise}
  \label{fig:7dof-MX-WMVN-Rhodonea}
\end{subfigure}
%\hspace{5mm}
\begin{subfigure}{.245\textwidth}
  \centering
  % include  image
  \includegraphics[width=\textwidth,trim={0.8cm 0.8cm 0.8cm 0.8cm},clip]{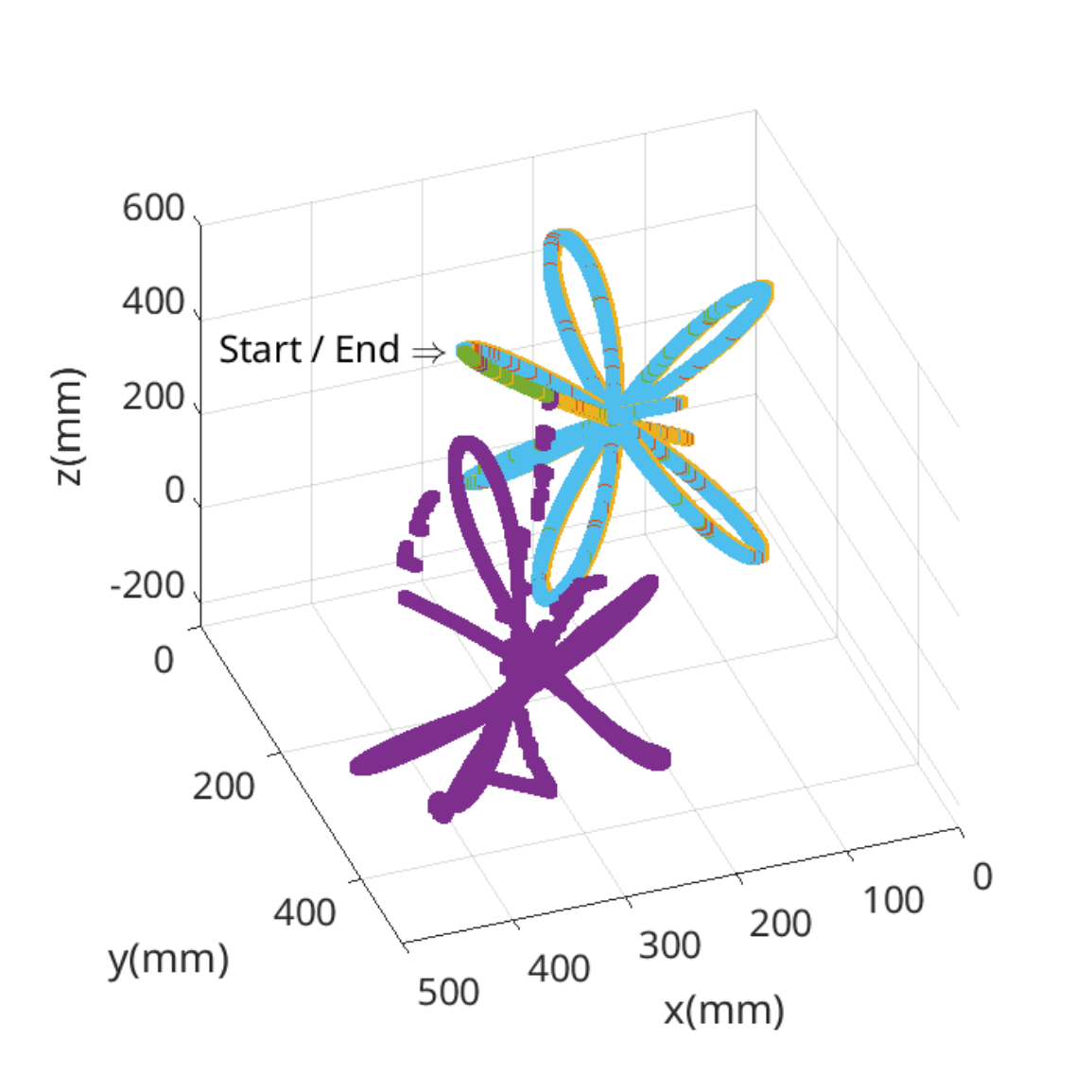}  
  \caption{\scriptsize MAN / MP / Path 2 / Zero noise}
  \label{fig:7dof-MP-MAN-Rhodonea}
\end{subfigure}
%\hspace{5mm}
\begin{subfigure}{.245\textwidth}
  \centering
  % include  image
  \includegraphics[width=\textwidth,trim={0.8cm 0.8cm 0.8cm 0.8cm},clip]{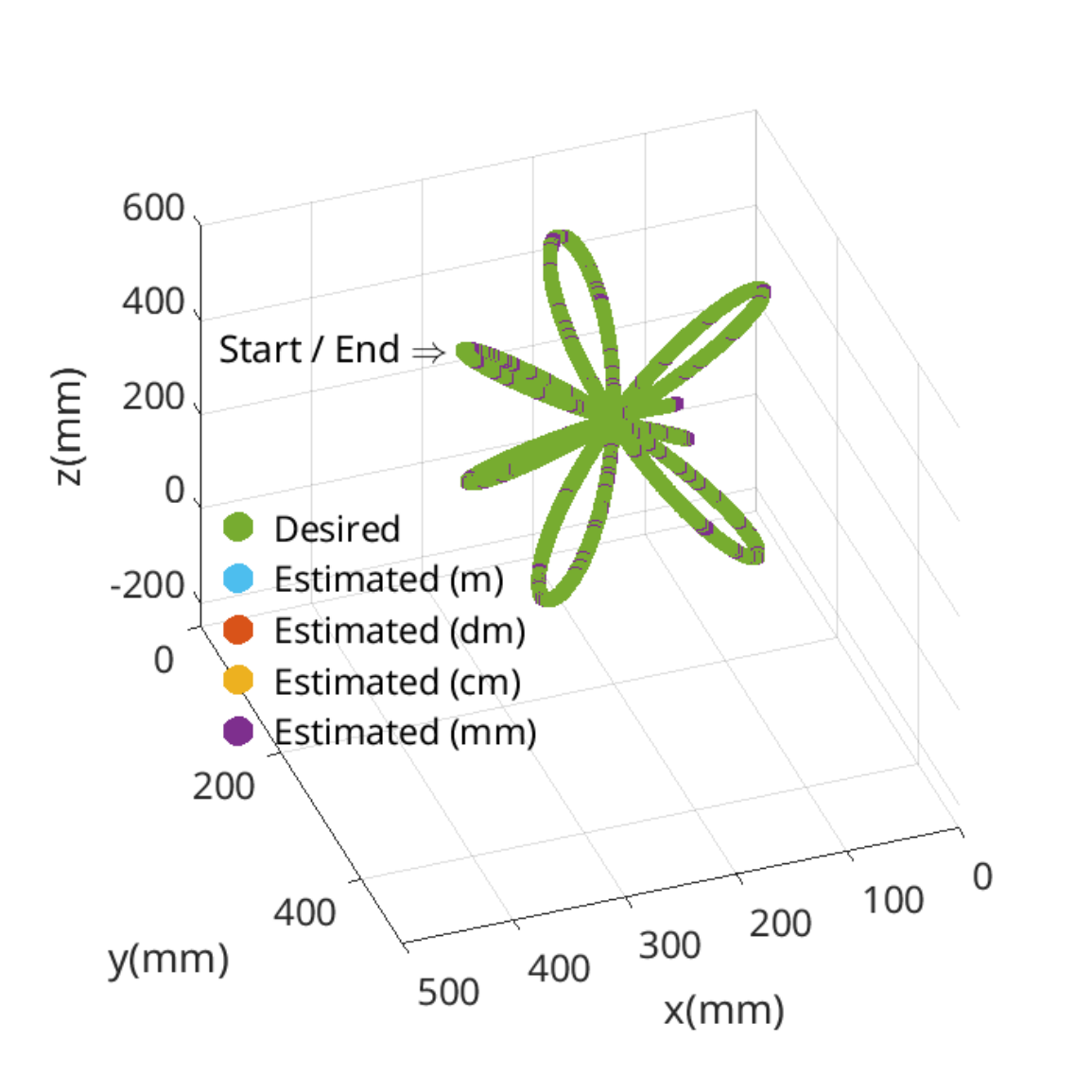}  
  \caption{\scriptsize MAN / MX / Path 2 / Zero noise}
  \label{fig:7dof-MX-MAN-Rhodonea}
\end{subfigure}
\begin{subfigure}{.245\textwidth}
  \centering
  % include  image
  \includegraphics[width=\textwidth,trim={0.8cm 0.8cm 0.8cm 0.8cm},clip]{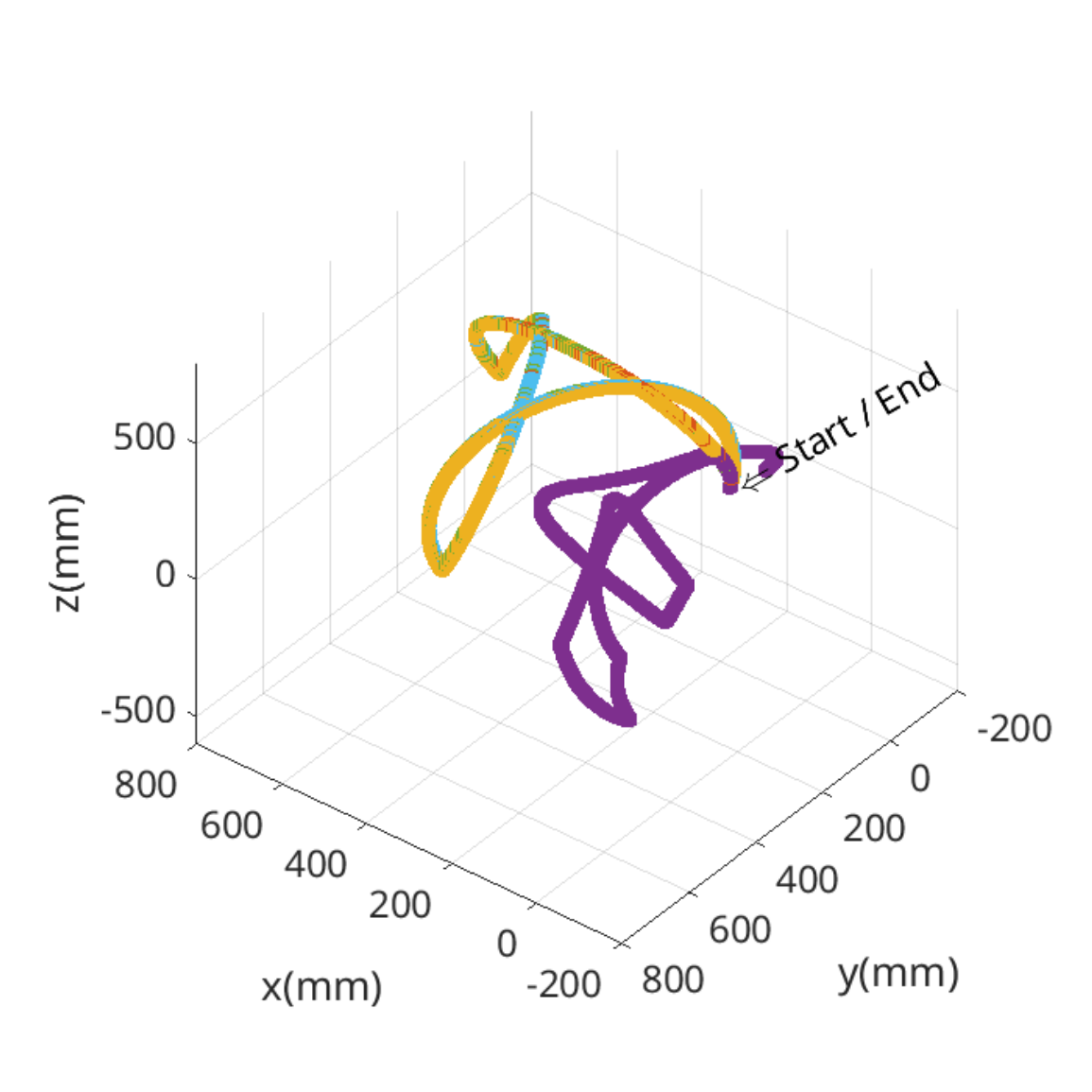}  
  \caption{\scriptsize WMVN / MP / Path 3 / Zero noise}
  \label{fig:7dof-MP-WMVN-Tricuspid}
\end{subfigure}
%\hspace{5mm}
\begin{subfigure}{.245\textwidth}
  \centering
  % include  image
  \includegraphics[width=\textwidth,trim={0.8cm 0.8cm 0.8cm 0.8cm},clip]{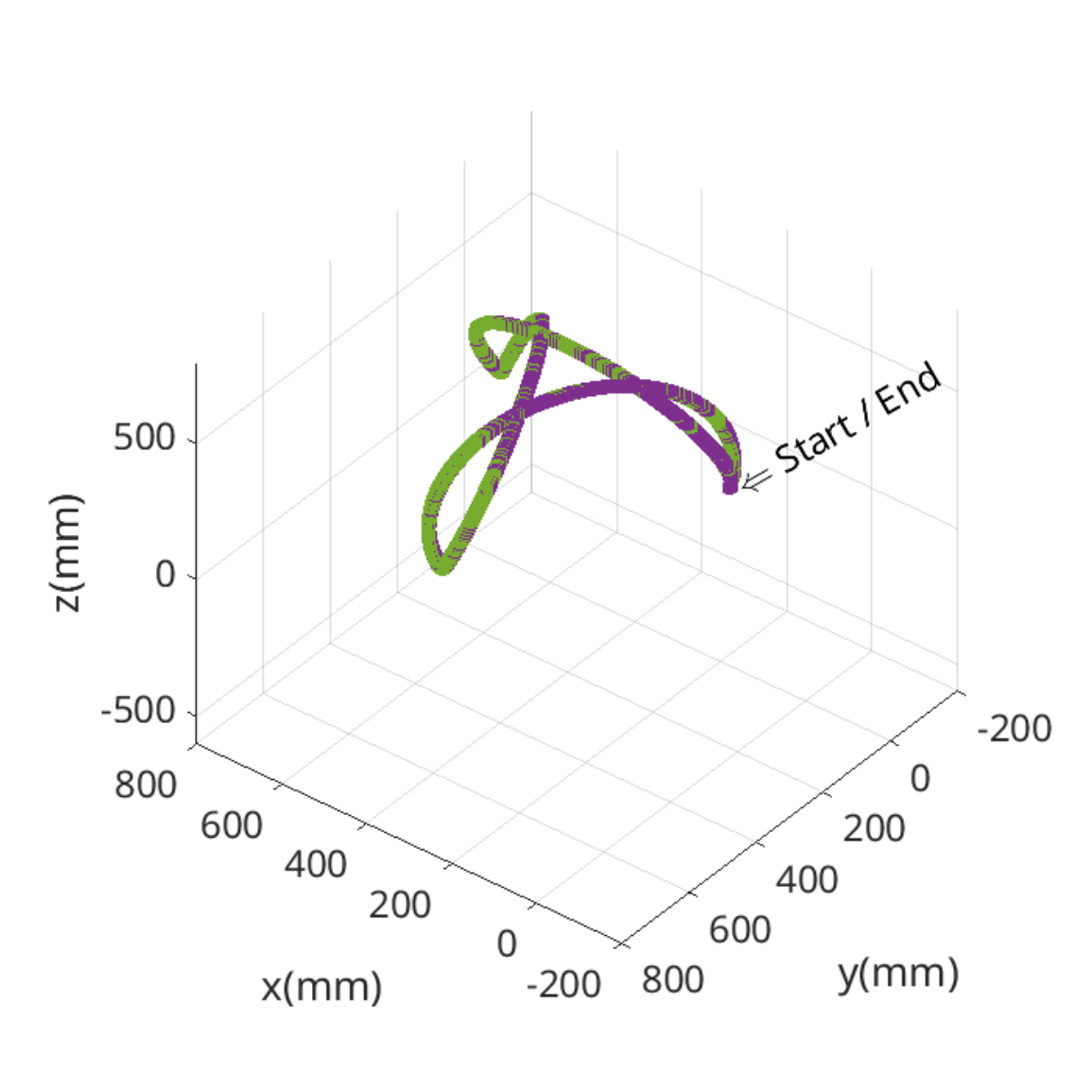}  
  \caption{\scriptsize WMVN / MX / Path 3 / Zero noise}
  \label{fig:7dof-MX-WMVN-Tricuspid}
\end{subfigure}
%\hspace{5mm}
\begin{subfigure}{.245\textwidth}
  \centering
  % include  image
  \includegraphics[width=\textwidth,trim={0.8cm 0.8cm 0.8cm 0.8cm},clip]{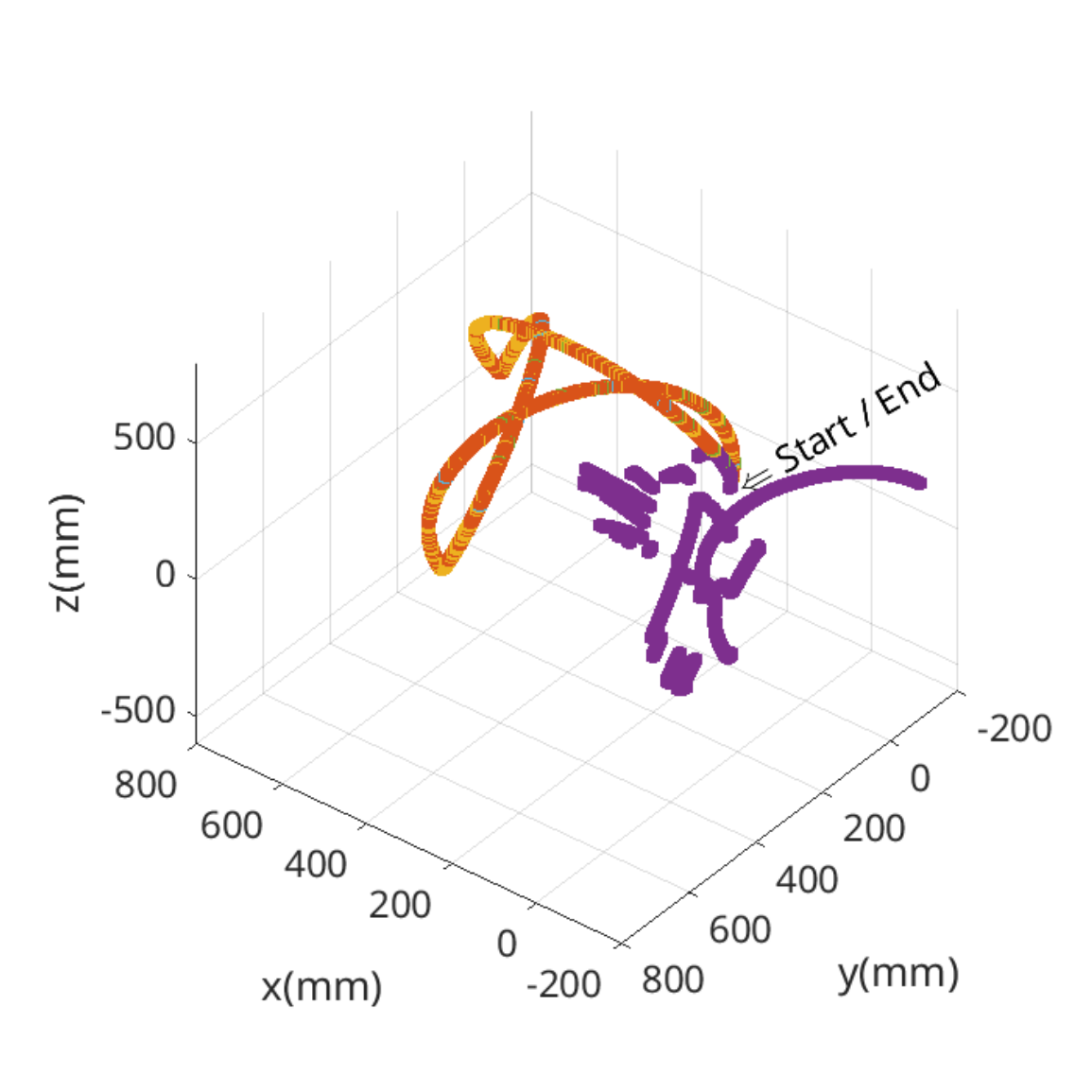}  
  \caption{\scriptsize MAN / MP / Path 3 / Zero noise}
  \label{fig:7dof-MP-MAN-Tricuspid}
\end{subfigure}
%\hspace{5mm}
\begin{subfigure}{.245\textwidth}
  \centering
  % include  image
  \includegraphics[width=\textwidth,trim={0.8cm 0.8cm 0.8cm 0.8cm},clip]{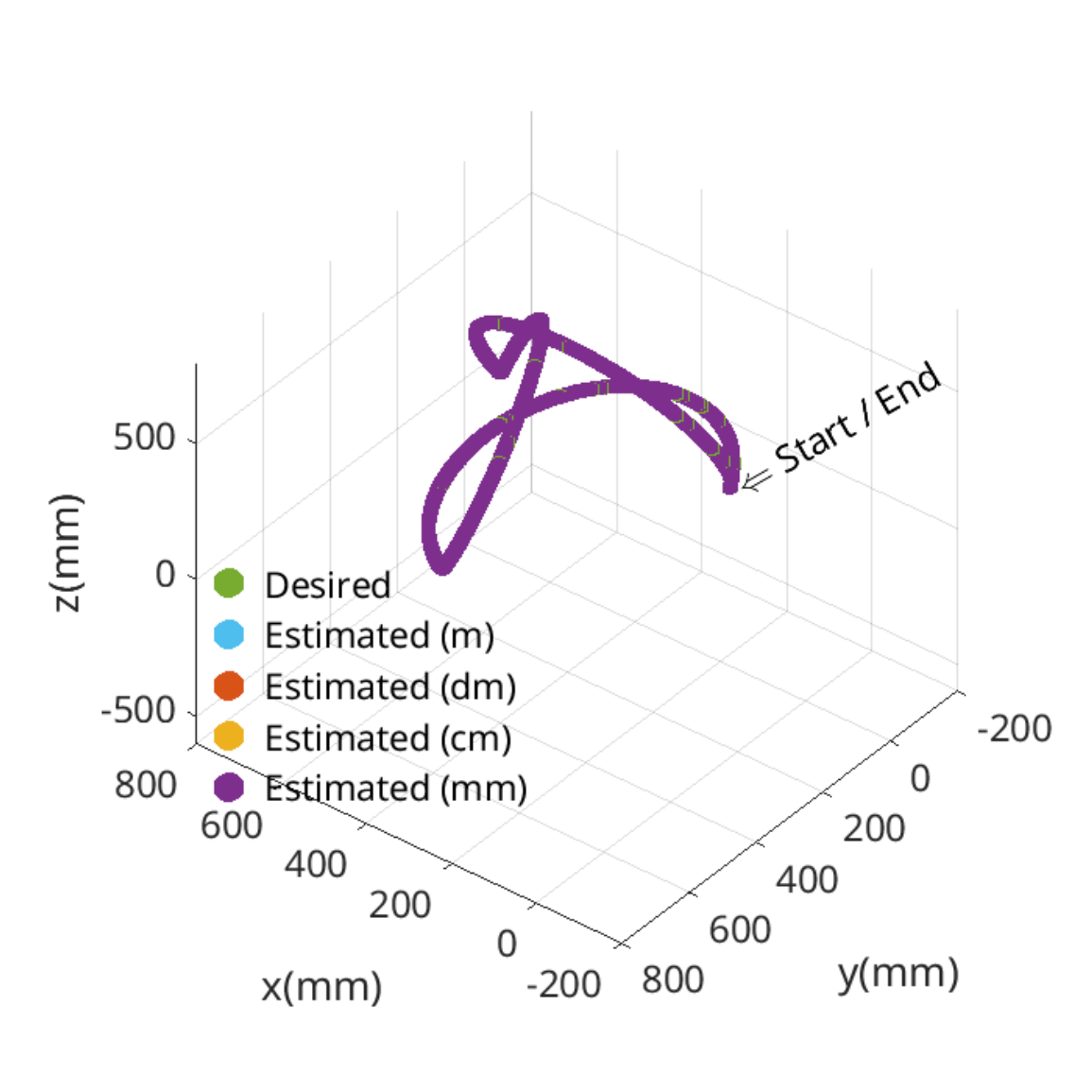}  
  \caption{\scriptsize MAN / MX / Path 3 / Zero noise}
  \label{fig:7dof-MX-MAN-Tricuspid}
\end{subfigure}
\caption{\footnotesize Desired path versus estimated paths while varying the units of the prismatic joint from $m$, $dm$, $cm$, to $mm$ for the 7DoF (2RP4R) using respectively the WMVN and MAN with both the MP and MX-GI's. Similar behaviors are observed with the other schemes.}
\label{fig:path-planning-7DoF}
\vspace{-4mm}
\end{figure*}

\subsection{In the case of the 7DoF-2RP4R redundant robot}
From rows 5 to 10 on Table \ref{tab:schemes-units-7DoFP-zero}, we observe that, except for one scheme, all investigated schemes were able to achieve less than $1mm$ average errors when the unit chosen for the 7DoF robot is $m$, for both the MP- and MX-GI's, and in the absence of noise: i.e. the A-SNS scheme achieved about $1.66mm$ average error for the interlaced circle path. For the 3-dimensional rhodonea path, the PID-PPP, MAN and  FPBM schemes achieved less than $1mm$ average errors in the $m$ case, while the WMVN, the V-SNS and the A-SNS respectively achieved $2.58mm$, $1.84mm$ and $6.01mm$. For the bent tricuspid path, all the MP-GI-based schemes achieved less than $1mm$ under the $m$ case. However, when the units were set from $m$ to $mm$ for all the investigated paths, the average error was no longer consistent across all units for the MP-GI-based schemes. After applying the MX-GI, the behavior of the robot end-effector remained the same across all units. As depicted in Figures \ref{fig:scheme-comparison-7DoF-Circle}, \ref{fig:scheme-comparison-7DoF-Rhodonea}, and \ref{fig:scheme-comparison-7DoF-Tricuspid}, the box-plots of the errors vary with the choice of unit for the MP-GI-based schemes, while they are the same for MX-GI-based schemes. Figure \ref{fig:path-planning-7DoF} illustrates the trajectories followed under the WMVN and MAN schemes without noise contamination. In Figures \ref{fig:7dof-MP-WMVN-Circle}, \ref{fig:7dof-MP-WMVN-Rhodonea}, and \ref{fig:7dof-MP-WMVN-Tricuspid}, the end-effector behaviors exhibited by the robot are not consistent across all units for the MP-GI-based WMVN scheme. However, Figures \ref{fig:7dof-MX-WMVN-Circle}, \ref{fig:7dof-MX-WMVN-Rhodonea}, and \ref{fig:7dof-MX-WMVN-Tricuspid} show that the end-effector behavior is consistent across all units when the MX-GI is used. The same observation can be made for the MAN scheme based on Figures \ref{fig:7dof-MP-MAN-Circle}, \ref{fig:7dof-MP-MAN-Rhodonea}, and \ref{fig:7dof-MP-MAN-Tricuspid}, which clearly show that the MP-GI schemes produce undesired end-effector behaviors when the units are varied from $m$ to $mm$, while MX-GI based schemes remain consistent as depicted in Figures \ref{fig:7dof-MX-MAN-Circle}, \ref{fig:7dof-MX-MAN-Rhodonea}, and \ref{fig:7dof-MX-MAN-Tricuspid}. 

The schemes were also tested under constant, time-varying, and random noise contamination. Once again, here, $\delta(t)$ was set as following: (1) constant noise $\delta(t) = u*[0.3, 0.5, 0.3]^{T}$ for velocity-level schemes, and $\delta(t) = u*[0.03, 0.05, 0.03]^{T}$ for acceleration-level schemes; (2) time-varying noise $\delta(t) = u*[0.3*sin(2t), 0.5*cos(2t), 0.3*sin(2t)]^{T}$ for velocity-level schemes and $\delta(t) = u*[0.03*sin(2t), 0.05*cos(2t), 0.03*sin(2t)]^{T}$ for acceleration-level schemes; (3) random noise $\delta(t)= u* [r_{1}, r_{2}, \dots, r_{m}]^{T}  \in \mathbb{R}^{m}$ with each $r_{i}$ being a seeded random number between $[0,1]$; where $t$ is the time and $u$ is either $1$, $10$, $100$, or $1000$ according to the choice of $m$, $dm$, $cm$, or $mm$ as unit. From rows 13 to 16, 20 to 23, and 27 to 30 on Table \ref{tab:schemes-units-7DoFP-zero}, we notice that the WMVN, V-SNS, MAN, and A-SNS schemes also failed the trajectory following task depending on the choice of units for the MP-GI. Similar to what happened in \cite{guo2017new}, in the case of the MP-GI-based MAN and A-SNS for $cm$ and/or $mm$ units, the simulation completely failed without completing ("WC") the task because MATLAB reached its highest precision as the tracking error grew unbounded. However, the MX-GI-based MAN and A-SNS remained consistent across all units even though the error was high due to the intrinsic lack of noise-suppression capability of the scheme. Similarly to the 3DoF case, the feedback-based PID-PPP (rows 12, 19, 26) and FPBM (rows 17, 24, and 31) schemes succeeded in their tasks in the presence of noise and provided consistent results across all units when the MX-GI was employed. Here also, the feedback gains used with the 7DoF were set to the same values as in the 3DoF case.

\begin{table}[!t] 
\notsotinyone
\caption{\footnotesize Total number of GI calls in the implementation of the investigated Pseudo-inverse-based Path-Planning (PPP) and Repetitive Motion Planning (PRMP) schemes}\label{tab:total-GI-calls-surveyed-schemes-methodologies}
    \centering
    \begin{threeparttable}
         \setlength\tabcolsep{3.7pt}
         \begin{tabular}{|p{1.4mm}|p{8mm}|c|c|c|c|c|c|c|} 
         %\begin{tblr}{
         %       colspec = {|c|c|c|c|},
                %row{4} = {gray9},
                %row{12} = {gray9},
                %row{14} = {gray9},
                %row{22} = {gray9},
                %row{31} = {gray9},
                %row{32} = {gray9},
                %column{3} = {teal7},
                %cell{2}{3} = {yellow7},
        %      }
         \hline
         %\multirow{3}{3.5em}{Schemes} & \multicolumn{6}{c|}{Zero Noise} \\
         %\cline{2-7} 
         1 & \multirow{2}{3.5em}{Schemes} & \multirow{2}{3.5em}{GI Calls}  & \multicolumn{2}{c|}{3DoF Total GI Calls} & \multicolumn{2}{c|}{7DoF Total GI Calls}\\  
         \cline{4-7}
         2 & & & MP-GI & MX-GI & MP-GI & MX-GI\\
         %\cline{3-14} 
         %3 & & $m$ & $cm$ & $mm$ & $m$ & $cm$ & $mm$ & $m$ & $cm$ & $mm$ & $m$ & $cm$ & $mm$ \\  
         \hline\hline
         %4 & &\multicolumn{6}{c|}{\textbf{Circle Total Time ($s$)}} &\multicolumn{6}{c|}{\textbf{ Interlaced Circle Total Time ($s$)}}\\ 
         %\hline
         5 & PID-PPP & $1$ & 1 MP & 1 UC & 1 MP & 5 MP + 5 UC \\
         \hline
         6 & WMVN &  $1$ & 1 MP & 1 UC  & 1 MP &  5 MP + 5 UC \\
         \hline
         7 & V-SNS &  $\geq 2$ & $\geq 2$ MP & $\geq 2$ UC & $\geq 2$ MP & $\geq 2$ * (5 MP + 5 UC) \\ 
         \hline
         8 & MAN & $1$ & 1 MP & 1 UC & 1 MP & 5 MP + 5 UC \\
         \hline
         9 & A-SNS & $\geq 2$ & $\geq 2$ MP & $\geq 2$ UC & $\geq 2$ MP &  $\geq 2$ * (5 MP + 5 UC) \\
         \hline
         10 & FPBM & $3$ & 3 MP & 3 UC & 3 MP &  $3$ * (5 MP + 5 UC) \\
         \hline
        %\end{tblr}
        \end{tabular}
        % Note under the table
        \begin{tablenotes}
        \small
        \item  
        \end{tablenotes}
    \end{threeparttable}
\vspace{-4mm}
\end{table}

\begin{table}[!h] 
\notsotinyone
\caption{\footnotesize Total computation time (in $seconds$) needed to traverse all the waypoints in a trajectory with a scheme under zero noise. All the investigated trajectories had 7000 waypoints.  The $dm$ results have been omitted for space consideration. When the unit is changed from $m$ to $cm$ and $mm$, "X" indicates that the scheme was unable to preserve the same end-effector behavior.}\label{tab:time-units}
    \centering
    \begin{threeparttable}
         \setlength\tabcolsep{3.7pt}
         \begin{tabular}{|p{1.4mm}|p{8mm}|c|c|c|c|c|c|c|c|c|c|c|c|} 
         %\begin{tblr}{
         %       colspec = {|c|c|c|c|},
                %row{4} = {gray9},
                %row{12} = {gray9},
                %row{14} = {gray9},
                %row{22} = {gray9},
                %row{31} = {gray9},
                %row{32} = {gray9},
                %column{3} = {teal7},
                %cell{2}{3} = {yellow7},
        %      }
         \hline
         %\multirow{3}{3.5em}{Schemes} & \multicolumn{6}{c|}{Zero Noise} \\
         %\cline{2-7} 
         1 & \multirow{3}{3.5em}{Schemes} & \multicolumn{6}{c|}{3DoF-RRP} & \multicolumn{6}{c|}{7DoF-RRPRRRR}\\  
         \cline{3-14}
         2 & & \multicolumn{3}{c|}{Original MP-GI} & \multicolumn{3}{c|}{Proposed MX-GI} & \multicolumn{3}{c|}{Original MP-GI} & \multicolumn{3}{c|}{Proposed MX-GI}\\
         \cline{3-14} 
         3 & & $m$ & $cm$ & $mm$ & $m$ & $cm$ & $mm$ & $m$ & $cm$ & $mm$ & $m$ & $cm$ & $mm$ \\  
         \hline\hline
         4 & &\multicolumn{6}{c|}{\textbf{Path 1 - Circle}} &\multicolumn{6}{c|}{\textbf{Path 1 - Interlaced Circle}}\\ 
         \hline
         5 & PID-PPP & 3.6 & X & X & 4.1 & 4.0 & 4.1 & 6.1 & X & X & 8.9 & 8.8 & 8.9  \\
         \hline
         6 & WMVN &  4.3 & X & X & 4.8 & 4.8 & 4.8 & 6.8 & X & X & 7.8 & 7.9 & 7.8 \\
         \hline
         7 & V-SNS &  4.3 & X & X & 4.9 & 4.9 & 4.9 & 6.8 & X & X & 12.7 & 12.8 & 12.6 \\ 
         \hline
         8 & MAN & 4.9 & X & X & 5.8 & 5.9 & 5.8 & 7.6 & X & X & 9.4 & 9.6 & 9.5 \\
         \hline
         9 & A-SNS & 5.2 & X & X & 6.5 & 6.8 & 6.2 & 7.8 & X & X & 20.1 & 19.8 & 20.0 \\
         \hline
         10 & FPBM & 5.4 & X & X & 6.8 & 6.9 & 6.9 & 8.5 & X & X & 11.7 & 11.8 & 11.5 \\
         \hline
         \hline
         11 & & \multicolumn{6}{c|}{\textbf{Path 2 - Rhodonea}} & \multicolumn{6}{c|}{\textbf{Path 2 - 3D Rhodonea}}\\
         \hline
         12 & PID-PPP & 3.6 & X & X & 4.1 & 4.1 & 4.0 & 6.1 & X  & X & 8.4 &  8.7 & 8.5\\
         \hline
         13 & WMVN &  4.8 & X & X & 6.1 & 6.0 & 5.9 & 6.3 &  X  &  X  &  9.3  &  9.3 &  9.2 \\
         \hline
         14 & V-SNS & 4.9 & X & X & 5.8 & 5.7 & 5.8 & 6.7  &  X & X  &  11.7 &  11.6  &  11.3 \\ 
         \hline
         15 & MAN & 5.0 & X & X & 6.2 & 6.3 & 6.6 & 6.8 & X & X & 8.6  &  8.7 & 8.6 \\
         \hline
         16 & A-SNS & 5.1 & X & X & 7.8 & 7.9 & 7.6 & 7.2 & X & X &  15.6  & 15.5 &  15.5 \\
         \hline
         17 & FPBM & 5.1 & X & X & 7.7 & 7.4 & 7.6 & 7.7 & X &  X &  11.0  & 11.1  & 11.1 \\
         \hline
         \hline
         18 & & \multicolumn{6}{c|}{\textbf{Path 3 - Tricuspid}} & \multicolumn{6}{c|}{\textbf{Path 3 - Bent Tricuspid}}\\
         \hline
         19 & PID-PPP & 3.6 & X & X & 4.0 & 4.0 & 4.0 & 6.2 & X & X & 8.5 & 8.4 & 8.2\\
         \hline
         20 & WMVN &  4.8 & X & X & 5.4 & 5.3 & 5.3 & 6.3 & X & X & 9.1 & 9.2 & 9.2 \\
         \hline
         21 & V-SNS &  4.8 & X & X & 5.4 & 5.3 & 5.5 & 6.4 & X & X & 11.3 & 11.4 & 11.4  \\ 
         \hline
         22 & MAN &  4.9 & X & X & 5.4 & 5.5 & 5.6 & 6.3 & X & X &  9.1 & 9.1 & 9.2 \\
         \hline
         23 & A-SNS &  5.0 & X & X & 6.5 & 6.4 & 6.3 & 6.7 & X & X & 28.9 & 28.8 & 28.8  \\
         \hline
         24 & FPBM &  5.3 & X & X & 6.8 & 6.9 & 6.9 & 7.3 & X & X & 16.9 & 16.9 & 16.9  \\
         \hline
        %\end{tblr}
        \end{tabular}
        % Note under the table
        \begin{tablenotes}
        \small
        \item  
        \end{tablenotes}
    \end{threeparttable}
\vspace{-6mm}
\end{table}

\subsection{Discussion on the runtime of the investigated MP-GI- and MX-GI-based schemes}
Table \ref{tab:total-GI-calls-surveyed-schemes-methodologies} shows the number of GI calls performed by all the investigated PPP and PRMP schemes. Based on the mathematical formulations of the PID-PPP, WMVN, MAN, and FPBM schemes as presented in Table \ref{tab:surveyed-schemes-methodologies}, it can be observed that these schemes respectively need 1, 1, 1, and 3 GI calls. However, the exact number of GI calls performed by the V-SNS and A-SNS schemes, before returning a solution, is not known as the readers can observe in Algorithm 1 in \cite{flacco2015control}. The V-SNS scheme adjusts the null space joint velocity vector $\dot{Q}_{N}$ until a feasible solution is obtained while the A-SNS scheme is similarly designed at the acceleration level. Since 1 iteration, of these schemes, performs at least 2 GI calls, they require $\geq 2$ GI calls before returning a solution. Therefore, depending on the trajectory type, the number of DoF, and the GI employed; these schemes can get slower in their execution.
Table \ref{tab:time-units} shows the total computation time (in $seconds$) needed to go through all the waypoints of the 3DoF and 7DoF paths (see Figure \ref{fig:paths-investigated}) with all the investigated MP-GI- and MX-GI-based schemes without noise consideration. In Table \ref{tab:time-units}, when the units are changed from $m$ to a $cm$ and $mm$, "X" indicates that the scheme was unable to preserve the same end-effector behavior.

In the case of the 3DoF robot, the MX-GI is reduced to the UC-GI as expressed by Equation (\ref{eq:MX-3DoF}) based on the MX-GI rule of thumb. That is the time complexity of the MX-GI is equivalent to the time complexity of the UC-GI. And based on Algorithms \ref{alg:MP} and \ref{alg:UC}, the UC-GI has one more scaling decomposition step that the MP-GI does not have, hence requiring slightly longer runtime. This fact can be observed in lines 5, 6, and 8 of Table \ref{tab:time-units} where for schemes requiring 1 GI call, the MX-GI has longer but comparable runtime to the MP-GI for path 1. For schemes requiring more than 1 GI call, the MX-GI runs a little bit longer as presented in lines 7, 9, and 10 of Table \ref{tab:time-units}. The same observations can be made by looking at Table \ref{tab:time-units} in lines 12 - 17 for path 2 and lines 19 - 24 for path 3.

In the case of the 7DoF robot, the MX-GI does not reduce to either the MP-GI or the UC-GI but instead combines both GI's to guarantee unit-consistency. In such a case, the MX-GI requires 5 MP-GI and 5 UC-GI calls as showed in Table \ref{tab:total-GI-calls-surveyed-schemes-methodologies}, for schemes requiring 1 GI call in their formulation. However, for schemes requiring more than 1 GI call ($\geq$ 2 GI Calls), the MX-GI require at least 10 MP-GI and 10 UC-GI calls. This fact is verified in lines 5, 6, and 8 of Table \ref{tab:time-units} where for schemes requiring 1 GI call, the MX-GI has longer but still comparable runtime to the MP-GI for path 1. For schemes requiring more than 1 GI call, the MX-GI-based gets slower as presented in lines 7, 9, and 10. Here also, the same observations can be made by looking at lines 12 - 17 for path 2 and lines 19 - 24 for path 3 in Table \ref{tab:time-units}.

\section{Conclusion}
This paper focused on demonstrating the issues related to unit consistency of six of the various velocity-level and acceleration-level PPP and PRMP methods in the literature. We also discussed the requirements for achieving unit consistency in PPP and PRMP schemes of redundant incommensurate robotic manipulators both in the absence of noise and presence of various types of noise. We demonstrated that widely used pseudo-inverse-based schemes, which rely on the MP-GI, fail to preserve consistent trajectory planning behavior for arbitrary selections of unit and in the presence of noise when applied to incommensurate robotic manipulators even if the scheme presents noise suppression properties. Experimental results based on 3DoF (2RP) and 7DoF (2RP4R) redundant incommensurate robots clearly show that these inconsistencies can be resolved by the use of the MX-GI whether the scheme is derived from a velocity-level or an acceleration-level.  Future research will investigate the effects of unit choices on Quadratic Programming-based \cite{flacco2012motion, flacco2015control, xiao2012acceleration} and Data-Driven-based \cite{xie2021acceleration} reformulations of PPP and PRMP schemes. Currently, the unit-consistency analysis was performed based on the kinematic model of the manipulators; future analyses will include dynamic models as well. Another future work will focus on applying auto-tuning techniques (e.g., evolutionary computation methods) on the gains of the feedback-based schemes.

\begin{table}[t!]
\notsotinyone 
\centering
\caption{\footnotesize Example behavior of the SVD and UI-SVD of the Analytical Jacobian ($J_{A}$) of the 3DoF (2RP) manipulator estimated at the joint configuration $\mathbold{Q} = [\theta_1 = 30^{o}, \theta_2=30^{o}, d_3=-0.7m]$. The prismatic joint $d_3$ is varied from $m$ (meter) to $mm$ (millimeter).} 
\begin{threeparttable}
         \begin{tabular}{|c|c|c|c|}          
         %\begin{tblr}{
         %       colspec = {|c|c|c|c|c|},
         %       row{9} = {gray9},
         %       row{10} = {gray9},
                %column{3} = {teal7},
                %cell{2}{3} = {yellow7},
         %     }
         \hline
         1 & \textbf{Var.} & $m$ & $mm$\\   
         \hline\hline
         2 & \multicolumn{3}{c|}{\textbf{Geometric Jacobian \cite{sciavicco2001modelling} at $\mathbold{Q} = [\theta_1 = 30^{o}, \theta_2=30^{o}, d_3=-0.7m]$}}\\  
         \hline
         3 & $J_{A}$ 
          & 
          ${\notsotinyone{} \begin{bmatrix}
            -1.80 &  -1.30  &  0.86\\
             0.80 &  -0.05  & -0.50\\
            \end{bmatrix}{\notsotinyone}}$
          & 
          ${\notsotinyone{} \begin{bmatrix}
             -1800 & -1300 &   0.86 \\
               800 &  -50 &  -0.50 \\
            \end{bmatrix}{\notsotinyone}}$ \\
         \hline
         4 & \multicolumn{3}{c|}{\textbf{Singular Value Decomposition (SVD) of $J_{G}$}}\\  
         \hline
         5 & $U$ 
          & 
        ${\notsotinyone{} \begin{bmatrix}
                0.94 &  0.31 \\
               -0.31 & 0.94 \\
            \end{bmatrix}{\notsotinyone}}$
          & 
        ${\notsotinyone{} \begin{bmatrix}      
                0.95 & 0.28\\
               -0.28 & 0.95 \\
            \end{bmatrix}{\notsotinyone}}$ \\
         \hline
         6 & $S$ 
          & 
        ${\notsotinyone{} \begin{bmatrix}
                   2.48 &   0.0 & 0.0\\
                   0.0 &  0.54 & 0.0\\
            \end{bmatrix}{\notsotinyone}}$
          & 
        ${\notsotinyone{} \begin{bmatrix}
                   2309.3 &  0.0 & 0.0 \\
                   0.0 &  489.32 & 0.0 \\
            \end{bmatrix}{\notsotinyone}}$ \\ 
         \hline
         7 & $V$ 
          & 
        ${\notsotinyone{} \begin{bmatrix}
                   -0.78 &   0.33 & 0.51 \\
                   -0.48 &  -0.85 & -0.19\\
                    0.36  &  -0.40 & 0.83\\
            \end{bmatrix}{\notsotinyone}}$
          & 
        ${\notsotinyone{} \begin{bmatrix}
                   -0.84  &  0.53 & 6.1E-4 \\
                   -0.53 &  -0.84  & 2.3E-4 \\
                    3.9E-4 & -5.2E-4 & 1\\
            \end{bmatrix}{\notsotinyone}}$ \\
         \hline
         8 & \multicolumn{3}{c|}{\textbf{Scaling Decomposition \cite{rothblum1992scalings}  of $J_{G}$}} \\  
         \hline
         9 & $E$ 
          & 
        ${\notsotinyone{} \begin{bmatrix}
                   0.46 &  0.0\\
                   0.0 &  2.13\\
            \end{bmatrix}{\notsotinyone}}$
          & 
        ${\notsotinyone{} \begin{bmatrix}
                   0.46 &  0.0\\
                   0.0 &  2.13\\
            \end{bmatrix}{\notsotinyone}}$ 
        \\
         \hline
         10 & $S_{UI}$  
          & 
        ${\notsotinyone{} \begin{bmatrix}
                   -0.70 &   -2.39 & 0.59\\
                   1.42 &  -0.42 & -1.68\\
            \end{bmatrix}{\notsotinyone}}$
          & 
        ${\notsotinyone{} \begin{bmatrix}
                   -0.70 &   -2.39 & 0.59\\
                   1.42 &  -0.42 & -1.68\\
            \end{bmatrix}{\notsotinyone}}$
        \\
         \hline
         11 & $D$ 
          & 
        ${\notsotinyone{} \begin{bmatrix}
                   0.83 &   0.0 & 0.0\\
                   0.0 &   3.92 & 0.0\\
                   0.0 &   0.0 &  1.58\\
            \end{bmatrix}{\notsotinyone}}$
          & 
        ${\notsotinyone{} \begin{bmatrix}
                   8.3E-4 &  0.0 & 0.0\\
                   0.0 & 3.9E-3  & 0.0\\
                   0.0 &  0.0 & 1.58\\
            \end{bmatrix}{\notsotinyone}}$
        \\
         \hline
         12 & \multicolumn{3}{c|}{\textbf{Singular Value Decomposition of $S_{UI}$}} \\  
         \hline
         13 & $U_{S_{UI}}$ 
          & 
        ${\notsotinyone{} \begin{bmatrix}
                   0.89 &  0.44\\
                   -0.44.0 &  0.89\\
            \end{bmatrix}{\notsotinyone}}$
          & 
        ${\notsotinyone{} \begin{bmatrix}
                   0.89 &  0.44\\
                   -0.44.0 &  0.89\\
            \end{bmatrix}{\notsotinyone}}$ 
        \\
         \hline         
         14 & $S_{S_{UI}}$  
          & 
        ${\notsotinyone{} \begin{bmatrix}
                   2.65 &   0.0 & 0.0 \\
                   0.0 &  2.13 & 0.0 \\
            \end{bmatrix}{\notsotinyone}}$
          & 
        ${\notsotinyone{} \begin{bmatrix}
                   2.65 &   0.0 & 0.0 \\
                   0.0 &  2.13 & 0.0 \\
            \end{bmatrix}{\notsotinyone}}$
        \\
         \hline         
         15 & $V_{S_{UI}}$ 
          & 
        ${\notsotinyone{} \begin{bmatrix}
           -0.47 &  0.45  &  0.75\\
           -0.73 &  -0.67 & -0.06\\
            0.47 &  -0.58  &  0.65\\
            \end{bmatrix}{\notsotinyone}}$
          & 
        ${\notsotinyone{} \begin{bmatrix}
           -0.47 &  0.45  &  0.75\\
           -0.73 &  -0.67 & -0.06\\
            0.47 &  -0.58  &  0.65\\
            \end{bmatrix}{\notsotinyone}}$
        \\
         \hline
        \end{tabular}
        %\end{tblr}
    \end{threeparttable}
    \label{tab:analysis-3DoF}
    \vspace{-2.5mm}
\end{table}

\begin{figure}[!t]
\centering
\begin{subfigure}[b]{.24\textwidth}
  \centering
  \includegraphics[width=\textwidth,trim={1cm 1cm 1.2cm 1.2cm},clip]{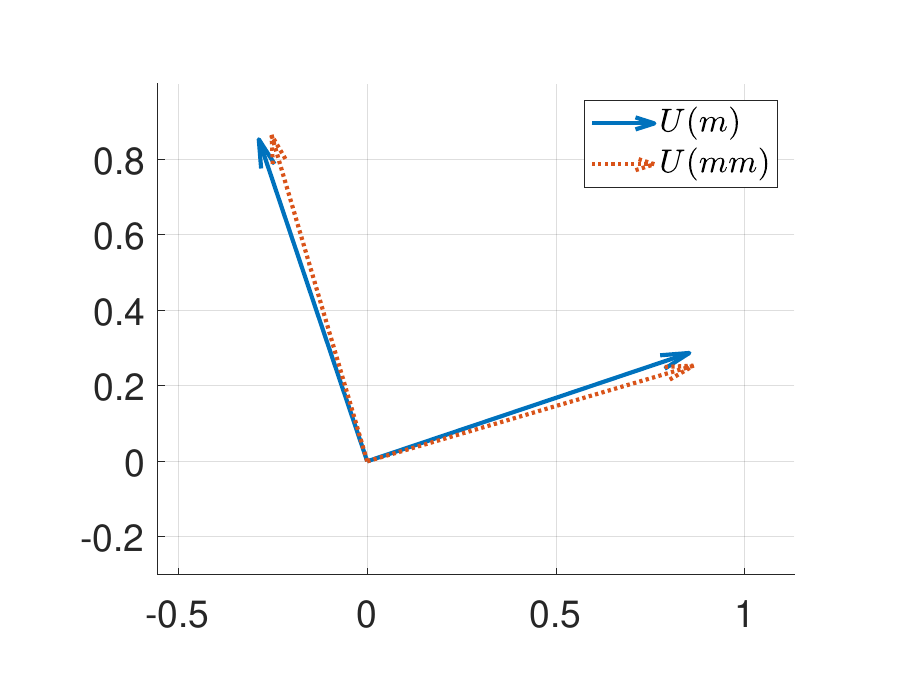} 
  \caption{\footnotesize SVD left singular vectors}
  \label{fig:left-U-SVD}
\end{subfigure}
\begin{subfigure}[b]{.24\textwidth}
  \centering
  \includegraphics[width=\textwidth,trim={1cm 0.9cm 1.2cm 1.2cm},clip]{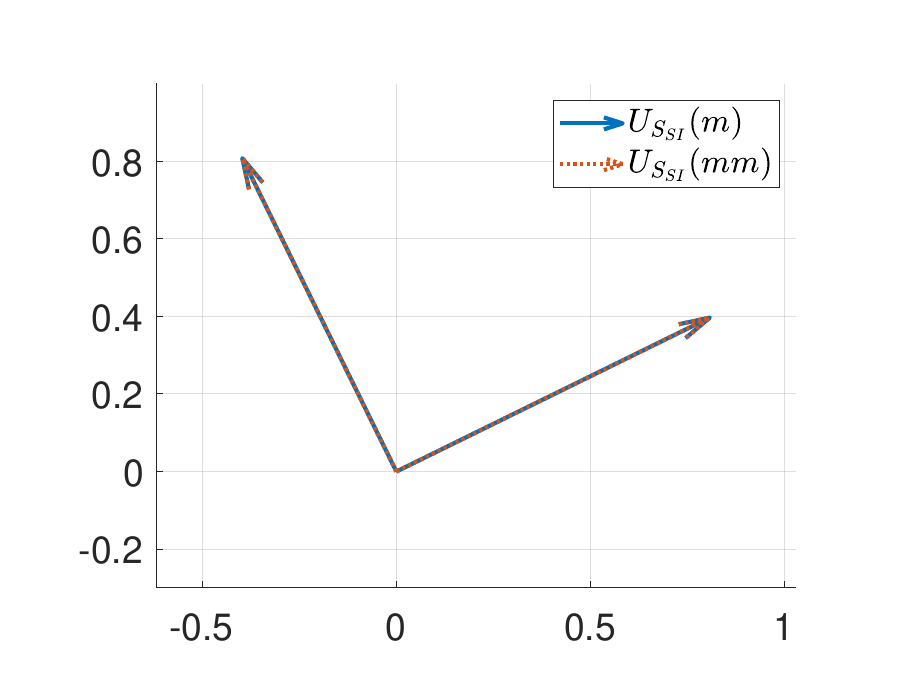} 
  \caption{\footnotesize UI-SVD left singular vectors}
  \label{fig:left-U-UI-SVD}
\end{subfigure}
\begin{subfigure}[b]{.24\textwidth}
  \centering
  \includegraphics[width=\textwidth,trim={3cm 1cm 3cm 1.2cm},clip]{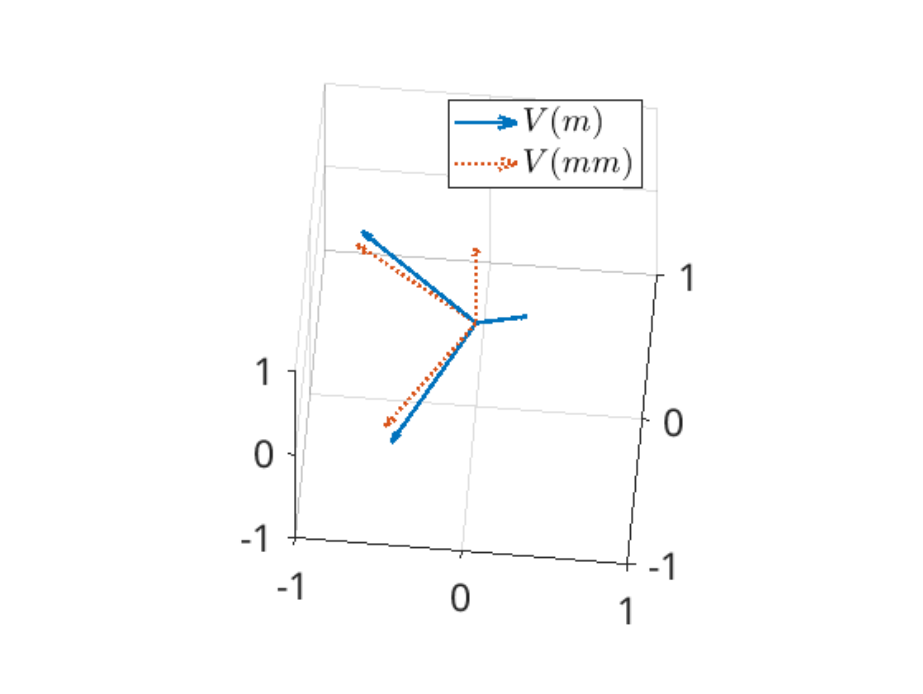} 
  \caption{\footnotesize SVD right singular vectors}
  \label{fig:right-V-SVD}
\end{subfigure}
\begin{subfigure}[b]{.24\textwidth}
  \centering
  \includegraphics[width=\textwidth,trim={3cm 0.9cm 3cm 1.2cm},clip]{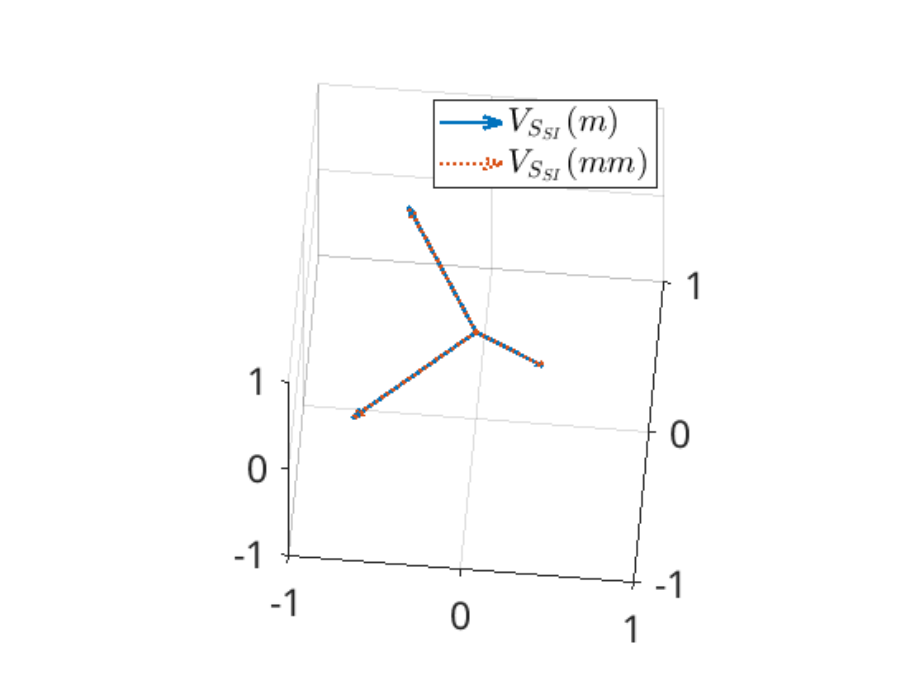} 
  \caption{\footnotesize UI-SVD right singular vectors}
  \label{fig:right-V-UI-SVD}
\end{subfigure}
\caption{\footnotesize Comparison between the left (\ref{fig:left-U-SVD}), (\ref{fig:left-U-UI-SVD})  and right singular vectors  (\ref{fig:right-V-SVD}), (\ref{fig:right-V-UI-SVD}) obtained respectively with the SVD and UI-SVD decompositions of $J_{G}$ in $m$ and $mm$.}
\label{fig:comparison-left-and-right-singular-vectors}
\vspace{-4mm}
\end{figure}

\appendix{}\label{sec:numerical-example}  
To understand the unit-consistency issues related to the use of the MP-GI for incommensurate robotic manipulators, consider the following example of a 3DoF (RRP) planar robot from Table \ref{tab:3DoF-DH-parameters}. Based on the Denavit-Hartenberg (DH) methodology and the Forward Kinematics $T$, the expression of its Analytical Jacobian ($J_{A}$) can be find by taking the derivative of $D$ (last column of $T$) as follows:

%\begin{equation}
%    \begin{split}
%        {}^{0}T_{n} &= \prod^{n}_{i=0} A_{i} = \begin{bmatrix}
%            cos(\theta)
%        \end{bmatrix}
%    \end{split}
%\end{equation}

\begin{equation}\label{eq:analytical-jacobian}
\notsotinyfour
    \begin{split}
        J_{A} &= \frac{\partial D}{\partial Q} = \begin{bmatrix}
            -a_{1}S_{1} - a_{2}S_{12} + d_{3}C_{12} & -a_{2}S_{12} + d_{3}C_{12} & S_{12} \\
            a_{1}C_{1} + a_{2}C_{12} + d_{3}S_{12} & a_{2}C_{12} + d_{3}S_{12} & -C_{12} \\
        \end{bmatrix}
    \end{split}
\end{equation}
where $D \in  \mathbb{R}^{m=2}$ is the position vector, $Q = \begin{bmatrix} \theta_{1}, \theta_{2}, d_{3} \end{bmatrix}^{T}  \in  \mathbb{R}^{n=3}$ is the joint configuration vector, $a_{1}$ and $a_{2}$ are link lengths as described in Table \ref{tab:3DoF-DH-parameters}, $S_{1} = sin(\theta_{1})$,  $C_{1} = cos(\theta_{1})$,  $S_{12} = sin(\theta_{1} + \theta_{2})$, and $C_{12} = cos(\theta_{1} + \theta_{2})$. From equation (\ref{eq:analytical-jacobian}), it can be clealy seen that changing the unit of the prismatic joint $d_{3}$ only affects the two first columns of $J_{A}$. For $Q =  \begin{bmatrix} \theta_1 = 30^{o},\theta_2=30^{o},d_3=-0.7m \end{bmatrix}^{T} \in \mathbb{R}^{3}$, $J_{A}$ is given in row 3 of Table \ref{tab:analysis-3DoF} when $d_{3}$ is expressed in $m$ and $mm$. From Table \ref{tab:analysis-3DoF}, the MP-GI ($J^{-MP}_{A}$) can be derived using the Singular Value Decomposition (SVD) of $J_{A}$ shown in rows 5, 6, and 7 which respectively denotes the left singular vectors ($U$), the singular values ($S$), and the right singular vectors ($V$). It can be clearly seen that choosing another unit to express the prismatic joint $d_{3}$ leads to completely different SVD decompositions as $U$, $S$, and $V$ in $m$ and $mm$ are not related by an order of magnitude or a scale. This observation is visualized in Figures \ref{fig:left-U-SVD} and \ref{fig:right-V-SVD} where the basis formed by those vectors in $m$ and $mm$ do not coincide. Hence, $J^{MP}_{A}$ will lead to inconsistencies in the calculations of joint velocities or joint accelerations in equations (\ref{eq:velocify-formulation}) and (\ref{eq:acceleration-formulation}). On the other hand, the UC-GI ($J^{-UC}_{A}$) relies on the Unit-Invariant Singular Value Decomposition \cite{uhlmann2018generalized} (UI-SVD) which combines the Scaling Decomposition \cite{rothblum1992scalings} (SD) and the Singular Value Decomposition (SVD). First, UI-SVD uses SD, as shown in rows 9, 10, and 11, to capture the change of units or scaling of $J_{A}$ from $m$ to $mm$ in the left ($E$) and right ($D$) scaling matrices. Such a decomposition clearly shows that $J_{A}$ is a scaling of $S_{UI}$. Then, UI-SVD applies SVD to $S_{UI}$. Free of any scaling influence, the unit-invariant left singular vectors ($U_{S_{UI}}$ ), singular values ($S_{S_{UI}}$), and right singular vectors ($V_{S_{UI}}$), as shown respectively in rows 13, 14 and 15, are identical for $m$ and $mm$. Figures \ref{fig:left-U-UI-SVD} and \ref{fig:right-V-UI-SVD} show how the basis formed by those vectors in $m$ and $mm$ coincide. That is, $J^{UC}_{G}$ will lead to unit-consistent results in equations (\ref{eq:velocify-formulation}) and (\ref{eq:acceleration-formulation}).

Only analysis of $d_{3}$ expressed in $m$ and $mm$ based on an Analytical Jacobian are provided in Table \ref{tab:analysis-3DoF} and Figure \ref{fig:comparison-left-and-right-singular-vectors}, however similar behaviors are observed with other jacobian types (Geometric, Numerical, and Elementary Transform Sequence Jacobians) and when $d_{3}$ is expressed in $dm$ and $cm$. That is the unit-consistency issues observed when using the MP-GI are unrelated to the type of Jacobians used.

\ifCLASSOPTIONcaptionsoff
  \newpage
\fi

% trigger a \newpage just before the given reference
% number - used to balance the columns on the last page
% adjust value as needed - may need to be readjusted if
% the document is modified later
%\IEEEtriggeratref{8}
% The "triggered" command can be changed if desired:
%\IEEEtriggercmd{\enlargethispage{-5in}}

% references section

% can use a bibliography generated by BibTeX as a .bbl file
% BibTeX documentation can be easily obtained at:
% http://mirror.ctan.org/biblio/bibtex/contrib/doc/
% The IEEEtran BibTeX style support page is at:
% http://www.michaelshell.org/tex/ieeetran/bibtex/
%\bibliographystyle{IEEEtran}
% argument is your BibTeX string definitions and bibliography database(s)
%\bibliography{IEEEabrv,../bib/paper}
%
% <OR> manually copy in the resultant .bbl file
% set second argument of \begin to the number of references
% (used to reserve space for the reference number labels box)
%\begin{thebibliography}{1}

%\bibitem{IEEEhowto:kopka}
%H.~Kopka and P.~W. Daly, \emph{A Guide to \LaTeX}, 3rd~ed.\hskip %1em plus
%  0.5em minus 0.4em\relax Harlow, England: Addison-Wesley, 1999.

%\end{thebibliography}

%\printbibliography[]
%\vspace{-2mm}
\bibliographystyle{IEEEtran}
\bibliography{IEEEabrv, main.bib}

% Generated by IEEEtran.bst, version: 1.14 (2015/08/26)
\begin{thebibliography}{10}
\providecommand{\url}[1]{#1}
\csname url@samestyle\endcsname
\providecommand{\newblock}{\relax}
\providecommand{\bibinfo}[2]{#2}
\providecommand{\BIBentrySTDinterwordspacing}{\spaceskip=0pt\relax}
\providecommand{\BIBentryALTinterwordstretchfactor}{4}
\providecommand{\BIBentryALTinterwordspacing}{\spaceskip=\fontdimen2\font plus
\BIBentryALTinterwordstretchfactor\fontdimen3\font minus
  \fontdimen4\font\relax}
\providecommand{\BIBforeignlanguage}[2]{{%
\expandafter\ifx\csname l@#1\endcsname\relax
\typeout{** WARNING: IEEEtran.bst: No hyphenation pattern has been}%
\typeout{** loaded for the language `#1'. Using the pattern for}%
\typeout{** the default language instead.}%
\else
\language=\csname l@#1\endcsname
\fi
#2}}
\providecommand{\BIBdecl}{\relax}
\BIBdecl

\bibitem{faroni2018predictive}
M.~Faroni, M.~Beschi, N.~Pedrocchi, and A.~Visioli, ``Predictive inverse
  kinematics for redundant manipulators with task scaling and kinematic
  constraints,'' \emph{IEEE Transactions on Robotics}, vol.~35, no.~1, pp.
  278--285, 2018.

\bibitem{faroni2020inverse}
M.~Faroni, M.~Beschi, and N.~Pedrocchi, ``Inverse kinematics of redundant
  manipulators with dynamic bounds on joint movements,'' \emph{IEEE Robotics
  and Automation Letters}, vol.~5, no.~4, pp. 6435--6442, 2020.

\bibitem{park2020trajectory}
S.-O. Park, M.~C. Lee, and J.~Kim, ``Trajectory planning with collision
  avoidance for redundant robots using jacobian and artificial potential
  field-based real-time inverse kinematics,'' \emph{International Journal of
  Control, Automation and Systems}, vol.~18, no.~8, pp. 2095--2107, 2020.

\bibitem{ademovic2016path}
A.~Ademovic and B.~Lacevic, ``Path planning for robotic manipulators using
  expanded bubbles of free c-space,'' in \emph{2016 IEEE International
  Conference on Robotics and Automation (ICRA)}.\hskip 1em plus 0.5em minus
  0.4em\relax IEEE, 2016, pp. 77--82.

\bibitem{lacevic2020improved}
B.~Lacevic and D.~Osmankovic, ``Improved c-space exploration and path planning
  for robotic manipulators using distance information,'' in \emph{2020 IEEE
  International Conference on Robotics and Automation (ICRA)}.\hskip 1em plus
  0.5em minus 0.4em\relax IEEE, 2020, pp. 1176--1182.

\bibitem{guo2020repetitive}
D.~Guo, Z.~Li, A.~H. Khan, Q.~Feng, and J.~Cai, ``Repetitive motion planning of
  robotic manipulators with guaranteed precision,'' \emph{IEEE Transactions on
  Industrial Informatics}, vol.~17, no.~1, pp. 356--366, 2020.

\bibitem{chan1995weighted}
T.~F. Chan and R.~V. Dubey, ``A weighted least-norm solution based scheme for
  avoiding joint limits for redundant joint manipulators,'' \emph{IEEE
  Transactions on Robotics and Automation}, vol.~11, no.~2, pp. 286--292, 1995.

\bibitem{pardi2020path}
T.~Pardi, V.~Maddali, V.~Ortenzi, R.~Stolkin, and N.~Marturi, ``Path planning
  for mobile manipulator robots under non-holonomic and task constraints,'' in
  \emph{2020 IEEE/RSJ International Conference on Intelligent Robots and
  Systems (IROS)}.\hskip 1em plus 0.5em minus 0.4em\relax IEEE, 2020, pp.
  6749--6756.

\bibitem{corke2011robotics}
P.~I. Corke and O.~Khatib, \emph{Robotics, vision and control: fundamental
  algorithms in MATLAB}.\hskip 1em plus 0.5em minus 0.4em\relax Springer, 2011,
  vol.~73.

\bibitem{flacco2012motion}
F.~Flacco, A.~De~Luca, and O.~Khatib, ``Motion control of redundant robots
  under joint constraints: Saturation in the null space,'' in \emph{2012 IEEE
  International Conference on Robotics and Automation}.\hskip 1em plus 0.5em
  minus 0.4em\relax IEEE, 2012, pp. 285--292.

\bibitem{flacco2015control}
------, ``Control of redundant robots under hard joint constraints: Saturation
  in the null space,'' \emph{IEEE Transactions on Robotics}, vol.~31, no.~3,
  pp. 637--654, 2015.

\bibitem{guo2017new}
D.~Guo, F.~Xu, and L.~Yan, ``New pseudoinverse-based path-planning scheme with
  pid characteristic for redundant robot manipulators in the presence of
  noise,'' \emph{IEEE Transactions on Control Systems Technology}, vol.~26,
  no.~6, pp. 2008--2019, 2017.

\bibitem{wang2019feedback}
Z.~Wang, B.~Wang, L.~Xu, and Q.~Xie, ``Feedback-added pseudoinverse-type
  balanced minimization scheme for kinematic control of redundant robot
  manipulators,'' \emph{IEEE Access}, vol.~7, pp. 23\,806--23\,815, 2019.

\bibitem{lanari1992control}
A.~De~Luca, L.~Lanari, and G.~Oriolo, ``Control of redundant robots on cyclic
  trajectories,'' in \emph{Proc. 1992 IEEE Internat. Conf. on Robotics and
  Automation, Nice}, 1992, pp. 500--506.

\bibitem{li2018new}
Z.~Li, B.~Liao, F.~Xu, and D.~Guo, ``A new repetitive motion planning scheme
  with noise suppression capability for redundant robot manipulators,''
  \emph{IEEE Transactions on Systems, Man, and Cybernetics: Systems}, vol.~50,
  no.~12, pp. 5244--5254, 2020.

\bibitem{xie2021acceleration}
Z.~Xie, L.~Jin, X.~Luo, B.~Hu, and S.~Li, ``An acceleration-level data-driven
  repetitive motion planning scheme for kinematic control of robots with
  unknown structure,'' \emph{IEEE Transactions on Systems, Man, and
  Cybernetics: Systems}, 2021.

\bibitem{sciavicco2001modelling}
L.~Sciavicco and B.~Siciliano, \emph{Modelling and control of robot
  manipulators}.\hskip 1em plus 0.5em minus 0.4em\relax Springer Science \&
  Business Media, 2001.

\bibitem{farzan2013dh}
S.~Farzan and G.~N. DeSouza, ``From dh to inverse kinematics: A fast numerical
  solution for general robotic manipulators using parallel processing,'' in
  \emph{2013 IEEE/RSJ International Conference on Intelligent Robots and
  Systems}.\hskip 1em plus 0.5em minus 0.4em\relax IEEE, 2013, pp. 2507--2513.

\bibitem{mayer1981differential}
G.~E. Mayer, R.~Paul, and B.~Shimano, ``Differential kinematic control
  equations for simple manipulators,'' \emph{IEEE Trans. Syst. Man Cybern},
  vol.~1, no.~1, 1981.

\bibitem{haviland2020systematic}
J.~Haviland and P.~Corke, ``A systematic approach to computing the manipulator
  jacobian and hessian using the elementary transform sequence,'' \emph{arXiv
  preprint arXiv:2010.08696}, 2020.

\bibitem{penrose1955generalized}
R.~Penrose, ``A generalized inverse for matrices,'' in \emph{Mathematical
  proceedings of the Cambridge philosophical society}, vol.~51, no.~3.\hskip
  1em plus 0.5em minus 0.4em\relax Cambridge University Press, 1955, pp.
  406--413.

\bibitem{chan1988general}
S.~K. Chan and P.~D. Lawrence, ``General inverse kinematics with the error
  damped pseudoinverse,'' in \emph{Proceedings. 1988 IEEE international
  conference on robotics and automation}.\hskip 1em plus 0.5em minus
  0.4em\relax IEEE, 1988, pp. 834--839.

\bibitem{chiaverini1991achieving}
S.~Chiaverini, O.~Egeland, and R.~Kanestrom, ``Achieving user-defined accuracy
  with damped least-squares inverse kinematics,'' in \emph{Fifth International
  Conference on Advanced Robotics' Robots in Unstructured Environments}.\hskip
  1em plus 0.5em minus 0.4em\relax IEEE, 1991, pp. 672--677.

\bibitem{chiaverini1994review}
S.~Chiaverini, B.~Siciliano, and O.~Egeland, ``Review of the damped
  least-squares inverse kinematics with experiments on an industrial robot
  manipulator,'' \emph{IEEE Transactions on control systems technology},
  vol.~2, no.~2, pp. 123--134, 1994.

\bibitem{buss2005selectively}
S.~R. Buss and J.-S. Kim, ``Selectively damped least squares for inverse
  kinematics,'' \emph{Journal of Graphics tools}, vol.~10, no.~3, pp. 37--49,
  2005.

\bibitem{sugihara2011solvability}
T.~Sugihara, ``Solvability-unconcerned inverse kinematics by the
  levenberg--marquardt method,'' \emph{IEEE Transactions on Robotics}, vol.~27,
  no.~5, pp. 984--991, 2011.

\bibitem{colome2012redundant}
A.~Colom{\'e} and C.~Torras, ``Redundant inverse kinematics: Experimental
  comparative review and two enhancements,'' in \emph{2012 IEEE/RSJ
  International Conference on Intelligent Robots and Systems}.\hskip 1em plus
  0.5em minus 0.4em\relax IEEE, 2012, pp. 5333--5340.

\bibitem{uhlmann2018generalized}
J.~Uhlmann, ``A generalized matrix inverse that is consistent with respect to
  diagonal transformations,'' \emph{SIAM Journal on Matrix Analysis and
  Applications}, vol.~39, no.~2, pp. 781--800, 2018.

\bibitem{angeles2003fundamentals}
J.~Angeles, \emph{Fundamentals of robotic mechanical systems: theory, methods,
  and algorithms}.\hskip 1em plus 0.5em minus 0.4em\relax Springer, 2003.

\bibitem{siciliano2008springer}
B.~Siciliano, O.~Khatib, and T.~Kr{\"o}ger, \emph{Springer handbook of
  robotics}.\hskip 1em plus 0.5em minus 0.4em\relax Springer, 2008, vol. 200.

\bibitem{zhang2019applying}
B.~Zhang and J.~Uhlmann, ``Applying a unit-consistent generalized matrix
  inverse for stable control of robotic systems,'' \emph{Journal of Mechanisms
  and Robotics}, vol.~11, no.~3, p. 034503, 2019.

\bibitem{zhang2020examining}
------, ``Examining a mixed inverse approach for stable control of a rover,''
  \emph{International Journal of Control Systems and Robotics}, vol.~5, 2020.

\bibitem{demby2020use}
U.~J.~T. Demby's, ``Use of jacobians for inverse kinematics of articulated
  robots: a study on approximate solutions,'' Ph.D. dissertation, University of
  Missouri--Columbia, 2020.

\bibitem{dembys2023choosing}
J.~Demby's, J.~Uhlmann, and G.~N. DeSouza, ``Choosing the correct generalized
  inverse for the numerical solution of the inverse kinematics of
  incommensurate robotic manipulators,'' \emph{arXiv preprint}, 2023.

\bibitem{schwartz2002noncommensurate}
E.~Schwartz, R.~Manseur, and K.~Doty, ``Noncommensurate systems in robotics,''
  \emph{International Journal of Robotics and Automation}, vol.~17, no.~2, pp.
  86--92, 2002.

\bibitem{schwartz2003non}
E.~M. Schwartz, R.~Manseur, and K.~L. Doty, ``Non-commensurate manipulator
  jacobian.'' in \emph{Robotics and Applications}, 2003, pp. 112--115.

\bibitem{klein1983review}
C.~A. Klein and C.-H. Huang, ``Review of pseudoinverse control for use with
  kinematically redundant manipulators,'' \emph{IEEE Transactions on Systems,
  Man, and Cybernetics}, no.~2, pp. 245--250, 1983.

\bibitem{zghal1990efficient}
H.~Zghal, R.~V. Dubey, and J.~A. Euler, ``Efficient gradient projection
  optimization for manipulators with multiple degrees of redundancy,'' in
  \emph{Proceedings., IEEE International Conference on Robotics and
  Automation}.\hskip 1em plus 0.5em minus 0.4em\relax IEEE, 1990, pp.
  1006--1011.

\bibitem{li2017novel}
S.~Li, H.~Wang, and M.~U. Rafique, ``A novel recurrent neural network for
  manipulator control with improved noise tolerance,'' \emph{IEEE transactions
  on neural networks and learning systems}, vol.~29, no.~5, pp. 1908--1918,
  2017.

\bibitem{deo1997minimum}
A.~S. Deo and I.~D. Walker, ``Minimum effort inverse kinematics for redundant
  manipulators,'' \emph{IEEE transactions on robotics and automation}, vol.~13,
  no.~5, pp. 767--775, 1997.

\bibitem{lu2002inverses}
T.-T. Lu and S.-H. Shiou, ``Inverses of 2$\times$ 2 block matrices,''
  \emph{Computers \& Mathematics with Applications}, vol.~43, no. 1-2, pp.
  119--129, 2002.

\bibitem{xiao2012acceleration}
L.~Xiao and Y.~Zhang, ``Acceleration-level repetitive motion planning and its
  experimental verification on a six-link planar robot manipulator,''
  \emph{IEEE Transactions on Control Systems Technology}, vol.~21, no.~3, pp.
  906--914, 2012.

\bibitem{rothblum1992scalings}
U.~G. Rothblum and S.~A. Zenios, ``Scalings of matrices satisfying line-product
  constraints and generalizations,'' \emph{Linear algebra and its
  applications}, vol. 175, pp. 159--175, 1992.

\end{thebibliography}

\end{document}